\ifdefined\PASubmission
  \documentclass[12pt,letterpaper]{article}
\else
  \documentclass[11pt,letterpaper]{article}
\fi
\usepackage{longtable}

\usepackage{booktabs}
\usepackage[compact]{titlesec}
\usepackage{titletoc}
\usepackage[table]{xcolor}% 
\usepackage{latexsym}
\usepackage{amssymb,amsmath, bm,pgfplots,tikz,bbm}
\usepackage{graphicx}
\graphicspath{{../}}
\usepackage{marvosym}
\usepackage{multirow,float}
\usepackage{algpseudocode}
\usepackage{caption}
\usepackage{mathtools}

\usepackage{subcaption}
\usepackage{comment}
\usepackage{enumitem}
\usepackage{longtable}

\usepackage{helvet}   % Helvetica for the figure only; does NOT change your body font

\newcommand{\bul}[1]{%
  \begin{itemize}[leftmargin=1.25em,itemsep=2pt,topsep=3pt,parsep=0pt,
                  label={\small\textbullet}]#1\end{itemize}}
                  
\usepackage{natbib}
\usepackage{color}
\usepackage[bookmarksopen=true, bookmarksnumbered=true,
pdfstartview=FitH, breaklinks=true, urlbordercolor={0 1 0}, citebordercolor={0 0 1}]{hyperref}
\usepackage{cleveref}
\usepackage{dcolumn}
\newcolumntype{.}{D{.}{.}{-1}}
\newcolumntype{d}[1]{D{.}{.}{#1}}

\usepackage{theorem}
\theoremstyle{plain}
\theoremheaderfont{\scshape}

\newtheorem{assumption}{Assumption}

\newcommand{\qed}{\hfill \ensuremath{\Box}}
\newcommand{\indep}{\mbox{$\perp\!\!\!\perp$}}

\usepackage{kantlipsum}
\allowdisplaybreaks

\usepackage{rotating}

\usepackage{arydshln}
\usepackage{threeparttable}

\usepackage{booktabs,tabularx,array}

\usepackage[compact]{titlesec}
\usepackage[listings,breakable]{tcolorbox}
\usepackage[ruled,linesnumbered,vlined]{algorithm2e}
\usepackage{pifont}

\newcommand\E{\mathbb{E}}

\newcommand\bD{\bm{D}}

\newcommand\bX{\bm{X}}

\newcommand\bZ{\bm{Z}}

\newcommand\cY{\mathcal{Y}}

\newcommand\bgamma{\boldsymbol{\gamma}}

\usetikzlibrary{decorations.markings}
\usetikzlibrary{positioning,calc}
\usetikzlibrary{decorations.pathmorphing}
\usetikzlibrary{shapes.geometric, arrows}
\usetikzlibrary{arrows,decorations.pathmorphing,backgrounds,positioning,fit,matrix}
\usetikzlibrary{shapes,decorations,arrows,calc,arrows.meta,fit,positioning}
\tikzset{auto,node distance =1 cm and 1 cm,semithick,
	state/.style ={circle, draw, minimum width = 0.7 cm},
	point/.style = {circle, draw, inner sep=0.04cm,fill,node contents={}},
	bidirected/.style={Latex-Latex,dashed},
	el/.style = {inner sep=2pt, align=left, sloped}
}

\newcommand{\tp}{{\scriptscriptstyle\mathsf{T}}}
\graphicspath{{../}{./}}
\usepackage{comment}
\usepackage{bbm}

\usepackage{booktabs}
\usepackage{array}
\usepackage{ragged2e}
\usepackage{newtxtext}

\ifdefined\PASubmission
  \usepackage{setspace}
\fi

\newcolumntype{L}[1]{>{\RaggedRight\arraybackslash}p{#1}}

\newcommand{\PaperAcknowledgements}{We thank António Câmara, Stephen Chaudoin, Danny Ebanks, Valentina González-Rostani, Musashi Hinck, Connor Jerzak, Gary King, Dean Knox, Eddie Yang, and participants of PolMeth 2026 for helpful comments.
We are also grateful to the five subject-matter experts who generously contributed their time and expertise by independently annotating texts for our validation study.
We also thank an anonymous reviewer from Harvard’s Alexander and Diviya Magaro Peer Pre-Review Program
for their thoughtful comments and suggestions.}
\newcommand{\PaperFinancialSupport}{This work was supported by the Institute for Quantitative Social Science (IQSS).}

\begin{document}
\ifdefined\PASubmission
  \doublespacing
\fi
% suggested alt titles: "Observational Equivalence in Human and LLM Text Annotation" and "LLM and Human Text Annotations Are Observationally Equivalent"
\title{\bf Observational Equivalence of LLM and Human Annotation%
\ifdefined\PASubmission\else
  \thanks{\PaperAcknowledgements\ We also gratefully acknowledge financial support from the Institute for Quantitative Social Science (IQSS).}
\fi}
\author{Kentaro Nakamura\thanks{Ph.D. student at John F. Kennedy School of Government, Harvard University, Email: \href{mailto:knakamura@g.harvard.edu}{knakamura@g.harvard.edu}} \and
Jing Ling Tan\thanks{Ph.D. student at Department of Government, Harvard University, Email: \href{mailto:jingling_tan@g.harvard.edu}{jingling\_tan@g.harvard.edu}} \and George Yean\thanks{Ph.D. student at Department of Government, Harvard University, Email: \href{mailto:gyean@g.harvard.edu}{gyean@g.harvard.edu}  }}
\date{
Last Updated: \today %\\
%\color{red} [Preliminary draft. Please do not circulate without permission.]
}
\maketitle
\ifdefined\PASubmission\else
  \thispagestyle{empty}
\fi
%\tableofcontents\thispagestyle{empty}

\begin{abstract}
\noindent 
In this paper, we show that humans and large language models (LLMs) are observationally equivalent in coding texts. Recent LLMs agree with attentive humans at rates comparable to those observed among humans, and cross-LLM disagreement concentrates
on texts that also divide humans. We demonstrate these patterns through
replications of text-classification tasks from 14 peer-reviewed political
science studies, with 13 LLMs, three human coders (Ph.D. students), and 165 crowd coders, independently
applying identical codebooks.
We further find that clarifying an underspecified codebook increases
agreement among humans and between humans and
LLMs. Together, these findings point to ambiguity in texts and coding
rules as a shared source of disagreement and challenge the presumption
that human labels should take precedence whenever humans and LLMs
differ. We therefore argue that the central challenge of text annotation
is not simply choosing between humans and LLMs, but developing
coding rules that minimize ambiguity and accounting for the ambiguity
that remains. To this end, we propose using cross-LLM disagreement to
identify difficult cases, guide human review, and refine codebooks, and
we develop ambiguity-aware bounds for downstream inference without
requiring ground truth for every text.
\end{abstract}

\noindent {\bf Keywords:}  large language models, automated text analysis, Text-as-Data, AI

\newpage

\section{Introduction} 

% paragraph 1
Large language models (LLMs) are changing quantitative text analysis by 
applying natural-language instructions directly to raw text. 
Earlier text-as-data methods typically relied on dictionaries, bag-of-words 
representations, or parametric models that abstracted away from 
how people interpret language \citep{grimmer_text_2013, grimmer2021machine, grimmer2022text}. 
Because these methods were generally regarded as imperfect approximations of human reading, 
researchers commonly validated automated annotations against hand-coded labels, treating disagreements as algorithmic errors. 
LLMs challenge this human-centered validation paradigm. By incorporating definitions, examples, and contextual information into a single prompt, researchers can instruct LLMs to apply codebooks in ways that resemble human annotation. This raises two fundamental questions: Can LLMs annotate texts as reliably as human coders, and what does their performance imply for how researchers should conduct and validate quantitative text analysis?

% paragraph 2 (finding 1)
In this paper, we evaluate LLM-based annotation by replicating binary
text-classification tasks from 14 peer-reviewed political science studies.
For each study, 13 LLMs, three human coders (Ph.D. students with training in
political science), and crowd workers independently classify the same texts using identical
codebooks, allowing direct comparison of LLM--human and human--human
agreement. We find that, except for an older, weaker model, recent LLMs agree with humans
at rates comparable to those among humans themselves. Across studies,
LLM--human agreement closely tracks human--human agreement: LLMs agree
more with humans when human agreement is higher, and less when it declines.

% paragraph 3 (finding 2)
We further find that disagreement among different 
annotators is systematically concentrated in the same texts. 
When humans disagree about a text, 
LLMs and crowd-sourced workers also exhibit substantially greater disagreement, although crowd workers’ overall agreement with humans is substantially lower than that of LLMs.
%Conversely, texts on which humans agree tend to elicit stable classifications across LLMs. 
These patterns point to a common source of disagreement: ambiguity in the text or coding rule
rather than errors specific to any one type of annotator. 
Consistent with this interpretation,
clarifying an underspecified codebook substantially reduces disagreement among humans and
sufficiently capable LLMs.

% paragraph 4 (implication)
These findings change how disagreement should be interpreted. When an LLM disagrees with a human coder, the disagreement should not automatically be treated as model error, because human coders are often divided on the same cases. Instead, disagreement can reflect ambiguity in the text or coding rule rather than a failure unique to the LLM. Under the same instructions, sufficiently capable LLMs agree with humans at rates comparable to those among humans themselves, leaving little empirical basis for preferring attentive human coding on annotation quality alone. Moreover, disagreement across LLMs and between LLMs and humans is informative: it tends to concentrate on texts that also divide attentive human coders, providing a useful signal of ambiguity. Validation should therefore ask not only whether an LLM reproduces a human label, but also whether the text and coding rule yield a stable classification.

% paragraph 5 (proposal)
Building on this reinterpretation, we propose a practical workflow for
LLM-based text annotation. Researchers should first develop a clear
codebook and manually apply it to a small random sample to identify
common ambiguities and refine coding rules. They should then verify that
candidate LLMs can reliably apply the codebook and use disagreement across
suitable LLMs to identify additional borderline cases for targeted review
before annotating the full corpus. Because some ambiguity may remain,
we develop ambiguity-aware bounds that use cross-LLM disagreement as a
scalable proxy for residual ambiguity, allowing downstream inference
without assuming a uniquely defined ground-truth annotation for every text.

% contribution
This paper makes two contributions. First, a longstanding annotation
literature argues that disagreement among attentive humans can reveal
ambiguity in texts or coding tasks rather than simply reflecting coding
error \citep{aroyo2015truth, uma2021learning}. We extend this insight to
LLM-based annotation, showing that cross-LLM disagreement provides an
informative, scalable signal of ambiguity rather than model failure.
Second, building on existing human--LLM annotation workflows
\citep{dunivin2025scaling, meng2024exploring, pangakis2025keeping,
xiong2025co, yuan2025case}, we propose using cross-LLM disagreement to select which
texts humans review, so that review effort targets the cases most
likely to reveal ambiguities in the codebook. 
We further develop ambiguity-aware bounds that account for residual
ambiguity in downstream inference without requiring a uniquely defined
target annotation for every text, an assumption common in methods for
bias correction in downstream inference using LLM-generated annotations
(e.g., \citealt{angelopoulos_prediction-powered_2023, carlson2025unifying,
egami2024usingimperfectsurrogatesdownstream, egami2024using, fong_machine_2021,
nakamura2025surrogate, zrnic_cross-prediction-powered_2024}).

\section{Paradigms of Quantitative Text Analysis: Before and After LLMs}\label{sec2}

\subsection{Goal and Philosophy of Text Annotation}\label{sec2.1}

The goal of content analysis is to make ``replicable and valid inferences from texts
(or other meaningful matter) to the contexts of their use'' (\citealt{krippendorff2018content}, p.24).
In this respect, coding text is not different in kind from measurement elsewhere in
social sciences. In the conceptual model of measurement proposed by \citet{adcock2001measurement},
researchers first have a \emph{background concept}, the broad constellation of meanings associated 
with a term such as ``democracy'' or ``security'', and then formulate it into a \emph{systematized
concept}, the specific formulation the researcher adopts for a given study.
The operationalization needs to track the systematized concept (validity) and be reproducible
by different humans (reliability).

Text annotation is a special case of exactly this process, but what makes text unique is its 
unstructured and high-dimensional nature. %Structured data are typically low-dimensional, and their operationalization, however arbitrary, leaves relatively little residual ambiguity once fixed. For example, scholars may define an interstate war as a conflict that produces at least one thousand battle deaths \citep{small1982}. The threshold is a debatable conceptual choice, but once the coding rules are fixed, adjudicating a case against them leaves far less residual ambiguity than reading a text. This is even if, as some note, applying such thresholds is not wholly mechanical, such as for determining start or end times of war \citep{sambanis2004what}.
%Text data, by contrast, are unstructured and high-dimensional. 
By unstructured, we mean that the concept of interest is not directly observed.
Instead, it must be inferred from language.
As \citet{krippendorff2018content} emphasized, texts have no ``reader-independent'' qualities: 
their meaning is not contained in words but is brought to them by readers.
Thus, two careful coders can read the same passage differently, not because one has made
an error, but because text genuinely admits multiple interpretations.
High dimensionality compounds this problem. As a corpus grows, the space of possible textual
configurations grows with it, continually producing borderline and outlier cases \citep{grimmer2022text}. 
The ambiguity that enters at the operationalization stage therefore cannot be fully eliminated 
in many cases, no matter how carefully the codebook is written.

This also points to an inherent tension between reliability and validity.
One could push annotation toward the structured-data ideal by
operationalizing a concept as a mechanical, easily replicated procedure, 
such as counting occurrences of a keyword. Such rules are
trivially replicable, but they typically sacrifice validity to the systematized concept 
\citep{kracauer1952challenge}. Moving in the opposite direction, i.e., toward 
more holistic human judgment, improves validity but
lowers reliability as it leaves more to the reader. 
Any codebook
that is at once usable and valid thus retains some irreducible ambiguity, and
that ambiguity is distributed unevenly across a corpus: some texts are read the
same way by everyone, while others are genuinely contestable.

Taken together, these features imply that text annotation is a demanding task
even for humans, and that there is no ground truth
to be recovered for some cases. 
Indeed, a large body of literature has recognized that inter-coder reliability
among attentive human coders is rarely perfect because of ambiguity, raising
questions about whether a unique ground truth exists
(e.g., \citealt{aroyo2015truth, mikhaylov2012coder, plank2022problem,
pavlick2019inherent, uma2021learning}).

What researchers can do is narrow the range of defensible
readings through operationalization and validate the measurement repeatedly \citep{krippendorff2018content}.

\subsection{Text-as-Data before Large Language Models}\label{sec2.2}
Automated text analysis addresses a problem of scale: corpora such as news articles and party manifestos often exceed what humans can annotate by hand. Automated methods extend systematic content analysis by translating human-defined concepts into reproducible computational procedures.

Before LLMs, automated text analysis relied on dictionary approaches, topic models, and supervised classifiers (see \citealt{grimmer_text_2013} for a review). Yet these methods do not explicitly model the data-generating processes underlying text production or human interpretation \citep{grimmer2021machine}. Accordingly, \citet{grimmer_text_2013} argue that automated text-analysis algorithms “use insightful, but wrong, models of political text to help researchers make inferences from their data” (p.270) and “will never replace careful and close reading of texts” (p.268). Researchers therefore evaluated automated methods primarily by their agreement with human judgments \citep{hase2022automated}, interpreting high agreement as evidence of accuracy and low agreement as algorithmic error or bias.

Researchers also used crowdsourced workers to construct validation sets (e.g., \citealt{budak2016fair, park2021grandstand, plank2022problem, rheault2019politicians, theocharis2016bad, ying2022}). Crowd workers offered a less expensive alternative to experts, but their annotations were generally noisier because they tended to have less training, expertise, and incentive to attend carefully to the task \citep{benoit_crowd-sourced_2016, snow2008cheap}. Nevertheless, their annotations were typically presumed more informative about textual meaning than automated outputs. This human-centered paradigm was reasonable before LLMs, when human annotations were assumed to approximate textual meaning more closely than any algorithm could.

\subsection{Text-as-Data after Large Language Models}\label{sec2.3}

LLMs can now perform tasks that, until recently, were thought to require sophisticated reasoning:
they pass graduate examinations, solve competition-level mathematics
problems, and score competitively on multitask language-understanding benchmarks
\citep{hendrycks2021mmlu, rathje2024gpt}. If these systems can handle such hard tasks, are we still confident that they cannot
perform text annotation as humans do?

Due to its importance, recent years have seen extensive validation of LLM annotation
(e.g., \citealt{gielens2026goodbye, gilardi_chatgpt_2023, goodall2026large,
Halterman2026CodebookLLMs, heseltine2024large, liu2025voices, rathje2024gpt,
tornberg2025large, yang2025data, ziems2024llm, zhu2023can}), but existing
evaluations provide mixed evidence. Several
studies find that LLMs outperform human annotators on tasks such as relevance,
topic, and frame classification \citep{heseltine2024large, tornberg2025large}. Other research 
documents substantial variation across tasks and models \citep{gong2026limits, ziems2024llm}. 
In political science, \citet{Halterman2026CodebookLLMs} show that
LLMs can struggle to follow political-science codebooks that operationalize complex, 
researcher-defined constructs. 
On the other hand, \cite{gilardi_chatgpt_2023} found that early LLMs agreed with ground truth 
generated by human annotations more often than crowd-sourced workers.

A fundamental difficulty with interpreting these results is that the truth is generally
unobserved in reality. Most existing evaluations compare model outputs with a human-coded reference and implicitly treat human labels as ground truth.\footnote{Among the few exceptions is \cite{yang2025data}, which collects
additional annotations from the authors as a robustness check. Our design
differs in two respects. First, their study focuses primarily on how LLM
choice affects downstream estimates rather than sources of annotation
disagreement. Second, their human coders first annotated 20 common texts
and reconciled disagreements before coding the remainder. Although this
calibration may improve human--human agreement, it also gave human coders
information unavailable to LLMs.}
Such comparisons can establish whether an LLM reproduces a particular set of human labels, but
they cannot distinguish among three possibilities: that the LLM is wrong, that the human label is
wrong, or that the text and coding rule admit multiple defensible interpretations. The last
possibility is especially important for text, because even attentive humans may disagree when
language or coding rules are ambiguous. Indeed, a growing literature questions whether human
annotations should automatically be treated as an error-free gold standard
(e.g., \citealt{bisbee_spirling_2026_llm_gold_standard, clark2021all, hosking2024human}).

% pragraph on what we want to assess
%We assess how attentive human coders are observational equivalent to LLMs in terms of text annotation process if we provide the same coding instruction and thus have access to the same information. We evaluate this from the two perspectives: (1) how LLMs and human coders reliably annotate texts and (2) whether the same texts generate the disagreement for both LLMs and human coders.
In this paper, rather than treating human annotations as ground truth, we assess whether LLMs and
attentive human coders are observationally equivalent under a common coding framework (i.e., holding fixed the
explicit operationalization and task information available to them). We evaluate this equivalence
from two perspectives. First, we examine whether LLM--human agreement approaches the agreement observed among humans themselves.
Second, we examine whether texts that generate disagreement among human coders also exhibit greater LLM--human and LLM--LLM disagreement.
If both patterns hold, then disagreement with any single human label is less naturally interpreted
as an error specific to the LLM and may instead reflect difficulty or ambiguity in applying the
common coding rule.

% What it is
% This test provides the indirect test of the quality of LLM annotations, following the tradition of the measurement literature. In social science, the construct of interest is not directly observed, and thus validation is not established by observing a latent ``true'' value directly, but by accumulating evidence supporting the intended interpretation of a measure \citep{messick1995validity}. In the quantitative text analysis, for example, this has been the main method of the validation (e.g., \citealt{quinn_how_2010}). 
These comparisons provide indirect evidence about the quality of LLM annotations without observing the underlying truth. In social science, the construct of interest is typically
latent, so validation cannot generally proceed by comparing a measure with the ground-truth.
Instead, the measurement literature views validity as an accumulation of evidence supporting the
intended interpretation of a measure \citep{ grimmer2022text, lowe2013validating,
messick1995validity, quinn_how_2010}, and human coding provides one such source of evidence as
it has long served as a substantively grounded operationalization of researcher-defined concepts.
Accordingly, if LLM annotations correspond to human annotations to approximately the same degree
that humans correspond to one another, and if both types of coders struggle with the same texts,
this provides validity-relevant evidence that LLMs preserve similar substantive information under
the specified operationalization.

%Yet, methodological practice has not fully adjusted to this change in the relative capabilities 
%of humans and machines, especially in recent work on downstream inference with machine-generated 
%annotations. 
%Social scientists are often interested not in annotations themselves but in quantities estimated 
%using annotated text, such as regression coefficients. To correct bias arising from machine-generated
%annotations, recent methods combine machine annotations observed for the full 
%corpus with human annotations observed for a random validation subset (e.g., \citealt{angelopoulos_prediction-powered_2023,  egami2024usingimperfectsurrogatesdownstream, egami2024using, carlson2025unifying, fong_machine_2021, nakamura2025surrogate, zrnic_cross-prediction-powered_2024}). Although much of this literature emerged
%after the rise of LLMs, it largely inherits the validation logic of the pre-LLM era: the human-coded
%subset is typically treated as the ground truth (or at least as more accurate
%than the machine-generated annotations) and is therefore used to anchor corrections for measurement
%error in LLM annotations.

\subsection{Testable Hypotheses}\label{sec2.4}

Our argument yields four sets of predictions. First, if recent LLMs apply a common coding
rule as reliably as humans, their agreement with humans should approach the agreement
observed among humans. 
%Because both groups receive the same texts and codebooks, this comparison isolates differences in how they apply the codebook.

\medskip

\noindent \textbf{Hypothesis 1 (Reliability).}
\emph{Recent LLMs achieve agreement with human coders comparable to the agreement observed among human coders themselves.}

\medskip

Second, if annotation disagreement reflects ambiguity in the text or codebook, the same
observations should generate disagreement across different annotators. 
%LLMs should disagree more with a held-out human on texts that divide the other humans, and LLM--human disagreement should increase with disagreement across LLMs.

\medskip

\noindent \textbf{Hypothesis 2-a (Human disagreement).}
\emph{LLM--human disagreement is greater for texts on which human coders disagree with one another.}

\medskip

\noindent \textbf{Hypothesis 2-b (LLM disagreement).}
\emph{Disagreement across LLMs is greater for texts on which human coders disagree with one another.}

\medskip

Third, if disagreement arises from ambiguities, clarifying the codebook should improve agreement
across annotators.

\medskip

\noindent \textbf{Hypothesis 3 (Clearer codebook).}
\emph{A revised codebook increases agreement among humans and between LLMs and humans.}

\medskip

Finally, the same relationship between textual ambiguity and disagreement should extend to
crowd-sourced coders. However, because crowd coders have less incentive to pay attention, 
we expect their annotations to correspond less closely with human annotations than LLM annotations do.

\medskip

\noindent \textbf{Hypothesis 4-a (Crowd disagreement).}
\emph{Disagreement among crowd-sourced coders is greater for texts on which humans disagree or LLM outputs vary across models.}

\medskip

\noindent \textbf{Hypothesis 4-b (LLMs versus crowd coders).}
\emph{LLMs achieve greater agreement with human coders than crowd-sourced coders do.}

\medskip

\section{Replication Studies by LLMs and Humans}\label{sec3}

\subsection{Procedure}

\noindent \textbf{Data Collection}: We focus on recent applications of automated text analysis in leading political science journals. We selected 13 articles using text classification from the \textit{American Journal of Political Science}, the \textit{American Political Science Review}, and the \textit{Journal of Politics} from 2018 through 2025. To ensure broad substantive coverage, we included three studies in American politics \citep{feltovich_campaign_2024, gilardi_policy_2021, lacombe2019political}, four in international relations \citep{ClarkDolan2021, mattingly_chinese_2025, Schub2022, thrall_informational_2025}, and six in comparative politics \citep{blaydes_mirrors_2018, BlumenauLauderdale2018, bush_facing_2023, jung_mobilizing_2020, parthasarathy_deliberative_2019, osnabrugge_playing_2021}. We further include \citet{arias2022securitizes} as a theoretically hard case. Recent work suggests that LLMs may struggle to classify concepts rare or absent from their training data \citep{Halterman2026CodebookLLMs}, so findings from general-interest journals may not fully generalize to specialized subfields. \citet{arias2022securitizes} provides a demanding robustness check because it examines \emph{climate securitization}, a construct unlikely to appear frequently in LLM training corpora and difficult to define even within international relations scholarship \citep{BuzanWaeverDeWilde1998, PhillisEtAl2018}.

For each study, we selected one concept and drew a simple random sample of 100 texts from its original corpus. These texts were independently annotated by LLM and human coders. Appendix~\ref{app:study_details} describes concept selection and summarizes the studies.

\medskip

\noindent \textbf{Codebook / Prompt generation}: For each paper, we construct a standardized codebook
template to ensure consistency across papers. The template consists of seven components: 
(1) role assignment, (2) topic definition, (3) paper and data description, (4) category definition,
(5) classification rules, (6) label wording, and (7) step-by-step reasoning instructions. 
We design this template based on the prompt engineering literature for large language models 
(e.g., \citealt{atreja_prompt_2024, Halterman2026CodebookLLMs, mu2024navigating, weber2024evaluation}). 
We then generate the paper-specific components (2)--(6) based on the descriptions in the article with the help of LLMs (GPT-5.5). For each component, we check the generated content against the corresponding descriptions in the published paper and revise it when necessary. This procedure standardizes codebooks across studies   (see Appendix~\ref{app:prompt_all} for
the detailed procedure). While our main
analysis uses the prompt containing all components, we also check how much the results
would change by systematically varying prompts (see Appendix~\ref{app:prompt_sensitivity}).

One limitation of our research design is the codebook quality. Because we are not experts on all 14 studies, our codebooks are based solely on descriptions in the original articles. This reflects a tradeoff between replicating many studies for broader external validity and the limited availability of published codebooks. Appendix~\ref{sec:validation_codebook} therefore replicates the same analysis procedure using three papers with original, more detailed codebooks and confirms that our main findings hold under these codebooks as well. %Consequently, low intercoder reliability in our main analysis should not be interpreted as evidence that the underlying concept is inherently difficult to classify. Some disagreement may instead arise from ambiguities introduced by our reconstructed coding rules. Our analysis of \citet{Schub2022} illustrates this distinction: providing more detailed instructions substantially increases agreement both among humans and between humans and sufficiently capable LLMs.

\medskip

\noindent \textbf{LLM Annotation}: We then annotate each text using 13 different LLMs that span a
wide range of parameter counts, developers, release dates, and likely training data. Seven are
open-weight models: Mistral~7B \citep{jiang2023mistral}, LLaMA~3.1-8B \citep{meta2024llama3},
Qwen2.5-7B and Qwen2.5-14B \citep{qwen2025qwen25}, Gemma~3-12B and Gemma~3-27B \citep{gemmateam2025gemma3},
and GPT-OSS-20B \citep{openai2025gptoss}. The remaining six are proprietary API models:
three from the GPT-5 family (GPT-5.4~mini, GPT-5.4, and GPT-5.5) and three from Anthropic
(Claude Haiku~4.5, Claude Sonnet~4.6, and Claude Opus~4.6).
Table~\ref{tab:models} lists each model with its release date.
\medskip

\begin{table}[t]
\centering
\small
\caption{Language models used in the replication, with release dates.}
\label{tab:models}
\begin{tabular}{llc}
\toprule
Model & Developer & Release date \\
\midrule
Mistral 7B (Instruct v0.3) & Mistral AI & May 2024 \\
LLaMA 3.1-8B               & Meta       & July 2024 \\
Qwen2.5-7B                  & Alibaba    & September 2024 \\
Qwen2.5-14B                 & Alibaba    & September 2024 \\
Gemma 3-12B                 & Google     & March 2025 \\
Gemma 3-27B                 & Google     & March 2025 \\
GPT-OSS-20B                 & OpenAI     & August 2025 \\
Claude Haiku 4.5            & Anthropic  & October 2025 \\
Claude Opus 4.6             & Anthropic  & February 2026 \\
Claude Sonnet 4.6           & Anthropic  & February 2026 \\
GPT-5.4 mini                & OpenAI     & March 2026 \\
GPT-5.4                     & OpenAI     & March 2026 \\
GPT-5.5                     & OpenAI     & April 2026 \\
\bottomrule
\end{tabular}
\end{table}

\noindent \textbf{Human Annotations}: Finally, the three authors of this paper, who are Ph.D. students with training in political science,
independently annotated the same texts using the same codebooks as the LLMs,
without discussing difficult cases or consulting external information.
We avoided calibration and reconciliation discussions because
these could introduce task-specific information or shared interpretations
not explicitly documented in the codebook and thus unavailable to LLMs.
Requiring both human and LLM annotators to follow the same explicit coding
instructions allows us to directly compare LLM--human and human--human
agreement under a common coding framework.

We assessed human intercoder reliability using Krippendorff's $\alpha$
\citep{krippendorff2018content}, which measures agreement among multiple
coders while adjusting for chance agreement. Formally,
\begin{align*}
    \alpha := 1 - \frac{\text{Observed within-text disagreement}}{\text{Baseline disagreement from the label distribution}}.
\end{align*}
Thus, $\alpha=1$ indicates perfect reliability and $\alpha=0$ indicates
chance-level agreement. $\alpha$ can be negative when observed disagreement
exceeds that expected by chance.

{
Table~\ref{tab:studies_icr} reports Krippendorff's $\alpha$ and mean pairwise agreement for each
study. Consistent with the literature
(e.g., \citealt{gong2026limits}), human intercoder reliability varies
substantially across studies, with some estimates falling below
conventional thresholds for reliable content analysis. We note that low
$\alpha$ partly reflects class imbalance, which inflates chance agreement:
in four studies where fewer than 14\% of labels are positive \citep{arias2022securitizes, mattingly_chinese_2025, parthasarathy_deliberative_2019, thrall_informational_2025}, raw
agreement exceeds 85\% even though $\alpha$ is below 0.45. With only 100 texts, changing
a single label can shift $\alpha$ dramatically for such cases.
Our procedure, however,
may also contribute. Unlike recommended practice
\citep{lombard2002content, neuendorf2002content}, our coders received no
training, pilot coding, or reconciliation, so our benchmark likely
understates what a trained coding team would achieve. Our comparison
therefore characterizes how LLMs and humans behave under a common coding
framework, not how LLMs compare with trained expert coders.

To assess whether our coders nonetheless provide a reasonable
benchmark under this framework, we conduct two validation exercises in Appendix~\ref{sec:validation_coding}.
First, for three replicated studies which have original human
annotations targeting the same binary concept
\citep{feltovich_campaign_2024, lacombe2019political, Schub2022},
we compare our annotations with those of the original annotators.
Second, we obtain independent annotations from scholars with
relevant substantive expertise for subsets of texts from
\citet{arias2022securitizes} and \citet{ClarkDolan2021}.
Across these exercises, we find no systematic evidence that
disagreement with the original annotators or subject-matter
specialists exceeds disagreement among our three human coders.
These results support the use of our human annotations as a
reasonable benchmark for evaluating LLM annotations.

}

\begin{table}[t]
\centering
\small
\caption{The 14 replicated studies: concept of interest and human intercoder reliability (ICR), measured by Krippendorff's $\alpha$ and mean pairwise agreement on each study's full 100-text sample. The studies are ordered in descending order of Krippendorff's $\alpha$.}
\label{tab:studies_icr}
\sbox0{\begin{tabular}{llcc}
\toprule
Paper & Concept of interest & Krippendorff's $\alpha$ & Mean pairwise agreement \\
\midrule
\quad \cite{feltovich_campaign_2024} & Negative campaigning & 0.869 & 0.987 \\
\quad \cite{bush_facing_2023} & Taxes & 0.745 & 0.933 \\
\quad \cite{ClarkDolan2021} & Fiscal policy & 0.734 & 0.880 \\
\quad \cite{BlumenauLauderdale2018} & Finance & 0.639 & 0.860 \\
\quad \cite{jung_mobilizing_2020} & Moral rhetoric & 0.460 & 0.847 \\
\quad \cite{arias2022securitizes} & Climate security & 0.431 & 0.873 \\
\quad \cite{lacombe2019political} & Gun control & 0.411 & 0.720 \\
\quad \cite{mattingly_chinese_2025} & Western dysfunction & 0.392 & 0.887 \\
\quad \cite{parthasarathy_deliberative_2019} & Fund allocation & 0.382 & 0.853 \\
\quad \cite{thrall_informational_2025} & Commercial issues & 0.372 & 0.900 \\
\quad \cite{blaydes_mirrors_2018} & Art of rulership & 0.363 & 0.753 \\
\quad \cite{Schub2022} & Adversary domestic politics & 0.266 & 0.640 \\
\quad \cite{gilardi_policy_2021} & Smoking ban enforcement & 0.224 & 0.700 \\
\quad \cite{osnabrugge_playing_2021} & Emotive rhetoric & -0.022 & 0.533 \\
\bottomrule
\end{tabular}}
\usebox0
\par\smallskip
\begin{minipage}{\wd0}
\end{minipage}
\end{table}

\subsection{Results}

\subsubsection{Observational Equivalence of LLM and Human Annotation}

We begin by testing Hypothesis~1 by comparing LLM--human agreement
with human--human agreement. Because human agreement varies substantially across studies, we present the results separately for higher-reliability tasks (Krippendorff's $\alpha>0.4$) and lower-reliability tasks (Krippendorff's $\alpha< 0.4$), yielding an even split of seven studies in each group.

\begin{figure}[h]
    \centering
    \includegraphics[width=\linewidth]{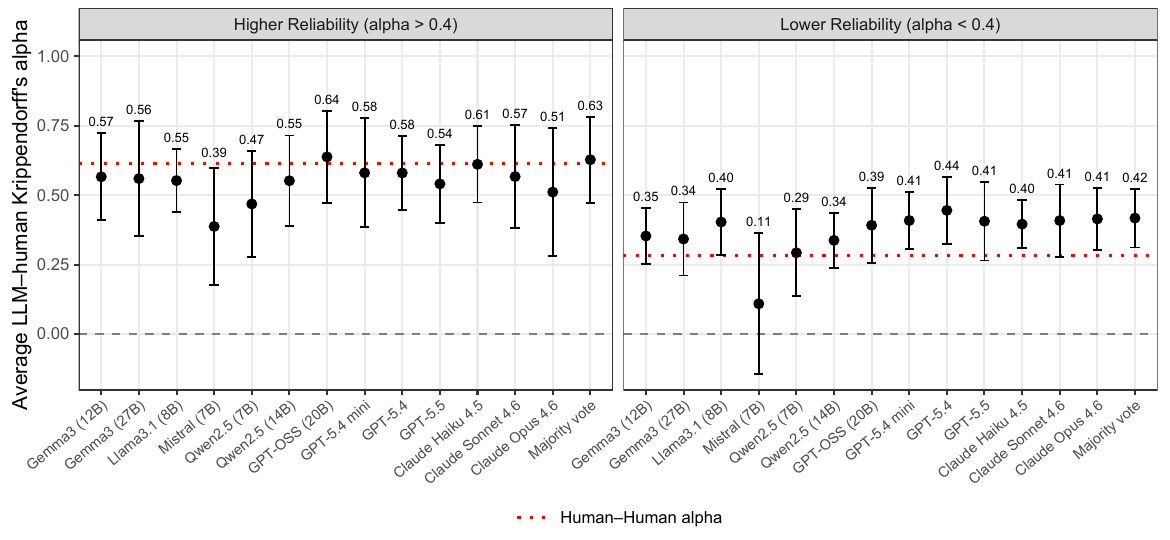}
    \caption{Average LLM--human agreement by model, shown separately for studies with relatively higher human intercoder reliability (ICR $>0.4$; 7 studies) and relatively lower human intercoder reliability (ICR $\leq 0.4$; 7 studies). Points represent model-specific averages across studies, with error bars indicating 95\% between-study confidence intervals. The red dotted line indicates average human--human ICR within each panel, while the gray dashed line indicates $\alpha=0$.}
    \label{fig:llm_expert_agreement}
\end{figure}

Figure~\ref{fig:llm_expert_agreement} shows that LLM--human agreement closely
tracks human--human agreement. 
Among the tasks with relatively higher human reliability, the average intercoder reliability among humans is 0.61, while all models except Mistral-7B achieve LLM--human agreement
between 0.47 and 0.64. Among the
lower-reliability tasks, LLM--human agreement for these models declines to between 0.29 and 0.44, but
human--human agreement also falls to 0.28. Thus, lower LLM--human agreement
occurs precisely in settings where humans themselves agree less.
 Overall, the results support Hypothesis~1:
with the exception of the oldest model (Mistral-7B), recent LLMs achieve agreement with
humans comparable to the agreement observed among humans themselves.

\subsubsection{Disagreement Concentrates in Texts That Divide Humans}\label{sec:disagreement_gini}

To test Hypothesis~2-a, we use a leave-one-human-out procedure. For each text,
we classify it as \emph{clear} when two humans agree and as \emph{ambiguous}
when they disagree. We then evaluate each LLM against the remaining held-out
human, repeating the procedure for all three choices of held-out human and
averaging the results.

\begin{figure}[h]
    \centering
    \includegraphics[width=\linewidth]{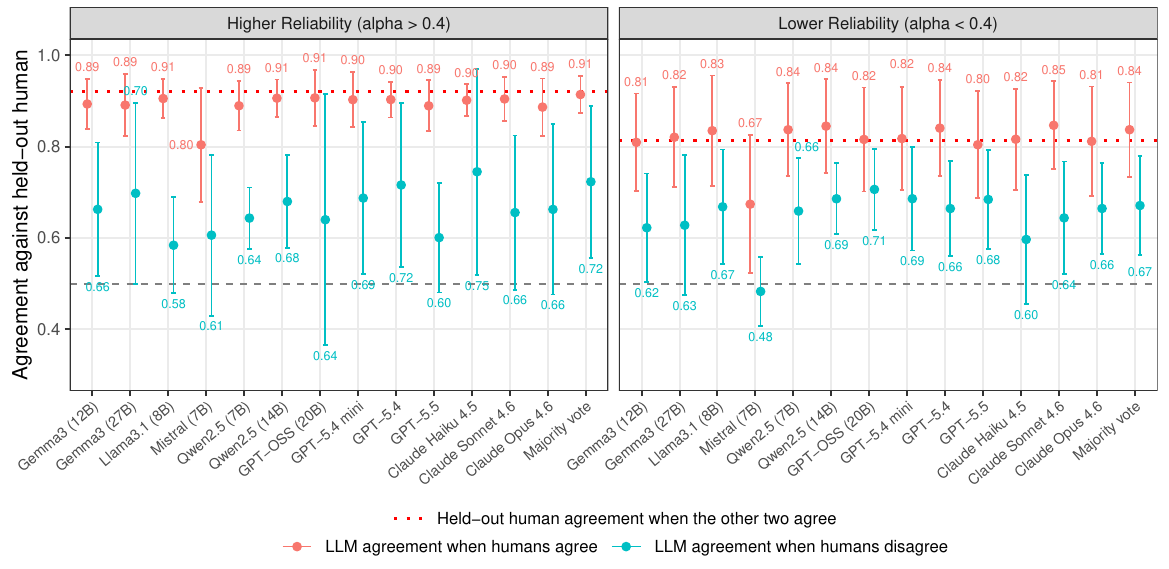}
    \caption{Agreement with a held-out human, by model and by whether the two
    remaining humans agree. The red dotted line indicates the held-out
    human's agreement with the common label of the other two humans on clear
    texts. Red and blue points report LLM agreement with the held-out human
    on clear and ambiguous texts, respectively. Error bars are 95\%
    between-study confidence intervals.}
    \label{fig:heldout_accuracy}
\end{figure}

Figure~\ref{fig:heldout_accuracy} shows that LLM agreement declines sharply on
texts that divide humans. Among the studies with higher reliability, agreement with the
held-out human, averaged across the 13 models, falls from approximately 0.89 on clear texts to 0.66 on
ambiguous texts. Among the studies with lower reliability, it falls from approximately 0.81
to 0.65.
These results support Hypothesis~2-a. LLM--human disagreement is concentrated
in the same texts that generate disagreement among humans, implying that
much of the disagreement reflects ambiguity in the text or codebook rather
than model-specific error.
Appendix~\ref{app:example_ambiguous} also provides
illustrative examples of these contested texts.

We then test Hypothesis~2-b by examining whether disagreement across LLMs is
greater for texts that divide humans. For each text $i$, let $p_i$ denote the
proportion of the 13 LLMs assigning the label 1. We measure cross-model
disagreement using Gini impurity,
\begin{align}
    \mathrm{Gini}_i = 2p_i(1-p_i),
\end{align}
which ranges from zero when all models agree to 0.5 when they are evenly
divided.

\begin{table}[!t]
\centering
\caption{Cross-model LLM disagreement (Gini impurity) by human agreement}
\label{tab:gini_disagreement}
\sbox0{\setlength{\tabcolsep}{4pt}\begin{tabular}{lccccc}
\toprule
 & \multicolumn{2}{c}{Mean LLM disagreement} & & & \\\cmidrule(lr){2-3}
Difficulty & Humans unanimous & Humans split & Diff. & 95\% CI & $N$ \\
\midrule
Higher Reliability ($\alpha > 0.4$) & 0.075 & 0.250 & 0.174 & [0.120, 0.229] & 565 / 135 \\
Lower Reliability ($\alpha < 0.4$) & 0.085 & 0.239 & 0.154 & [0.091, 0.217] & 440 / 260 \\
\bottomrule
\end{tabular}}
\usebox0
\par\smallskip
%\begin{minipage}{\wd0}
%\footnotesize\emph{Notes:} Cells are mean Gini impurity $2p(1-p)$ over the 13 LLM votes per text. $N$ counts texts (unanimous\,/\,split); CIs are cluster-robust by paper.
%\end{minipage}
\end{table}

Table~\ref{tab:gini_disagreement} shows that cross-model disagreement is much
higher for texts on which humans disagree. In the studies with higher reliability, mean Gini
impurity rises from 0.075 to 0.250; in the studies with lower reliability, it rises from
0.085 to 0.239. Both differences are statistically significant. These results 
support Hypothesis~2-b and are robust to different measures 
of heterogeneity (see Appendix~\ref{app:entropy_expert}). 
In Appendix~\ref{app:prompt_sensitivity}, we obtain similar results by varying the prompts: 
for most models, classifications are significantly more sensitive to small changes in prompt 
for texts on which humans disagree.
This pattern is analogous to disagreement among attentive human coders, which prior work
has argued can reflect genuine ambiguity in the annotation task rather than simple coding error
\citep{aroyo2015truth, uma2021learning}.%\footnote{In Appendix~\ref{sec:predict_disagreement}, we further explore what explains the prevalence of disagreements.}.

\subsubsection{Clarifying the Codebook Improves Agreement}\label{sec3.2.3}

Finally, we test Hypothesis~3 using the classification task from
\citet{Schub2022}. Specifically, we developed a revised codebook that goes beyond our standardized template by providing clearer definitions, more explicit inclusion and exclusion rules, and guidance for mixed cases.
The three humans and 13 LLMs then reclassified the same texts using the revised instructions.

\begin{figure}[h]
    \centering
    \includegraphics[width=0.75\linewidth]{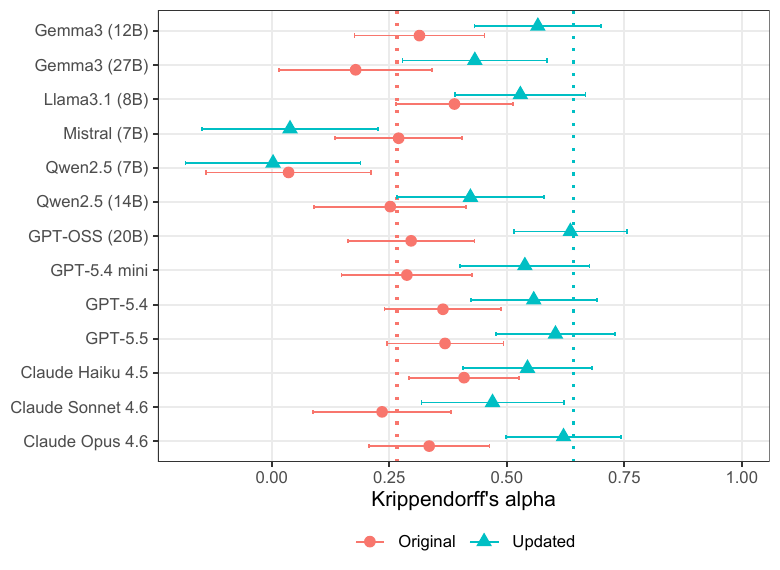}
    \caption{LLM--human agreement under the original (red) and revised codebook (blue) for the \citet{Schub2022} annotation task. Each point reports a model's average Krippendorff's $\alpha$ with the three humans. The colored dotted lines indicate human--human agreement measured by Krippendorff's $\alpha$.}
    \label{fig:schub_codebook}
\end{figure}

Figure~\ref{fig:schub_codebook} shows that the revised codebook raises
human--human agreement from approximately 0.27 to 0.64. It also increases
LLM--human agreement for most recent models, several of which reach between
0.54 and 0.64 and approach the human benchmark. The improvement is not uniform,
however. Mistral-7B and Qwen2.5-7B perform
worse under the more detailed instructions, suggesting that a clearer but more
complex codebook may be difficult for weaker models to apply consistently. 
Together, these findings show that improving the codebook can increase
agreement among humans and between humans and recent capable LLMs.
%Switching annotators alone need not resolve disagreement arising from unclear instructions.

%Taken together, these findings show that, under the same coding instructions, recent LLMs show agreement with humans comparable to that among humans themselves, and texts that divide humans also tend to divide LLMs. This does not mean that either produces sufficiently reliable annotations when the instructions are unclear. Rather, clarifying the codebook improves agreement among humans and between humans and most recent LLMs. Improving the coding instructions is therefore important for both types of annotators; simply switching from LLMs to attentive humans need not resolve the problem.

\subsubsection{Same finding for Crowdsourced Workers, but they are less reliable than LLMs}\label{sec4}

Finally, we extend our analysis to crowdsourced annotation.
We recruited 165 U.S. participants through Cloud Research Connect and randomly
assigned each participant to one of the 14 papers we used in replication or to the revised
codebook version of \citet{Schub2022} analyzed in the previous section. For each paper, we
limited the coding task to 20 texts to reduce respondent fatigue while retaining enough 
observations for analysis. Specifically, we selected the 10 texts with the highest agreement
across LLMs and the 10 texts with the lowest agreement to sample both easy and difficult cases.
Participants were asked to apply the same codebook used in the previous section to annotate 
the assigned texts. To reduce potential LLM-assisted responses, we disabled copying and pasting and implemented multiple bot-detection and data-quality checks. We detail the annotation procedure in Appendix~\ref{app:crowd_procedure} and preregistered hypotheses, results, and robustness checks in Appendix~\ref{app:crowd_results}.

\begin{figure}[!ht]
    \centering
    \includegraphics[width=\linewidth]{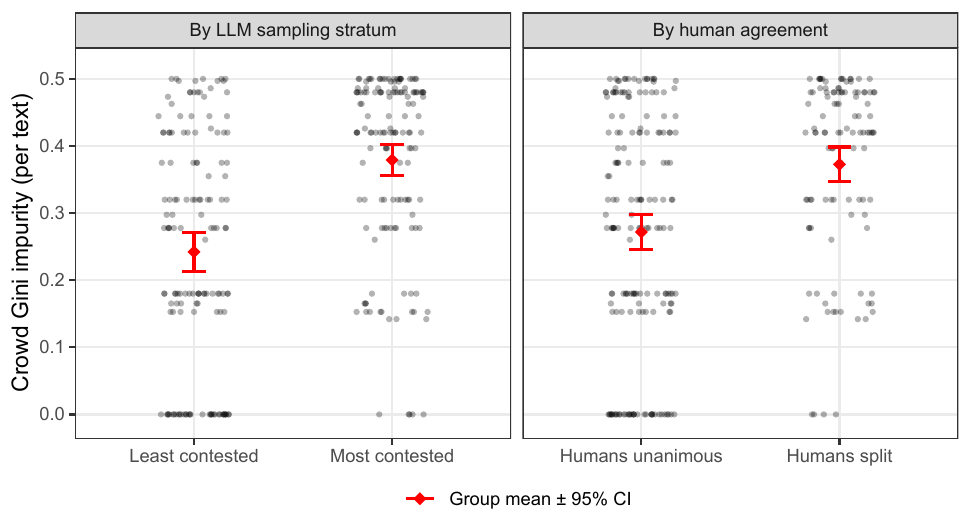}
    \caption{(Left) For each paper, texts are grouped as the 10
    least contested texts across LLM coding versus 10 most contested ones, based on our stratified sampling. (Right) Texts
    are grouped into texts on which the three human coders are unanimous or split. In red are the group means with 95\% confidence intervals; gray points are individual texts. We exclude the revised codebook version of \cite{Schub2022} for consistency.  }
    \label{fig:h1_replication}
\end{figure}

We obtain two main findings. First, we find that crowd workers exhibit greater disagreement
on ambiguous texts, regardless of whether ambiguity is measured by disagreement among LLMs or
among humans. The difference in Gini impurity between ambiguous and non-ambiguous texts is both
statistically and substantively significant: Gini impurity is 14 percentage points higher for the
texts with the lowest versus highest LLM agreement and 10 percentage points higher for texts on which
humans split versus those on which they reached unanimous agreement (Figure~\ref{fig:h1_replication}).

\begin{figure}[!ht]
    \centering
    \includegraphics[width=\linewidth]{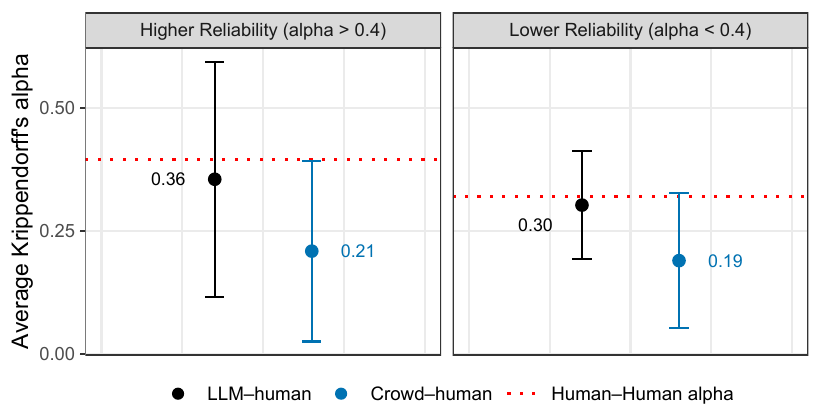}
    \caption{Survey extension of main analyses. All agreement measures use Krippendorff's $\alpha$. The red dotted line indicates human--human agreement; black points and bars show LLM--human agreement, and blue points and bars show crowd--human agreement, with 95\% between-study confidence intervals.}
    \label{fig:h2_replication}
\end{figure}

Second, we find that human--human and LLM--human agreement both
decline for lower-reliability tasks, whereas crowd--human agreement
remains consistently lower (Figure~\ref{fig:h2_replication}). This extends \citet{gilardi_chatgpt_2023}
by comparing intercoder agreement without presuming human labels
are ground truth. Most strikingly, LLM--human agreement closely tracks agreement among
humans across both types of tasks, which is not the case for crowd-sourced workers.

%Second, we find that agreement between humans and LLMs is substantially higher than agreement between humans and crowd workers. To facilitate this comparison, we replot our main analysis from Figure~\ref{fig:llm_expert_agreement} using only the same 20 texts annotated by all three groups. This standardized comparison, based on identical texts and codebooks, replicates and extends  the findings of \cite{gilardi_chatgpt_2023} by comparing inter-coder agreement rather than  evaluating performance against an assumed ground truth. We find not only a substantial gap between human--LLM and human--crowd agreement, but also that the attainable level of agreement varies systematically with task difficulty. As shown in Figure~\ref{fig:h2_replication},  human--human agreement (red dotted line) declines considerably for lower-reliability tasks (right panel) relative to higher-reliability tasks (left panel). Most strikingly, LLM--human agreement closely tracks agreement among humans across both types of tasks, whereas crowd--human agreement remains consistently lower.

\section{Practical Takeaways}\label{sec5}

Given these results, we propose a procedure that uses multiple LLMs to efficiently identify borderline cases, 
refine the codebook, and ultimately conduct honest statistical inference. Figure~\ref{fig:annotation-pipeline} summarizes the procedure. 
Below, we explain each step in detail.

\begin{figure}[!hp]
\centering

\begin{tikzpicture}[
  font=\sffamily\small,
  line width=0.5pt,
  >=stealth,
  box/.style={
    draw=black,
    rounded corners=2pt,
    fill=white,
    align=left,
    inner sep=7pt,
    text width=110mm
  },
  smallbox/.style={
    draw=black,
    rounded corners=2pt,
    fill=white,
    align=left,
    inner sep=7pt
  },
  cond/.style={
    draw=black,
    rounded corners=2pt,
    fill=white,
    align=center,
    inner sep=6pt
  },
  sub/.style={
    draw=black!55,
    rounded corners=2pt,
    fill=white,
    align=center,
    inner sep=5pt,
    font=\sffamily\footnotesize
  },
  arr/.style={
    ->,
    line width=0.5pt
  }
]

% -------------------------------------------------
% STEP 1
% -------------------------------------------------
\node[box] (step1)
{
  \textbf{Step 1: Generate a clear Codebook}
  \bul{
    \item Construct a detailed codebook to classify the topic of interest
    \item Include clear classification rules %Suggested components: definition, classification rule, label wording, background context \emph{[For LLMs: add role, step-by-step reasoning]}
  }
};

% -------------------------------------------------
% STEP 2
% This is now an ordinary node with exactly the
% same style and width as Step 1.
% -------------------------------------------------
\node[box, below=7mm of step1] (step2)
{
\begin{minipage}{110mm}
\textbf{Step 2: Test Codebook with LLMs / Humans}

%\vspace{3mm}

%\centering
%\begin{tikzpicture}[
%  sub/.style={
%    draw=black!55,
%    rounded corners=2pt,
%    fill=white,
%    align=center,
%    inner sep=5pt,
%    font=\sffamily\footnotesize
%  }
%]
%\node[sub, text width=47mm] (s2a)
%{
%  \textbf{2A: LLM Annotation}\\[2pt]
%  Agreement across multiple models or prompts
%};

%\node[sub, text width=47mm, right=5mm of s2a] (s2b)
%{
%  \textbf{2B: Expert Annotation}\\[2pt]
%  Researchers annotate a subset of texts
%};
%\end{tikzpicture}

%\raggedright
%\vspace{2mm}

\bul{
  \item Randomly select a subset of the corpus; annotate with LLMs/humans.
  \item \emph{For LLMs:} use multiple models, including large open models and proprietary models.
  \item Check how much humans/LLMs agree on the random subset
}
\end{minipage}
};

% Exactly vertical arrow
\draw[arr] (step1.south) -- (step2.north);

% -------------------------------------------------
% AGREEMENT BRANCH
% -------------------------------------------------
\node[
  cond,
  text width=36mm,
  below=13mm of step2,
  xshift=-22mm
] (ifhigh)
{
  \textbf{High Disagreement}
};

\node[
  cond,
  text width=36mm,
  below=13mm of step2,
  xshift=22mm
] (iflow)
{
  \textbf{Low Disagreement}
};

\coordinate (fork) at ([yshift=-6mm]step2.south);

\draw (step2.south) -- (fork);
\draw[arr] (fork) -- (ifhigh.north);
\draw[arr] (fork) -- (iflow.north);

% -------------------------------------------------
% STEP 3
% -------------------------------------------------
\node[
  smallbox,
  text width=66mm,
  below=9mm of ifhigh
] (step3)
{
  \textbf{Inspect Disagreements}
  \bul{
    \item Check where LLMs and humans disagree
    \item \textbf{Improve the codebook:} reduce ambiguity in the
          classification rules; repeat Step 1
  }
};

% -------------------------------------------------
% STEP 4
% -------------------------------------------------
\node[
  smallbox,
  text width=110mm,
  below=11mm of step3,
  xshift=22mm
] (step4)
{
  \textbf{Step 3: Run Downstream Inference under Residual Ambiguity}
  \bul{
    \item Run LLM annotations for the entire corpus
    \item Report LLM--LLM, human--human, and/or LLM--human
          agreement calculated on the random subset of the corpus
    \item Report the codebook used
    \item \textbf{Use ambiguity-aware bounds} for honest inference
  }
};

% -------------------------------------------------
% ARROWS
% -------------------------------------------------
\draw[arr] (ifhigh.south) -- (step3.north);

\draw[arr]
  (iflow.south)
  --
  (iflow.south |- step4.north);

% Return arrow from Step 3 to Step 1
\coordinate (returnleft) at ([xshift=-10mm]step3.west);

\draw[arr]
  (step3.west)
  --
  (returnleft)
  |-
  (step1.west);

\end{tikzpicture}

\caption{Proposed analysis procedure for text-as-data using LLMs.}
\label{fig:annotation-pipeline}
\end{figure}

\subsection{STEP 1: Generate a clear codebook}\label{step1}
The first step is to construct a codebook. Researchers should first
define what they want to measure and how to operationalize it. This step
is consequential, as our results show that much apparent annotation
error, whether by LLMs or humans, is ambiguity introduced during
operationalization.
While the best codebook depends on the setting and concept of interest,
clear classification rules generally improve LLM agreement with human
annotations (see Appendix~\ref{app:prompt_sensitivity}). Researchers should
state the classification rules clearly and revise them through trial and
error in STEP~2.

\subsection{STEP 2: Test codebook with LLMs / humans and inspect disagreements}\label{step2}
Before annotating the full corpus, researchers should manually code a
small simple random sample to assess how often the codebook yields clear
classifications and identify common weaknesses. Drawing on their
substantive understanding, they should review disagreements and difficult
cases and revise the codebook. This step evaluates the codebook rather
than the annotator and is necessary regardless of who performs the full
annotation.

Next, researchers should apply the revised codebook to multiple LLMs
across additional subsets for further pilots. They should expand human
review by targeting texts on which LLMs disagree, rather than additional
random texts, because these cases are more likely to reveal remaining
codebook ambiguity.\footnote{Appendix~\ref{app:stochastic_decoding} shows
that repeated runs of a single model can also signal ambiguity, although
their informativeness varies across models.}
Researchers should review these cases and refine the codebook until
clearer instructions no longer resolve disagreements, following iterative
human-review approaches in the literature
\citep{dunivin2025scaling, pangakis2025keeping, xiong2025co}.

Because model capabilities evolve rapidly, we recommend pilot-based
selection rather than fixed models or prompts. Ideally, multiple humans
should annotate a small subset. Exclude models whose LLM--human agreement
falls substantially below human--human agreement, since their disagreement
may reflect limited capability or poor instruction following rather than
ambiguity. In our studies, recent medium- and large-scale models perform
similarly and open-source models perform
comparably to proprietary state-of-the-art models, consistent with the
literature (e.g., \citealt{yang2025data}). Once the codebook is clear, we find that prompting technique, such as role
assignment, contribute relatively little
to agreement compared with explicit classification rules
(Appendix~\ref{app:prompt_sensitivity}).

\subsection{STEP 3: Account for residual ambiguity}\label{step3}

Once the codebook has stabilized, researchers can annotate the full corpus. Nevertheless, some texts may remain
ambiguous even under well-developed codebooks \citep{aroyo2015truth}, as our
replications illustrate. Forcing these texts into a category (for example,
coding all unclear cases as zero) may increase agreement while compromising
validity.
Residual ambiguity may therefore distort estimates and inference in
downstream analyses, such as regressions.

We propose \emph{ambiguity-aware bounds} to accommodate annotations that
are undefined under the codebook. For the sake of explanation,
consider the average of binary annotations
across $N$ texts,
\begin{align}
    \theta := \frac{1}{N}\sum_{i = 1}^N Y_i,
\end{align}
where $Y_i$ is the true annotation of text $i$. The problem here is that,
if we cannot define the truth for some texts, the quantity above is also undefined.
Nevertheless, knowing which texts are
ambiguous allows us to bound the possible averages without selecting a
unique label for each text. Let $A_i=1$ for ambiguous texts and $A_i=0$
otherwise. Since annotations are binary,
\begin{align}
    \theta \in \biggl[ \frac{1}{N}\sum_{i = 1}^N (1 - A_i) \hat Y_i, \ \ \frac{1}{N}\sum_{i = 1}^N \biggl((1 - A_i) \hat Y_i + A_i\biggr) \biggr], \label{bound1}
\end{align}
where $\hat Y_i$ is the observed human or LLM annotation. We assume
$\hat Y_i=Y_i$ for unambiguous texts.\footnote{We  
relax this assumption in Appendix~\ref{app:sensitivity}.} The lower and upper bounds assign all ambiguous texts zero
and one, respectively.

Researchers should assess ambiguity relative to their intended
operationalization, providing an independent substantive judgment of
whether multiple classifications are defensible. Importantly, errors have asymmetric
consequences: classifying unambiguous texts as ambiguous widens the bounds
conservatively, whereas missing ambiguous texts may make them too narrow.
We therefore recommend coding $A_i$ conservatively, treating a text as
ambiguous whenever a substantively defensible alternative classification
cannot be ruled out.

To estimate these bounds without reviewing every text, we use cross-LLM
disagreement, measured by entropy or Gini impurity, as an informative but
imperfect proxy for ambiguity. After finalizing the codebook, researchers
assess $A_i$ in a random subset, with $S_i=1$ indicating inclusion and
$S_i=0$ otherwise. They then annotate the full corpus with multiple LLMs
to construct $\hat A_i$ from their disagreement and calculate
\begin{align}
    \theta \in \biggl[ \frac{1}{N}\sum_{i = 1}^N (1 - \tilde A_i) \hat Y_i, \ \ \frac{1}{N}\sum_{i = 1}^N \biggl((1 - \tilde A_i) \hat Y_i + \tilde A_i \biggr) \biggr] \ \ \text{where} \ \ \tilde A_i = \hat A_i + \frac{S_i}{\Pr(S_i = 1)}\biggl(A_i - \hat{A}_i\biggr). \label{bound2}
\end{align}
Here, $\tilde A_i$ is the design-adjusted ambiguity measure
\citep{angelopoulos_prediction-powered_2023, egami2024using}, and the endpoints
are unbiased estimates of those in Equation~\eqref{bound1}.
%\footnote{To see why, first notice that
%\begin{align*}
%    \E[\tilde A_i \mid A_i, \hat A_i, \hat Y_i] =  \hat A_i + \frac{\Pr(S_i = 1)}{\Pr(S_i = 1)}\biggl(A_i - \hat A_i\biggr) = A_i,
%\end{align*}
%where the first equality is because the subset researchers read is chosen randomly.
%As a result,
%\begin{align*}
%    \E\biggl[ \frac{1}{N}\sum_{i = 1}^N (1 - \tilde A_i) \hat Y_i\biggr]
%    &= \frac{1}{N}\sum_{i = 1}^N \E[(1 - \tilde A_i) \hat Y_i ] = \frac{1}{N}\sum_{i = 1}^N \E\biggl[ \biggl(1 - \E[\tilde A_i \mid A_i, \hat A_i, \hat Y_i ]\biggr) \hat Y_i \biggr] = \E\biggl[ \frac{1}{N}\sum_{i = 1}^N (1 - A_i) \hat Y_i\biggr]
%\end{align*}
%where the second equality follows from the law of iterated expectations. The same argument applies to the upper bound. Importantly, the argument does not assume that the proxy $\hat A_i$ is binary.}
Appendix~\ref{app:sensitivity} presents extensions beyond averages and an
empirical application. Compared with bounds and sensitivity analyses for
non-random measurement error
\citep{imai_causal_2010, bisbee_spirling_2026_llm_gold_standard}, our approach
gains informativeness from text-level variation in ambiguity and its
empirical proxy. When ambiguity is widespread, the bounds may be uninformative; they complement rather than replace codebook development and validation.

\section{Conclusion}\label{sec6}
Using replications of 14 political science papers, we evaluate LLM-based
annotation by comparing annotations from 13 LLMs, three human coders
(Ph.D. students with training in political science), and 165 crowd-sourced workers applying identical codebooks.
We find that recent LLMs agree with humans at rates comparable to those among humans
themselves, establishing an observational equivalence between LLM and
human coding. We further show that this equivalence arises because
different annotators struggle with the same texts: LLM--human disagreement
concentrates in texts that divide humans, and cross-LLM disagreement is
substantially greater on those same cases. Clarifying an underspecified
codebook reduces disagreement among humans and sufficiently capable LLMs.
Crowd-sourced workers exhibit the same ambiguity-related pattern, although
their overall agreement with humans is substantially lower than that of
recent LLMs.

Because humans and LLMs often disagree on the same borderline cases, divergence from human labels need not indicate model error. Researchers
should therefore prioritize conceptualization and codebook development,
using cross-LLM disagreement to identify difficult cases and refine coding
rules. Because some ambiguity may remain, we develop ambiguity-aware
bounds that do not require a unique ground-truth label for every text.
Reliable text analysis requires capable annotators, careful
operationalization, and statistical inference that explicitly propagates
residual ambiguity rather than treating it as ordinary annotation error.

\ifdefined\PASubmission
\section*{Funding Statement}
\PaperFinancialSupport

\section*{AI Statement}
Large language models are used as research tools in this study, 
as described in the main text and Appendix. In particular, OpenAI's 
GPT-5.5, accessed through ChatGPT in 2026, was used to generate initial 
paper-specific components of the standardized classification codebooks 
from source articles. The authors manually reviewed and, where necessary, 
revised these outputs against the original articles. The study also uses 
multiple large language models as text annotators. The models, versions, 
prompts, and annotation procedures are reported in the manuscript and Appendix.

The authors also used OpenAI's ChatGPT during manuscript preparation to assist 
with language editing, organization, and the drafting and revision of text. 
OpenAI's Codex and Anthropic's Claude Code were used to assist with writing, 
debugging, and revising research code. All AI-assisted text and code were manually 
reviewed and modified by the authors, who take full responsibility for the manuscript, 
analyses, and accompanying code.

\section*{Data Availability Statement}
Replication data and code will be posted upon acceptance of the paper.

\section*{Acknowledgements}
\PaperAcknowledgements
\fi

\section*{Research with Human Subjects}

The crowd-sourced annotation study was reviewed and approved by the Harvard
University Institutional Review Board (Protocol No.~26-0698). All
participants provided informed consent prior to participation. The study was
preregistered on the Open Science Framework (\url{https://osf.io/5wp2k}).

% Bibliography should come before \appendix
\clearpage
\bibliography{references,ref_add}

\begin{thebibliography}{}

\bibitem[Adcock and Collier, 2001]{adcock2001measurement}
Adcock, R. and Collier, D. (2001).
\newblock Measurement validity: A shared standard for qualitative and quantitative research.
\newblock {\em American Political Science Review}, 95(3):529--546.

\bibitem[Angelopoulos et~al., 2023]{angelopoulos_prediction-powered_2023}
Angelopoulos, A.~N., Bates, S., Fannjiang, C., Jordan, M.~I., and Zrnic, T. (2023).
\newblock Prediction-{Powered} {Inference}.
\newblock arXiv:2301.09633 [stat].

\bibitem[Arias, 2022]{arias2022securitizes}
Arias, S.~B. (2022).
\newblock Who securitizes? climate change discourse in the united nations.
\newblock {\em International Studies Quarterly}, 66(2):sqac020.

\bibitem[Aroyo and Welty, 2015]{aroyo2015truth}
Aroyo, L. and Welty, C. (2015).
\newblock Truth is a lie: Crowd truth and the seven myths of human annotation.
\newblock {\em AI magazine}, 36(1):15--24.

\bibitem[Atreja et~al., 2024]{atreja_prompt_2024}
Atreja, S., Ashkinaze, J., Li, L., Mendelsohn, J., and Hemphill, L. (2024).
\newblock Prompt {Design} {Matters} for {Computational} {Social} {Science} {Tasks} but in {Unpredictable} {Ways}.
\newblock arXiv:2406.11980 [cs].

\bibitem[Benoit et~al., 2016]{benoit_crowd-sourced_2016}
Benoit, K., Conway, D., Lauderdale, B.~E., Laver, M., and Mikhaylov, S. (2016).
\newblock Crowd-sourced {Text} {Analysis}: {Reproducible} and {Agile} {Production} of {Political} {Data}.
\newblock {\em American Political Science Review}, 110(2):278--295.

\bibitem[Bisbee and Spirling, 2026]{bisbee_spirling_2026_llm_gold_standard}
Bisbee, J. and Spirling, A. (2026).
\newblock What to do when humans are no longer the gold standard: Large language models, state of the art and robustness for politics research.
\newblock Working paper, Vanderbilt University and Princeton University.
\newblock First draft: June 13, 2025. This version: February 10, 2026.

\bibitem[Blaydes et~al., 2018]{blaydes_mirrors_2018}
Blaydes, L., Grimmer, J., and McQueen, A. (2018).
\newblock Mirrors for {Princes} and {Sultans}: {Advice} on the {Art} of {Governance} in the {Medieval} {Christian} and {Islamic} {Worlds}.
\newblock {\em The Journal of Politics}, 80(4):1150--1167.

\bibitem[Blumenau and Lauderdale, 2018]{BlumenauLauderdale2018}
Blumenau, J. and Lauderdale, B.~E. (2018).
\newblock Never let a good crisis go to waste: Agenda setting and legislative voting in response to the {EU} crisis.
\newblock {\em The Journal of Politics}, 80(2):462--478.

\bibitem[Budak et~al., 2016]{budak2016fair}
Budak, C., Goel, S., and Rao, J.~M. (2016).
\newblock Fair and balanced? quantifying media bias through crowdsourced content analysis.
\newblock {\em Public Opinion Quarterly}, 80(S1):250--271.

\bibitem[Bueno~de Mesquita et~al., 2015]{bueno2015measuring}
Bueno~de Mesquita, E., Fair, C.~C., Jordan, J., Rais, R.~B., and Shapiro, J.~N. (2015).
\newblock Measuring political violence in pakistan: Insights from the bfrs dataset.
\newblock {\em Conflict Management and Peace Science}, 32(5):536--558.

\bibitem[Bush and Clayton, 2023]{bush_facing_2023}
Bush, S.~S. and Clayton, A. (2023).
\newblock Facing {Change}: {Gender} and {Climate} {Change} {Attitudes} {Worldwide}.
\newblock {\em American Political Science Review}, 117(2):591--608.

\bibitem[Buzan et~al., 1998]{BuzanWaeverDeWilde1998}
Buzan, B., W{\ae}ver, O., and de~Wilde, J. (1998).
\newblock {\em Security: A New Framework for Analysis}.
\newblock Lynne Rienner Publishers, Boulder, CO.

\bibitem[Carlson and Dell, 2025]{carlson2025unifying}
Carlson, J. and Dell, M. (2025).
\newblock A unifying framework for robust and efficient inference with unstructured data.
\newblock {\em arXiv preprint arXiv:2505.00282}.

\bibitem[Clark et~al., 2021]{clark2021all}
Clark, E., August, T., Serrano, S., Haduong, N., Gururangan, S., and Smith, N.~A. (2021).
\newblock All that’s ‘human’is not gold: Evaluating human evaluation of generated text.
\newblock In {\em Proceedings of the 59th Annual Meeting of the Association for Computational Linguistics and the 11th International Joint Conference on Natural Language Processing (Volume 1: Long Papers)}, pages 7282--7296.

\bibitem[Clark and Dolan, 2021]{ClarkDolan2021}
Clark, R. and Dolan, L.~R. (2021).
\newblock Pleasing the principal: {U.S.} influence in world bank policymaking.
\newblock {\em American Journal of Political Science}, 65(1):36--51.

\bibitem[Dunivin, 2025]{dunivin2025scaling}
Dunivin, Z.~O. (2025).
\newblock Scaling hermeneutics: a guide to qualitative coding with llms for reflexive content analysis.
\newblock {\em EPJ Data Science}, 14(1):28.

\bibitem[Egami et~al., 2023]{egami2024usingimperfectsurrogatesdownstream}
Egami, N., Hinck, M., Stewart, B., and Wei, H. (2023).
\newblock Using imperfect surrogates for downstream inference: Design-based supervised learning for social science applications of large language models.
\newblock {\em Advances in Neural Information Processing Systems}, 36:68589--68601.

\bibitem[Egami et~al., 2024]{egami2024using}
Egami, N., Hinck, M., Stewart, B.~M., and Wei, H. (2024).
\newblock Using large language model annotations for the social sciences: A general framework of using predicted variables in downstream analyses.
\newblock {\em Preprint from November}, 17:2024.

\bibitem[Feltovich and Giovannoni, 2024]{feltovich_campaign_2024}
Feltovich, N. and Giovannoni, F. (2024).
\newblock Campaign {Messages}, {Polling}, and {Elections}: {Theory} and {Experimental} {Evidence}.
\newblock {\em American Journal of Political Science}, 68(2):408--426.
\newblock \_eprint: https://onlinelibrary.wiley.com/doi/pdf/10.1111/ajps.12722.

\bibitem[Fong and Tyler, 2021]{fong_machine_2021}
Fong, C. and Tyler, M. (2021).
\newblock Machine {Learning} {Predictions} as {Regression} {Covariates}.
\newblock {\em Political Analysis}, 29(4):467--484.

\bibitem[{Gemma Team} et~al., 2025]{gemmateam2025gemma3}
{Gemma Team}, Kamath, A., Ferret, J., Pathak, S., Vieillard, N., Merhej, R., Perrin, S., Matejovicova, T., Ramé, A., Rivière, M., et~al. (2025).
\newblock Gemma 3 technical report.
\newblock {\em arXiv preprint arXiv:2503.19786}.

\bibitem[Gielens et~al., 2026]{gielens2026goodbye}
Gielens, E., Sowula, J., and Leifeld, P. (2026).
\newblock Goodbye human annotators? content analysis of social policy debates using chatgpt.
\newblock {\em Journal of Social Policy}, 55(2):385--404.

\bibitem[Gilardi et~al., 2023]{gilardi_chatgpt_2023}
Gilardi, F., Alizadeh, M., and Kubli, M. (2023).
\newblock {ChatGPT} outperforms crowd workers for text-annotation tasks.
\newblock {\em Proceedings of the National Academy of Sciences}, 120(30):e2305016120.

\bibitem[Gilardi et~al., 2021]{gilardi_policy_2021}
Gilardi, F., Shipan, C.~R., and Wüest, B. (2021).
\newblock Policy {Diffusion}: {The} {Issue}-{Definition} {Stage}.
\newblock {\em American Journal of Political Science}, 65(1):21--35.
\newblock \_eprint: https://onlinelibrary.wiley.com/doi/pdf/10.1111/ajps.12521.

\bibitem[Gong et~al., 2026]{gong2026limits}
Gong, L., Hopkins, D.~J., and Wolken, S. (2026).
\newblock The limits of de-politicizing--and also of annotation: A case study in russian media outlets’ social media posts, 2016--2024.
\newblock In {\em Proceedings of the International AAAI Conference on Web and Social Media}, volume~20, pages 889--909.

\bibitem[Goodall et~al., 2026]{goodall2026large}
Goodall, L.~S., Shilton, D., Mullins, D.~A., and Whitehouse, H. (2026).
\newblock Large language models struggle with ethnographic text annotation.
\newblock {\em arXiv preprint arXiv:2601.12099}.

\bibitem[Grattafiori et~al., 2024]{meta2024llama3}
Grattafiori, A., Dubey, A., Jauhri, A., Pandey, A., Kadian, A., Al-Dahle, A., Letman, A., Mathur, A., Schelten, A., Vaughan, A., et~al. (2024).
\newblock The llama 3 herd of models.
\newblock {\em arXiv preprint arXiv:2407.21783}.

\bibitem[Grimmer et~al., 2021]{grimmer2021machine}
Grimmer, J., Roberts, M.~E., and Stewart, B.~M. (2021).
\newblock Machine learning for social science: An agnostic approach.
\newblock {\em Annual Review of Political Science}, 24:395--419.

\bibitem[Grimmer et~al., 2022]{grimmer2022text}
Grimmer, J., Roberts, M.~E., and Stewart, B.~M. (2022).
\newblock {\em Text as data: A new framework for machine learning and the social sciences}.
\newblock Princeton University Press.

\bibitem[Grimmer and Stewart, 2013]{grimmer_text_2013}
Grimmer, J. and Stewart, B.~M. (2013).
\newblock Text as {Data}: {The} {Promise} and {Pitfalls} of {Automatic} {Content} {Analysis} {Methods} for {Political} {Texts}.
\newblock {\em Political Analysis}, 21(3):267--297.

\bibitem[Halterman and Keith, 2026]{Halterman2026CodebookLLMs}
Halterman, A. and Keith, K.~A. (2026).
\newblock Codebook llms: Evaluating llms as measurement tools for political science concepts.
\newblock {\em Political Analysis}, 34(2):188--204.

\bibitem[Hase, 2022]{hase2022automated}
Hase, V. (2022).
\newblock Automated content analysis.
\newblock In {\em Standardisierte Inhaltsanalyse in der Kommunikationswissenschaft--Standardized content analysis in communication research: Ein Handbuch-a handbook}, pages 23--36. Springer.

\bibitem[Hendrycks et~al., 2021]{hendrycks2021mmlu}
Hendrycks, D., Burns, C., Basart, S., Zou, A., Mazeika, M., Song, D., and Steinhardt, J. (2021).
\newblock Measuring massive multitask language understanding.
\newblock In {\em International Conference on Learning Representations (ICLR)}.

\bibitem[Heseltine and Clemm~von Hohenberg, 2024]{heseltine2024large}
Heseltine, M. and Clemm~von Hohenberg, B. (2024).
\newblock Large language models as a substitute for human experts in annotating political text.
\newblock {\em Research \& Politics}, 11(1):20531680241236239.

\bibitem[Hosking et~al., 2024]{hosking2024human}
Hosking, T., Blunsom, P., and Bartolo, M. (2024).
\newblock Human feedback is not gold standard.
\newblock In {\em International Conference on Learning Representations}, volume 2024, pages 55864--55883.

\bibitem[Imai and Yamamoto, 2010]{imai_causal_2010}
Imai, K. and Yamamoto, T. (2010).
\newblock Causal {Inference} with {Differential} {Measurement} {Error}: {Nonparametric} {Identification} and {Sensitivity} {Analysis}.
\newblock {\em American Journal of Political Science}, 54(2):543--560.
\newblock \_eprint: https://onlinelibrary.wiley.com/doi/pdf/10.1111/j.1540-5907.2010.00446.x.

\bibitem[Jiang et~al., 2023]{jiang2023mistral}
Jiang, A.~Q., Sablayrolles, A., Mensch, A., Bamford, C., Chaplot, D.~S., de~las Casas, D., Bressand, F., Lengyel, G., Lample, G., Saulnier, L., Lavaud, L.~R., Lachaux, M.-A., Stock, P., Le~Scao, T., Lavril, T., Wang, T., Lacroix, T., and El~Sayed, W. (2023).
\newblock Mistral 7{B}.
\newblock {\em arXiv preprint arXiv:2310.06825}.

\bibitem[Jiang and Marneffe, 2022]{jiang2022investigating}
Jiang, N.-J. and Marneffe, M.-C.~d. (2022).
\newblock Investigating reasons for disagreement in natural language inference.
\newblock {\em Transactions of the Association for Computational Linguistics}, 10:1357--1374.

\bibitem[Jung, 2020]{jung_mobilizing_2020}
Jung, J.-H. (2020).
\newblock The {Mobilizing} {Effect} of {Parties}' {Moral} {Rhetoric}.
\newblock {\em American Journal of Political Science}, 64(2):341--355.
\newblock \_eprint: https://onlinelibrary.wiley.com/doi/pdf/10.1111/ajps.12476.

\bibitem[Kennedy et~al., 2020]{kennedy_shape_2020}
Kennedy, R., Clifford, S., Burleigh, T., Waggoner, P.~D., Jewell, R., and Winter, N. J.~G. (2020).
\newblock The shape of and solutions to the {MTurk} quality crisis.
\newblock {\em Political Science Research and Methods}, 8(4):614--629.

\bibitem[Kracauer, 1952]{kracauer1952challenge}
Kracauer, S. (1952).
\newblock The challenge of qualitative content analysis.
\newblock {\em Public opinion quarterly}, pages 631--642.

\bibitem[Krippendorff, 2018]{krippendorff2018content}
Krippendorff, K. (2018).
\newblock {\em Content analysis: An introduction to its methodology}.
\newblock Sage publications.

\bibitem[Lacombe, 2019]{lacombe2019political}
Lacombe, M.~J. (2019).
\newblock The political weaponization of gun owners: The national rifle association’s cultivation, dissemination, and use of a group social identity.
\newblock {\em The Journal of Politics}, 81(4):1342--1356.

\bibitem[Levy, 1995]{Levy1995}
Levy, M.~A. (1995).
\newblock Is the environment a national security issue?
\newblock {\em International Security}, 20(2):35--62.

\bibitem[Liu and Sun, 2025]{liu2025voices}
Liu, A. and Sun, M. (2025).
\newblock From voices to validity: Leveraging large language models (llms) for textual analysis of policy stakeholder interviews.
\newblock {\em AERA Open}, 11:23328584251374595.

\bibitem[Lombard et~al., 2002]{lombard2002content}
Lombard, M., Snyder-Duch, J., and Bracken, C.~C. (2002).
\newblock Content analysis in mass communication: Assessment and reporting of intercoder reliability.
\newblock {\em Human communication research}, 28(4):587--604.

\bibitem[Lowe and Benoit, 2013]{lowe2013validating}
Lowe, W. and Benoit, K. (2013).
\newblock Validating estimates of latent traits from textual data using human judgment as a benchmark.
\newblock {\em Political analysis}, 21(3):298--313.

\bibitem[Mattingly et~al., 2025]{mattingly_chinese_2025}
Mattingly, D., Incerti, T., Ju, C., Moreshead, C., Tanaka, S., and Yamagishi, H. (2025).
\newblock Chinese state media persuades a global audience that the “{China} model” is superior: {Evidence} from a 19-country experiment.
\newblock {\em American Journal of Political Science}, 69(3):1029--1046.
\newblock \_eprint: https://onlinelibrary.wiley.com/doi/pdf/10.1111/ajps.12887.

\bibitem[Meng et~al., 2024]{meng2024exploring}
Meng, H., Yang, Y., Fu, W., Lee, J., Li, Y., and Lee, Y.-C. (2024).
\newblock Exploring the human-llm synergy in advancing theory-driven qualitative analysis.
\newblock {\em arXiv preprint arXiv:2405.05758}.

\bibitem[Messick, 1995]{messick1995validity}
Messick, S. (1995).
\newblock Validity of psychological assessment: Validation of inferences from persons' responses and performances as scientific inquiry into score meaning.
\newblock {\em American psychologist}, 50(9):741.

\bibitem[Mikhaylov et~al., 2012]{mikhaylov2012coder}
Mikhaylov, S., Laver, M., and Benoit, K.~R. (2012).
\newblock Coder reliability and misclassification in the human coding of party manifestos.
\newblock {\em Political analysis}, 20(1):78--91.

\bibitem[Mu et~al., 2024]{mu2024navigating}
Mu, Y., Wu, B.~P., Thorne, W., Robinson, A., Aletras, N., Scarton, C., Bontcheva, K., and Song, X. (2024).
\newblock Navigating prompt complexity for zero-shot classification: A study of large language models in computational social science.
\newblock {\em arXiv preprint arXiv:2305.14310}.

\bibitem[M{\"u}ller and Proksch, 2024]{muller2024nostalgia}
M{\"u}ller, S. and Proksch, S.-O. (2024).
\newblock Nostalgia in european party politics: a text-based measurement approach.
\newblock {\em British Journal of Political Science}, 54(3):993--1005.

\bibitem[Nakamura, 2025]{nakamura2025surrogate}
Nakamura, K. (2025).
\newblock Surrogate representation inference for text and image annotations.
\newblock {\em arXiv preprint arXiv:2509.12416}.

\bibitem[Neuendorf et~al., 2002]{neuendorf2002content}
Neuendorf, K.~A., Skalski, P.~D., Cajigas, J.~A., and Allen, J.~C. (2002).
\newblock {\em The content analysis guidebook}, volume 300.
\newblock sage Thousand Oaks, CA.

\bibitem[{OpenAI} et~al., 2025]{openai2025gptoss}
{OpenAI}, Agarwal, S., Ahmad, L., Ai, J., Altman, S., Applebaum, A., Arbus, E., Arora, R.~K., Bai, Y., Baker, B., Bao, H., Barak, B., Bennett, A., Bertao, T., Brett, N., Brevdo, E., Brockman, G., Bubeck, S., et~al. (2025).
\newblock gpt-oss-120b \& gpt-oss-20b model card.
\newblock {\em arXiv preprint arXiv:2508.10925}.

\bibitem[Osnabrügge et~al., 2021]{osnabrugge_playing_2021}
Osnabrügge, M., Hobolt, S.~B., and Rodon, T. (2021).
\newblock Playing to the {Gallery}: {Emotive} {Rhetoric} in {Parliaments}.
\newblock {\em American Political Science Review}, 115(3):885--899.

\bibitem[Pangakis and Wolken, 2025]{pangakis2025keeping}
Pangakis, N. and Wolken, S. (2025).
\newblock Keeping humans in the loop: Human-centered automated annotation with generative ai.
\newblock In {\em Proceedings of the International AAAI Conference on Web and Social Media}, volume~19, pages 1471--1492.

\bibitem[Park, 2021]{park2021grandstand}
Park, J.~Y. (2021).
\newblock When do politicians grandstand? measuring message politics in committee hearings.
\newblock {\em The Journal of Politics}, 83(1):214--228.

\bibitem[Parthasarathy et~al., 2019]{parthasarathy_deliberative_2019}
Parthasarathy, R., Rao, V., and Palaniswamy, N. (2019).
\newblock Deliberative {Democracy} in an {Unequal} {World}: {A} {Text}-{As}-{Data} {Study} of {South} {India}’s {Village} {Assemblies}.
\newblock {\em American Political Science Review}, 113(3):623--640.

\bibitem[Pavlick and Kwiatkowski, 2019]{pavlick2019inherent}
Pavlick, E. and Kwiatkowski, T. (2019).
\newblock Inherent disagreements in human textual inferences.
\newblock {\em Transactions of the Association for Computational Linguistics}, 7:677--694.

\bibitem[Peer et~al., 2014]{PeerVosgerauAcquisti2014}
Peer, E., Vosgerau, J., and Acquisti, A. (2014).
\newblock Reputation as a sufficient condition for data quality on {Amazon Mechanical Turk}.
\newblock {\em Behavior Research Methods}, 46(4):1023--1031.

\bibitem[Phillis et~al., 2018]{PhillisEtAl2018}
Phillis, Y.~A., Chairetis, N., Grigoroudis, E., et~al. (2018).
\newblock Climate security assessment of countries.
\newblock {\em Climatic Change}, 148:25--43.

\bibitem[Pitler and Nenkova, 2008]{pitler2008revisiting}
Pitler, E. and Nenkova, A. (2008).
\newblock Revisiting readability: A unified framework for predicting text quality.
\newblock In {\em Proceedings of the 2008 conference on empirical methods in natural language processing}, pages 186--195.

\bibitem[Plank, 2022]{plank2022problem}
Plank, B. (2022).
\newblock The “problem” of human label variation: On ground truth in data, modeling and evaluation.
\newblock In {\em Proceedings of the 2022 conference on empirical methods in natural language processing}, pages 10671--10682.

\bibitem[Poesio and Artstein, 2005]{poesio2005reliability}
Poesio, M. and Artstein, R. (2005).
\newblock The reliability of anaphoric annotation, reconsidered: Taking ambiguity into account.
\newblock In {\em Proceedings of the workshop on frontiers in corpus annotations ii: Pie in the sky}, pages 76--83.

\bibitem[Quinn et~al., 2010]{quinn_how_2010}
Quinn, K.~M., Monroe, B.~L., Colaresi, M., Crespin, M.~H., and Radev, D.~R. (2010).
\newblock How to {Analyze} {Political} {Attention} with {Minimal} {Assumptions} and {Costs}.
\newblock {\em American Journal of Political Science}, 54(1):209--228.
\newblock \_eprint: https://onlinelibrary.wiley.com/doi/pdf/10.1111/j.1540-5907.2009.00427.x.

\bibitem[{Qwen Team}, 2024]{qwen2025qwen25}
{Qwen Team} (2024).
\newblock Qwen2.5 technical report.
\newblock {\em arXiv preprint arXiv:2412.15115}.

\bibitem[Rathje et~al., 2024]{rathje2024gpt}
Rathje, S., Mirea, D.-M., Sucholutsky, I., Marjieh, R., Robertson, C.~E., and Van~Bavel, J.~J. (2024).
\newblock {GPT} is an effective tool for multilingual psychological text analysis.
\newblock {\em Proceedings of the National Academy of Sciences}, 121(34):e2308950121.

\bibitem[Rheault et~al., 2019]{rheault2019politicians}
Rheault, L., Rayment, E., and Musulan, A. (2019).
\newblock Politicians in the line of fire: Incivility and the treatment of women on social media.
\newblock {\em Research \& Politics}, 6(1):1--7.

\bibitem[R{\o}nnfeldt, 1997]{Ronnfeldt1997}
R{\o}nnfeldt, C.~F. (1997).
\newblock Review essay: Three generations of environment and security research.
\newblock {\em Journal of Peace Research}, 34(4):473--482.

\bibitem[Schub, 2022]{Schub2022}
Schub, R. (2022).
\newblock Informing the leader: Bureaucracies and international crises.
\newblock {\em American Political Science Review}, 116(4):1460--1476.

\bibitem[Snow et~al., 2008]{snow2008cheap}
Snow, R., O’connor, B., Jurafsky, D., and Ng, A.~Y. (2008).
\newblock Cheap and fast--but is it good? evaluating non-expert annotations for natural language tasks.
\newblock In {\em Proceedings of the 2008 conference on empirical methods in natural language processing}, pages 254--263.

\bibitem[Theocharis et~al., 2016]{theocharis2016bad}
Theocharis, Y., Barber{\'a}, P., Fazekas, Z., Popa, S.~A., and Parnet, O. (2016).
\newblock A bad workman blames his tweets: The consequences of citizens' uncivil twitter use when interacting with party candidates.
\newblock {\em Journal of Communication}, 66(6):1007--1031.

\bibitem[Thrall, 2025]{thrall_informational_2025}
Thrall, C. (2025).
\newblock Informational lobbying and commercial diplomacy.
\newblock {\em American Journal of Political Science}, 69(3):1147--1162.
\newblock \_eprint: https://onlinelibrary.wiley.com/doi/pdf/10.1111/ajps.12873.

\bibitem[T{\"o}rnberg, 2025]{tornberg2025large}
T{\"o}rnberg, P. (2025).
\newblock Large language models outperform expert coders and supervised classifiers at annotating political social media messages.
\newblock {\em Social Science Computer Review}, 43(6):1181--1195.

\bibitem[Uma et~al., 2021]{uma2021learning}
Uma, A.~N., Fornaciari, T., Hovy, D., Paun, S., Plank, B., and Poesio, M. (2021).
\newblock Learning from disagreement: A survey.
\newblock {\em Journal of Artificial Intelligence Research}, 72:1385--1470.

\bibitem[Vincze et~al., 2008]{vincze2008bioscope}
Vincze, V., Szarvas, G., Farkas, R., M{\'o}ra, G., and Csirik, J. (2008).
\newblock The bioscope corpus: biomedical texts annotated for uncertainty, negation and their scopes.
\newblock {\em BMC bioinformatics}, 9(Suppl 11):S9.

\bibitem[Vogler, 2023]{Vogler2023}
Vogler, A. (2023).
\newblock Tracking climate securitization: Framings of climate security by civil and defense ministries.
\newblock {\em International Studies Review}, 25(2):viad010.

\bibitem[Weber and Reichardt, 2024]{weber2024evaluation}
Weber, M. and Reichardt, M. (2024).
\newblock Evaluation is all you need: Prompting generative large language models for annotation tasks in the social sciences: A primer using open models.
\newblock {\em arXiv preprint arXiv:2401.00284}.

\bibitem[Westwood, 2025]{Westwood2025}
Westwood, S.~J. (2025).
\newblock The potential existential threat of large language models to online survey research.
\newblock {\em Proceedings of the National Academy of Sciences}, 122(47):e2518075122.

\bibitem[Xiong et~al., 2025]{xiong2025co}
Xiong, C., Ni, J., Fan, Y., Zouhar, V., Rooein, D., Calvo-Bartolom{\'e}, L., Hoyle, A.~M., Jin, Z., Sachan, M., Leippold, M., et~al. (2025).
\newblock Co-detect: Collaborative discovery of edge cases in text classification.
\newblock In {\em Proceedings of the 2025 Conference on Empirical Methods in Natural Language Processing: System Demonstrations}, pages 354--364.

\bibitem[Yang et~al., 2025]{yang2025data}
Yang, E., Wang, Z., Zhou, C., and Xu, Y. (2025).
\newblock Data annotation with large language models: Lessons from a large empirical evaluation.

\bibitem[Ying et~al., 2022]{ying2022}
Ying, L., Montgomery, J.~M., and Stewart, B.~M. (2022).
\newblock Topics, concepts, and measurement: A crowdsourced procedure for validating topics as measures.
\newblock {\em Political Analysis}, 30(4):570--589.

\bibitem[Yuan et~al., 2025]{yuan2025case}
Yuan, M., Chen, J., Xing, Z., Mohammadi, G., and Quigley, A. (2025).
\newblock A case study of scalable content annotation using multi-llm consensus and human review.
\newblock {\em arXiv preprint arXiv:2503.17620}.

\bibitem[Zhang and de~Marneffe, 2021]{zhang2021identifying}
Zhang, X.~F. and de~Marneffe, M.-C. (2021).
\newblock Identifying inherent disagreement in natural language inference.
\newblock In {\em Proceedings of the 2021 Conference of the North American Chapter of the Association for Computational Linguistics: Human Language Technologies}, pages 4908--4915.

\bibitem[Zhu et~al., 2023]{zhu2023can}
Zhu, Y., Zhang, P., Haq, E.-U., Hui, P., and Tyson, G. (2023).
\newblock Can chatgpt reproduce human-generated labels? a study of social computing tasks.
\newblock {\em arXiv preprint arXiv:2304.10145}.

\bibitem[Ziems et~al., 2024]{ziems2024llm}
Ziems, C., Held, W., Shaikh, O., Chen, J., Zhang, Z., and Yang, D. (2024).
\newblock Can large language models transform computational social science?
\newblock {\em Computational Linguistics}, 50(1):237--291.

\bibitem[Zrnic and Candès, 2024]{zrnic_cross-prediction-powered_2024}
Zrnic, T. and Candès, E.~J. (2024).
\newblock Cross-{Prediction}-{Powered} {Inference}.
\newblock arXiv:2309.16598 [stat].

\end{thebibliography}

\clearpage
\appendix
\ifdefined\PASubmission
  \singlespacing
  \setcounter{page}{1}
\fi
\startcontents[appendix]

% Supplemental numbering
\setcounter{equation}{0}
\setcounter{figure}{0}
\setcounter{table}{0}
\setcounter{section}{0}

\renewcommand{\theequation}{S\arabic{equation}}
\renewcommand{\thefigure}{S\arabic{figure}}
\renewcommand{\thetable}{S\arabic{table}}
\renewcommand{\thesection}{S\arabic{section}}

\begin{center}
\ifdefined\PASubmission
    {\LARGE\bfseries Online Appendix}\par\vspace{0.75em}
    {\large Observational Equivalence of LLM and Human Annotation}\par\vspace{0.5em}
    {\large Kentaro Nakamura, Jing Ling Tan, and George Yean}
\else
    {\LARGE\bfseries Appendix}
\fi
\end{center}

\section*{Appendix Contents}

% 1 = begin with sections
% 2 = include through subsections
\printcontents[appendix]{}{1}[3]{}

\clearpage

{
\section{Validating our Research Design}

Our main analysis compares annotations produced by LLMs and the three
authors of this paper under standardized codebooks.
This section examines whether our findings depend on the use of standardized codebooks or annotations produced by the authors of this paper. To address
these concerns, we compare our annotations with external expert
annotations and extend our analysis to studies with original expert-developed codebooks.

\subsection{Validation of Human Coding Against External Benchmarks}\label{sec:validation_coding}

The three human coders in our main analysis are the authors of this
paper, rather than subject-matter experts in all of the substantive
domains examined.
We therefore conduct two supplementary
comparisons to assess how closely our annotations correspond to the annotations made by domain-specific experts.

\medskip
\noindent \textbf{Comparison with source-author annotations.}
First, we compare our annotations with the original human annotations
from the three replicated studies whose coders classified the same binary
concept that we code: \cite{feltovich_campaign_2024},
\cite{lacombe2019political}, and \cite{Schub2022}. Consistent with the
premise of this paper, we do not treat the original expert annotations as a gold
standard. Instead, we ask whether the original annotators behave like
additional members of our coding panel. Specifically, for each text, we compute the share of
disagreeing pairs among (i) the three pairs formed by the original annotation
and each of our coders and (ii) the three pairs of our own coders. Within each study, we then estimate
the mean difference between (i) and (ii) across texts, together with its
95\% confidence interval. If our human coding departs systematically from the
original expert coding, for example because of differences in the coding
instructions or expertise, disagreement with the original annotations should be
significantly higher than disagreement among our coders, making the difference estimates positive. Because we
developed a revised codebook for \cite{Schub2022}, we report results
under both our original and revised codebooks.

\begin{figure}[h]
    \centering
    \includegraphics[width=1.0\linewidth]{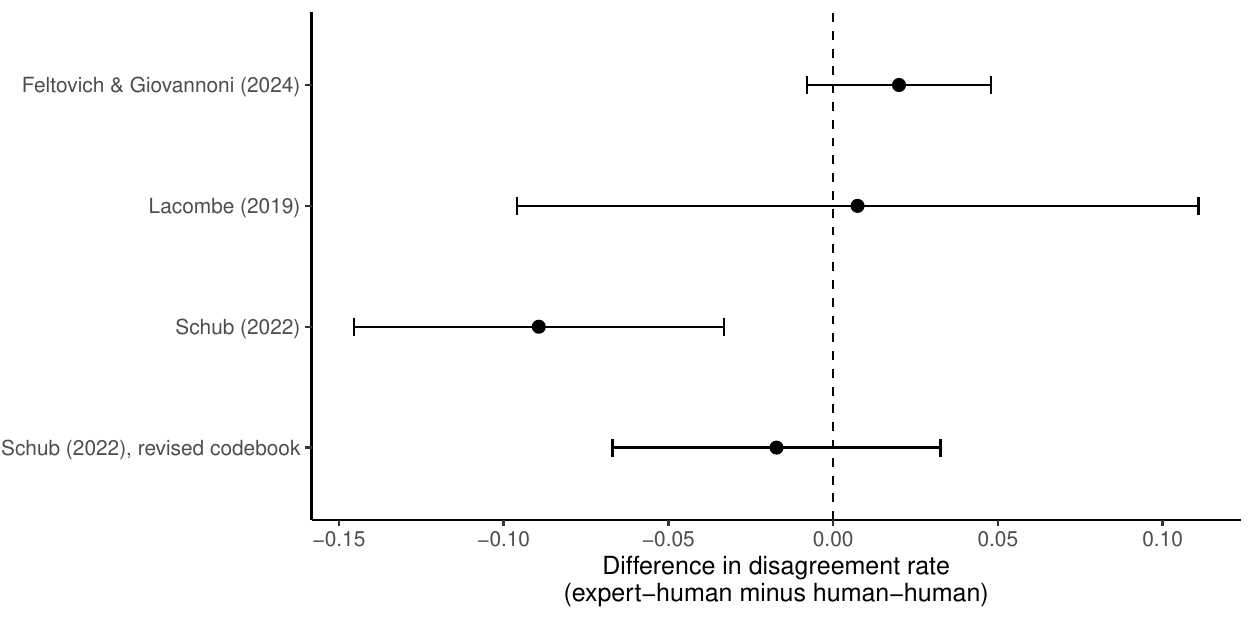}
    \caption{Difference between disagreement with the original expert annotations and disagreement among our three human coders. For each study, the point estimate is the mean difference between (i) the average disagreement rate between the original expert annotation and each of our three human coders and (ii) the average pairwise disagreement rate among our three human coders. Horizontal bars indicate 95\% confidence intervals, and the dashed vertical line marks zero. Positive values indicate that our human coders disagree with the original expert annotations more often than they disagree with one another, while negative values indicate the reverse. For \cite{Schub2022}, we report results under both the original and revised codebooks.}
    \label{fig_disagreement}
\end{figure}

Figure~\ref{fig_disagreement} reports the results. Across all three studies, we find no evidence that disagreement between our human coders and the original expert annotations is greater than disagreement among our human coders themselves. For \cite{feltovich_campaign_2024}, \cite{lacombe2019political}, and \cite{Schub2022} under the revised codebook, the two disagreement rates are nearly identical (the difference is almost zero by point estimates) and not statistically significantly different.
For \cite{Schub2022} under the original codebook, the estimated difference is even negative, indicating that our human coders agree with the expert annotations more often than they agree with one another. 
Overall, these results provide no indication that our human annotations differ from the original expert annotations beyond the level of disagreement observed among our own human coders, supporting the use of our human annotations as a reasonable benchmark.

\medskip
\noindent \textbf{Comparison with subject-matter specialists.}
Because the original studies provide only a single reference label per text, the comparison above can reveal whether our coders diverge from expert annotations, but cannot show how much experts disagree with one another or whether their disagreements are concentrated on the same texts as those of other coders.

We therefore recruited two subject-matter experts for
\citet{arias2022securitizes} and three for
\citet{ClarkDolan2021}, all holding Ph.D. degrees in political science and
working and/or publishing in areas corresponding directly to the topics of each paper. Each independently
annotated the same 20 texts used in our crowd-sourced
study, comprising the 10 texts with the highest and
with the lowest Gini impurity across LLMs, following the same codebooks
as our coders, crowd workers, and LLMs. For these texts,
we examine disagreement both within and across
four groups: subject-matter experts, our coders
(attentive non-experts), crowd workers, and LLMs. This design allows us to examine two questions. First, does the level of disagreement among subject-matter experts differ systematically from that among other annotator groups (our coders, crowd workers, and LLMs)? Second, does disagreement across all four groups concentrate on the same texts, as would be expected if it also reflects ambiguity in the text and coding rule that expertise does not resolve?

Table~\ref{tab:expert_group_agreement} presents the results.
For texts with low cross-LLM disagreement (Panel A: Agree stratum),
experts agree with one another 93\% of the time, and agreement within
and between our human coders and LLMs is similarly high.
In contrast, crowd workers exhibit substantially lower agreement rates.
For texts with high cross-LLM disagreement (Panel B: Disagree stratum),
however, expert--expert agreement declines sharply, from 93\% to 60\%,
closely mirroring the declines in human--human and LLM--LLM agreement.
Agreement between experts, our coders, and LLMs also declines
substantially. Interestingly, we do not observe a comparable decline
in within-group agreement among crowd workers, whose agreement remains
relatively low in both strata.

These findings provide further evidence that annotation disagreement
reflects shared difficulties in interpreting texts and applying coding
rules, even among subject-matter experts. Importantly, cross-LLM
disagreement identifies texts that also divide experts, suggesting
that it can serve as an effective tool for identifying difficult cases.
Thus, our main findings also appear to extend beyond our three human
coders to subject-matter experts.

\begin{table}[htbp]
\centering
\caption{Agreement within and across annotator groups.
Panel A (agree) and Panel B (disagree) contain the 10 texts
with the lowest and highest Gini impurity across LLMs, respectively. Each cell reports the proportion of identical annotations. Diagonal entries measure
within-group agreement, while off-diagonal entries measure
between-group agreement.}
\label{tab:expert_group_agreement}
\small
\setlength{\tabcolsep}{4pt}

\begin{minipage}[t]{0.49\linewidth}
\centering
\textbf{Panel A: Agree stratum}
\par\smallskip
\begin{tabular}{@{}lcccc@{}}
\toprule
 & Expert & Human & LLM & Crowd \\
\midrule
Expert & 0.933 & 0.900 & 0.895 & 0.683 \\
Human  & 0.900 & 0.933 & 0.959 & 0.664 \\
LLM    & 0.895 & 0.959 & 0.986 & 0.683 \\
Crowd  & 0.683 & 0.664 & 0.683 & 0.604 \\
\bottomrule
\end{tabular}
\end{minipage}
\hfill
\begin{minipage}[t]{0.49\linewidth}
\centering
\textbf{Panel B: Disagree stratum}
\par\smallskip
\begin{tabular}{@{}lcccc@{}}
\toprule
 & Expert & Human & LLM & Crowd \\
\midrule
Expert & 0.600 & 0.555 & 0.564 & 0.658 \\
Human  & 0.555 & 0.600 & 0.582 & 0.514 \\
LLM    & 0.564 & 0.582 & 0.571 & 0.566 \\
Crowd  & 0.658 & 0.514 & 0.566 & 0.656 \\
\bottomrule
\end{tabular}
\end{minipage}
\end{table}

\subsection{Validation Using Original Expert-Developed Codebooks}
\label{sec:validation_codebook}

Another concern with our replication design is the quality of the codebooks used.
Our main analysis uses codebooks developed through a standardized procedure because most published studies do not make their original coding instructions available.
Although this approach facilitates comparisons across studies, our reconstructed codebooks may not preserve all substantively important distinctions in the original coding schemes.
It therefore remains unclear whether our findings generalize to tasks using more detailed, expert-developed codebooks.

To address this concern, we conduct additional analyses of three studies for which independently developed coding instructions and original human annotations are available: \citet{benoit_crowd-sourced_2016}, \citet{muller2024nostalgia}, and the BFRS dataset on political violence in Pakistan, originally developed by \citet{bueno2015measuring} and subsequently reanalyzed by \citet{Halterman2026CodebookLLMs}.
For the first two studies, we use the coding instructions published by the original researchers, while for BFRS we use the detailed codebook adapted by \citet{Halterman2026CodebookLLMs}.
For each study, the three authors of this paper and the same 13 LLMs used in our main analysis independently classify the same 100 texts.
We preserve the original category definitions and classification rules, making only minimal adjustments when adapting the instructions for LLMs, such as specifying the required output format.
Unlike our main analysis, which focuses on binary classification, we retain the original three-category classification task for \citet{benoit_crowd-sourced_2016} and the 12-category classification task for BFRS.
The \citet{muller2024nostalgia} task remains binary, focusing on whether a sentence contains nostalgic rhetoric.
See Appendix~\ref{app:eachstudy_validation} for details of the studies and Appendix~\ref{app:validate_codebook} for the coding instructions.

We first examine whether our main finding on LLM--human agreement holds under these codebooks.
Specifically, we calculate nominal Krippendorff's $\alpha$ separately between each LLM and each individual human coder, and then average these coefficients within each group of human coders.
We consider both our three human coders and the experts (i.e., the original annotators), comprising six coders for \citet{benoit_crowd-sourced_2016}, four coders for \citet{muller2024nostalgia}, and a single reference label per text for BFRS.
For comparison, we also calculate the average pairwise Krippendorff's $\alpha$ within human coders (the three authors of this paper) and the coders in the original paper.

\begin{figure}[ht]
    \centering
    \includegraphics[width=\linewidth]{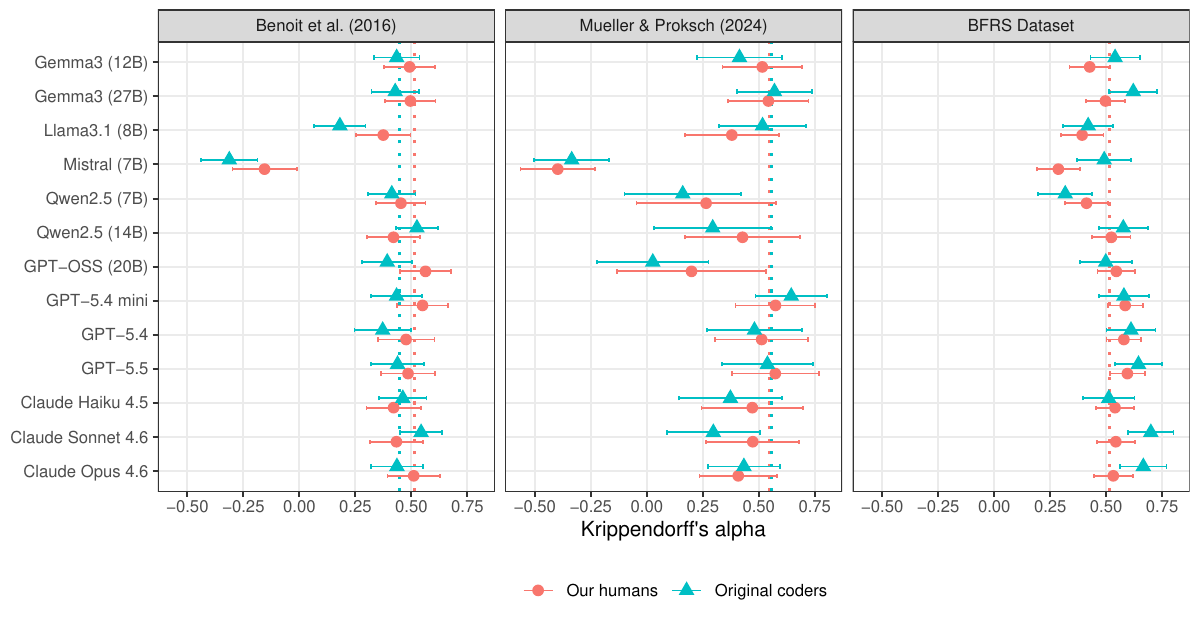}
    \caption{Agreement between LLMs and human annotators using original or externally developed codebooks.
    Each panel presents results for 100 texts classified using the coding instructions from the corresponding study.
    Red circles indicate the average pairwise Krippendorff's $\alpha$ between each LLM and our three human coders, while blue triangles indicate the corresponding average between each LLM and the original annotators.
    The original annotations comprise six individual coders for \cite{benoit_crowd-sourced_2016}, four for \cite{muller2024nostalgia}, and one reference label per text for BFRS.
    Horizontal error bars indicate 95\% confidence intervals.
    Dotted vertical lines indicate the average pairwise Krippendorff's $\alpha$ within our human group and, where available, within the original human group.
    The latter is unavailable for BFRS because only one original reference label per text is observed.}
    \label{fig:original_codebook_models}
\end{figure}

Figure~\ref{fig:original_codebook_models} presents the results.
First, the intercoder reliability among our three human coders is comparable to that among the original coders.
For \citet{benoit_crowd-sourced_2016}, the average pairwise Krippendorff's $\alpha$ is 0.51 among our coders and 0.45 among the original coders.
For \citet{muller2024nostalgia}, the corresponding values are 0.55 and 0.56.
These results provide additional evidence that the reliability of our human annotations is not substantially lower than that of the original human annotations, even when using the original coding instructions.

Second, we see that LLM--human agreement is comparable to human--human or expert--expert agreement for sufficiently capable LLMs even when using the original detailed codebooks.
For \citet{benoit_crowd-sourced_2016}, GPT-5.5 and Claude Opus 4.6 achieve agreement with the original annotators of approximately 0.44, nearly identical to the average agreement of 0.45 among the original human coders themselves.
Similarly, for \citet{muller2024nostalgia}, GPT-5.5 and Gemma 3-27B achieve agreement with the original annotators of 0.54 and 0.57, respectively, comparable to the human--human benchmark of 0.56.
For the BFRS dataset, GPT-5.5 and Claude Opus 4.6 achieve agreement with the original annotations of 0.64 and 0.67, respectively, exceeding the average agreement of 0.46 between our human coders and the original annotations.
Although weaker models do not consistently reach these levels of agreement, these results confirm that sufficiently capable LLMs can achieve LLM--human agreement comparable to human--human agreement even under detailed, independently developed codebooks.

We next examine whether our finding that LLM--human disagreement concentrates on texts that divide humans also holds under these original codebooks.
We first replicate the leave-one-human-out analysis in Section~\ref{sec:disagreement_gini} using our three human coders.
For each text, we hold out one human coder and classify the text as clear when the remaining two humans agree and as ambiguous when they disagree.
We then calculate each LLM's agreement with the held-out human separately for clear and ambiguous texts.
We repeat this procedure for each of the three choices of held-out human and aggregate the results. The procedure is identical to Section~\ref{sec:disagreement_gini} except that we used the detailed, expert-developed codebook.

\begin{figure}[ht]
    \centering
    \includegraphics[width=\linewidth]{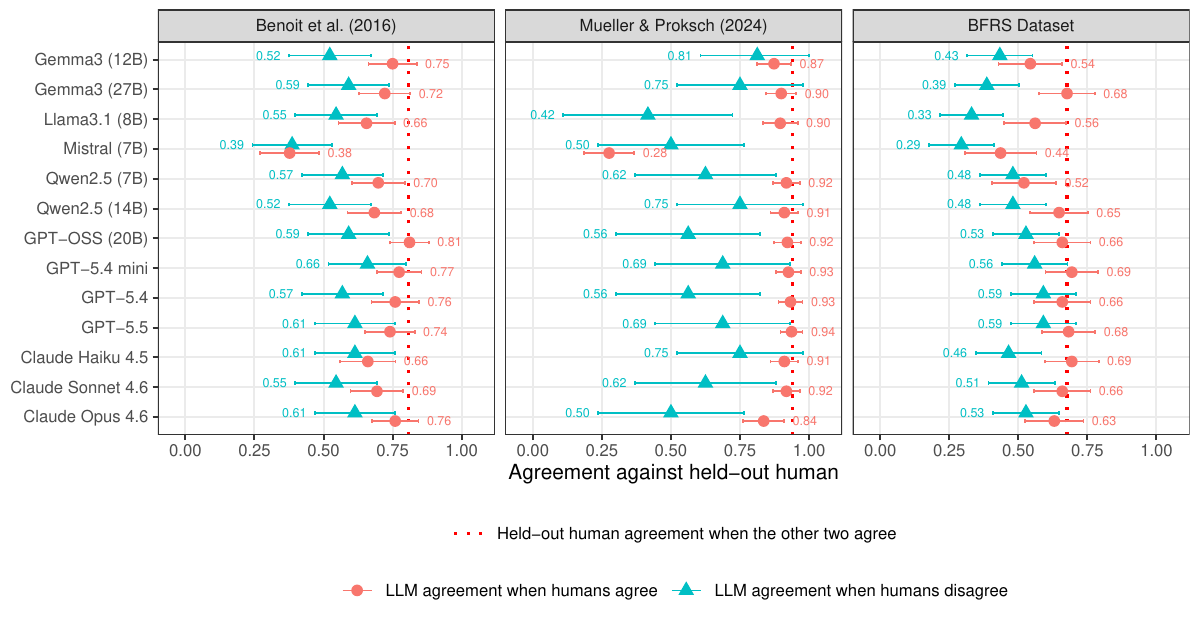}
    \caption{Agreement with a held-out human coder under original or externally developed codebooks.
    For each of the three validation studies, the figure reports agreement between each LLM and a held-out human coder, separately for texts on which the other two human coders agree or disagree.
    The red dotted line indicates the held-out human's agreement with the common label of the other two humans when they agree.
    Points represent agreement rates averaged over the three choices of held-out human, with horizontal error bars indicating 95\% delete-one-text jackknife confidence intervals.}
    \label{fig:original_codebook_heldout}
\end{figure}

Figure~\ref{fig:original_codebook_heldout} presents the results.
Consistent with our main analysis, many recent LLMs exhibit substantially lower agreement with a held-out human on texts where the other two humans disagree.
This pattern appears across the three validation studies, despite their differences in substantive content, number of classification categories, and coding instructions.
The magnitude of the difference varies across models, with weaker models not always exhibiting the same pattern.
These findings provide additional evidence that LLM--human disagreement is concentrated on texts that also generate disagreement among human coders, even when using detailed coding instructions developed independently of our main replication procedure.

Finally, we directly examine whether disagreement among the original human annotators is associated with disagreement across LLMs, which provides a further validation of Hypothesis~2-b.
Specifically, for each text $i$, we measure cross-LLM disagreement using nominal Gini impurity,
\begin{align}
    \mathrm{Gini}_i = 1-\sum_{k=1}^{K}p_{ik}^2,
\end{align}
where $p_{ik}$ is the proportion of LLMs assigning category $k$.
This generalizes the binary Gini impurity used in our main analysis to tasks with multiple categories.
We then compare mean cross-LLM disagreement between texts on which human coders agree and those on which they disagree, using both our human annotations and the original individual annotations. Here, as in the main paper, the disagreement is operationalized through the absence of a strict majority.
Because BFRS provides only one original reference label per text, we conduct this comparison using only our three human coders for that dataset.
We estimate the difference in mean Gini impurity between the two groups of texts and report 95\% confidence intervals from Welch's two-sample $t$-test.

\begin{table}[ht]
\centering
\small
\caption{Cross-LLM disagreement by human agreement under original codebooks.}
\label{tab:original_codebook_gini}
\resizebox{\linewidth}{!}{%
\begin{tabular}{llccccc}
\toprule
Study & Human coders & Agree & Disagree & Difference & 95\% CI & $p$-value \\
\midrule
\cite{benoit_crowd-sourced_2016}
 & Our coders & 0.228 & 0.377 & 0.148 & [0.070, 0.227] & $<0.001$ \\
 & Original coders & 0.276 & 0.364 & 0.088 & [-0.003, 0.178] & 0.057 \\
\addlinespace
\cite{muller2024nostalgia}
 & Our coders & 0.159 & 0.386 & 0.227 & [0.149, 0.305] & $<0.001$ \\
 & Original coders & 0.163 & 0.304 & 0.141 & [0.071, 0.211] & $<0.001$ \\
\addlinespace
BFRS Dataset
 & Our coders & 0.238 & 0.376 & 0.137 & [0.045, 0.230] & 0.004 \\
\bottomrule
\end{tabular}}
\par\smallskip
\begin{minipage}{\linewidth}
\footnotesize
\emph{Notes:} Entries report mean nominal Gini impurity across the 13 LLM predictions on texts where human coders agree or disagree.
%For our coders and the original coders in Müller and Proksch (2024), disagreement indicates that at least one coder assigns a different label.
%For the original coders in Benoit et al. (2016), disagreement indicates that no category receives a strict majority of the six votes.
Differences are calculated as disagreement minus agreement.
Confidence intervals and $p$-values are based on two-sided Welch tests across texts.
Because the number of classification categories differs across studies, absolute Gini levels should not be compared directly across datasets.
\end{minipage}
\end{table}

Table~\ref{tab:original_codebook_gini} presents the results.
Using our three human coders, cross-LLM disagreement is substantially greater on texts where humans disagree in all three studies.
The differences in mean Gini impurity are 0.148 for \citet{benoit_crowd-sourced_2016}, 0.227 for \citet{muller2024nostalgia}, and 0.137 for BFRS, all statistically significant at the 1\% level.
Importantly, we observe similar patterns when using the original human annotations.
For \citet{muller2024nostalgia}, mean cross-LLM Gini impurity increases from 0.163 on texts where the original coders agree to 0.304 where they disagree, a statistically significant difference of 0.141.
For \citet{benoit_crowd-sourced_2016}, cross-LLM Gini impurity also increases, from 0.276 when the original coders reach a majority to 0.364 when they do not.
Although this difference of 0.088 is not statistically significant at the 5\% level ($p=0.057$), its direction is consistent with our main finding.
These results provide further evidence that disagreement across LLMs is concentrated on texts that divide human annotators, including those whose annotations were collected independently of our study.

In sum, these validation studies provide additional support for our main findings.
The reliability of our human coders is comparable to that of the original human coders in the two studies where individual original annotations are available.
Moreover, recent LLMs can achieve substantial agreement with human annotators under independently developed codebooks, although their relative performance varies across tasks.
Most importantly, cross-LLM disagreement is systematically associated with disagreement among both our human coders and the original human annotators.
Thus, our main finding that human and LLM disagreement reflects common difficulties in applying coding rules does not appear to be driven solely by the standardized codebooks constructed for our primary analysis.

\newpage 

\section{Details of Replication Studies}
\subsection{Summary of Each Study (Main Paper)}\label{app:study_details}
For each of the 14 main replication studies, we selected one concept for text classification. When a study examined multiple concepts or topics, we selected the one most closely related to its substantive findings. If multiple concepts were equally relevant, we randomly selected one among them.

In this section, we summarize the paper, the text corpus used, the topics into which those texts are classified, how the topic classification is used in the original analysis, and the topic of interest in our reanalysis. We also report the summary of text classification tasks in  Table~\ref{app:paper_sum}.

\paragraph{Arias (2022), \textit{Who Securitizes? Climate Change Discourse in the United Nations}.} Arias asks why some states frame climate change as a security issue in the United Nations, and which states are most likely to make such securitizing moves. The paper argues that securitization is shaped by agenda-control incentives: P5 states benefit from moving issues toward the Security Council, while highly vulnerable states such as SIDS may resist losing agenda control. Empirically, the paper finds that UN climate discourse overall is not securitized, P5 states are more likely to securitize climate change, and SIDS are less likely to do so. The text data are speeches by state representatives in the UN General Assembly General Debate, filtered to speech segments discussing climate change. The analysis uses 4,525 climate-related speech segments from 1,987 speeches, with the earliest climate-related speech segment appearing in 1984.

\paragraph{Blaydes et al. (2018), \textit{Mirrors for princes and sultans: advice on the art of governance in the medieval Christian and Islamic worlds}.} The paper compares medieval Christian and Islamic political advice texts (“mirrors for princes”) to examine how ideas of governance evolved across regions. Using text-as-data methods, it identifies shared thematic structures and tracks their evolution over time. It finds broad similarities across traditions but key divergences, including declining religious emphasis in Europe and shifting ruler-centered discourse in the Islamic world. The data consist of translated medieval political advice texts addressed to rulers in Christian Europe and the Islamic world. These texts include guidance on governance, morality, religion, and social order, often written by elites for kings, courts, or political audiences.

\paragraph{Blumenau \&
Lauderdale (2018), \textit{Never let a good crisis go to waste: Agenda setting and legislative voting in response to the EU crisis}.} Blumenau and Lauderdale study how the 2008 global financial and sovereign debt crises affected agenda setting and legislative voting in the European Parliament. They argue that crises weaken legislators' attachment to the status quo, allowing pro-integration agenda setters to pass more integrationist policy than would otherwise have been possible. Empirically, they show that crisis-related votes increasingly divided MEPs along the pro- versus anti-European integration dimension rather than the left-right dimension. The text data are European Parliament legislative summary texts linked to roll-call votes from the sixth and seventh European Parliaments, covering 2004-2014. These summaries describe the purpose, background, and content of legislation under discussion, and are used to identify whether votes concern crisis-relevant financial and economic policy areas.

\paragraph{Bush \& Clayton (2023), \textit{Facing Change: Gender and Climate Change
Attitudes Worldwide}.} The article argues that the gender gap in climate concern widens with national wealth
because men in richer countries perceive greater costs from climate mitigation. The text
analysis draws on open-ended survey responses, collected in an original ten-country survey,
in which respondents describe the personal consequences they associate with acting on
climate change. A structural topic model recovers recurring topics -- among them denying
any personal cost, anticipating higher taxes or prices, and pointing to environmental
benefits. The topics are used at the response level, with the prevalence of each estimated
directly as a function of respondent gender and national income. We reanalyze the
material-costs topic, which most directly operationalizes the paper's proposed mechanism.

\paragraph{Clark \& Dolan (2021), \textit{Pleasing the Principal: U.S. Influence in World Bank Policymaking.}} The paper examines how World Bank policy conditionality reflects the preferences of powerful states, especially the United States. It shows that countries politically aligned with the U.S. receive fewer and less stringent policy conditions. The mechanism is primarily indirect, with World Bank staff designing programs consistent with U.S. preferences rather than explicit bargaining. The data consist of World Bank Development Policy Financing (DPF) loan conditions from 2005–2018, including detailed text and categorical coding of policy requirements. Each observation captures policy conditions (prior actions) imposed on borrowing countries as part of loan agreements.

\paragraph{Feltovich \& Giovannoni (2024), \textit{Campaign Messages, Polling, and
Elections}.}
The article examines how candidates' track records and campaign messaging jointly affect
voter welfare, pairing a formal model with a laboratory experiment. The texts are campaign
messages written by experimental subjects playing incumbent and challenger politicians in
repeated elections. Coders classify each message along five dimensions that may co-occur:
positive self-promotion, negative campaigning against the opponent, claims about one's own
competence, claims about past performance, and promises about future performance. These
message-level labels are used directly as dependent variables in regressions on the
experimental conditions. We reanalyze negative campaigning, the behavior behind the
paper's sharpest prediction -- that challengers attack incumbents more as incumbent
performance worsens.

\paragraph{Gilardi et al. (2021), \textit{Policy Diffusion: The Issue-Definition Stage}.}
The paper studies policy diffusion at the issue-definition stage, asking whether  prior smoking-ban adoptions in other U.S. states predict how later states frame the issue.  Using structural topic models, it finds that prior adoptions predict practical, observable  frames such as rules, enforcement, bars and restaurants, casinos, and local legislation,  but not normative frames such as freedom. The paper concludes that diffusion shapes  policymaking before adoption by affecting how issues are publicly defined. The data are newspaper paragraphs about anti-smoking laws from 49 newspapers covering  49 U.S. states between 1996 and 2013. The final corpus contains 52,675 relevant paragraphs  after broad keyword retrieval and filtering with crowd annotation and machine-learning  classification.

\paragraph{Jung (2020), \textit{The Mobilizing Effect of Parties' Moral Rhetoric}.}
The article asks whether parties' use of moral rhetoric mobilizes their own supporters.
The texts are party manifestos from six English-speaking democracies (Australia, Canada,
Ireland, New Zealand, the United Kingdom, and the United States), drawn from the
Comparative Manifesto Project and divided into quasi-sentences. Each quasi-sentence is
classified as moral or non-moral adapting from the Moral Foundations Dictionary, which flags
language invoking concerns such as care, fairness, loyalty, authority, and sanctity. These
document-level labels are not used directly: they are aggregated to the manifesto level as
the share of moral quasi-sentences, which serves as an independent variable predicting
copartisan turnout. We reanalyze the moral/non-moral distinction itself, the construct on
which the paper's entire argument rests.

\paragraph{Lacombe (2019), \textit{The Political Weaponization of Gun Owners}.} Lacombe studies why the NRA has been unusually effective at mobilizing gun owners and influencing U.S. gun politics. The paper argues that the NRA cultivated a politicized gun-owner social identity through long-term membership communications and programs, then used perceived threats to that identity to mobilize political action. The conclusion is that this identity-based mobilization is an important, often overlooked channel of interest-group power. The paper analyzes two original text corpora: NRA American Rifleman editorials from 1930-2008 and gun-control-related letters to the editor from four major U.S. newspapers: the New York Times, Arizona Republic, Atlanta Journal-Constitution, and Chicago Tribune. The newspaper letters are ordinary citizens' public arguments about gun control, excluding elites and NRA officials.

\paragraph{Mattingly et al. (2025), \textit{Chinese State Media Persuades a Global
Audience}.}
The article shows, through a nineteen-country survey experiment, that exposure to Chinese
state media moves global audiences toward seeing the ``China model'' as superior to the
American one. The text analysis characterizes the messaging itself, using the descriptions
of videos published by China's CGTN and the United States' ShareAmerica. A topic model
sorts these descriptions into topics, grouped by the authors into portrayals of China's
political model (responsive institutions, competent leadership, and Western dysfunction)
and its economic model (poverty alleviation, infrastructure, and trade and innovation),
alongside other news and cultural content. The topics are used in aggregate, as corpus-level
proportions that describe typical state-media messaging and motivate the choice of
representative videos used as experimental stimuli; they do not enter a regression. We
reanalyze the Western-dysfunction topic, the most clearly defined and substantively
interpretable of the individual topics from our reading.

\paragraph{Osnabrugge et al. (2021), \textit{Playing to the Gallery: Emotive Rhetoric in Parliaments.}} The paper studies when and why politicians use emotive rhetoric in parliamentary debates. It argues that emotive rhetoric is strategically deployed to appeal to voters, especially when speeches reach a broader public audience. Using large-scale data from UK and Irish parliaments, it finds that high-profile debates (e.g., Prime Minister’s Questions) feature significantly more emotive language.
The data consist of nearly one million parliamentary speeches from the UK House of Commons (2001–2019) and a comparable dataset from the Irish Parliament. These are formal legislative speeches delivered in different types of debates that vary in public visibility and audience size.

\paragraph{Parthasarathyetal et al (2019), \textit{Deliberative Democracy in an Unequal World}.} The data are verbatim transcripts and translations of 50 village assembly meetings held on Republic Day 2014 in rural Tamil Nadu, India. Each document is an uninterrupted speech, with speaker gender and position identified; the full corpus contains 1,736 speeches from gram sabha proceedings. A text should be labeled Yes for fund allocation when Discussion of government funds, budgets, allocations, or rupee amounts.

\paragraph{Schub (2022), \textit{Informing the Leader: Bureaucracies and International Crises}.} The paper examines how bureaucratic position shapes the information advisers provide to leaders during international crises. Using over 5,400 advisory texts from U.S. Cold War crises, it argues that bureaucratic specialization affects informational content and uncertainty rather than policy preferences. Foreign policy bureaucracies emphasize political attributes of adversaries and express greater uncertainty, whereas military bureaucracies emphasize military characteristics. The data consist of internal U.S. foreign policy deliberation texts during Cold War international crises, including memoranda, NSC discussions, presidential advisory meetings, and FRUS documents. The corpus contains adviser-level speech acts from senior officials across bureaucracies such as the State Department, Defense Department, CIA, NSC, and Joint Chiefs of Staff.

\paragraph{Thrall (2025), \textit{Informational Lobbying and Commercial Diplomacy.}} The paper examines how informational lobbying shapes bilateral diplomacy, arguing that diplomats prioritize issues emphasized by the interest groups supplying them with information. Using the expansion of American Chambers of Commerce abroad and oral histories from U.S. diplomats, the paper shows that diplomats exposed to active AmChams devote greater attention to commercial diplomacy. The findings suggest that business interests influence foreign policy by subsidizing diplomats’ information environment rather than through coercive lobbying. The paper uses approximately 1,500 oral history interviews conducted with retired U.S. diplomats by the Association for Diplomatic Studies and Training (ADST). The interviews describe diplomats’ experiences across embassy postings worldwide from roughly the 1940s–2000s, including discussions of trade, investment, security, human rights, and bilateral relations.

\begin{table}[!ht]
\centering
\footnotesize
\caption{Summary of Text Classification Tasks for 14 Studies}
\label{app:paper_sum}
\begin{tabular}{L{2.4cm} L{2.3cm} L{2.5cm} L{2.6cm} L{3.3cm}}
\toprule
Paper & Texts (unit) & Classification method & Reanalysis target & Use in original analysis \\
\midrule

Arias (2022) &
UN General Assembly General Debate speech &
Structural topic model &
Climate security &
Analyzed as dependent variable \\

\addlinespace
Blaydes et al.\ (2018) &
Political advice texts &
Structural topic model &
Multiple political content &
Describes topic trends over time \\

\addlinespace
Blumenau and Lauderdale (2018) &
European Parliament roll-call voting records &
Structural topic model &
Finances &
Distinguishes crisis-related votes to estimate how crises altered voting cleavages \\

\addlinespace
Bush \& Clayton (2023) &
Open-ended survey responses (response) &
Structural topic model &
Material-costs topic &
Used directly; topic prevalence modeled on gender and income \\

\addlinespace
Clark \& Dolan (2021) &
World Bank policy conditionality &
Pre-existing categories &
Fiscal policy &
Reanalyzed as dependent variable \\

\addlinespace
Feltovich \& Giovannoni (2024) &
Lab campaign messages (message) &
Human coders, majority of three &
Negative campaigning &
Used directly; dependent variable on experimental conditions \\

\addlinespace
Gilardi et al.\ (2021) &
Newspaper paragraphs on smoking restrictions &
Structural topic modeling &
Smoking ban enforcement &
Measures issue framing to examine how policy diffusion influences the definition of policy problems before adoption \\

\bottomrule
\end{tabular}
\end{table}

\newpage

\begin{table}[!ht]
\ContinuedFloat
\centering
\footnotesize
\caption[]{Summary of Text Classification Tasks for 14 Studies (continued)}
\begin{tabular}{L{2.4cm} L{2.3cm} L{2.5cm} L{2.6cm} L{3.3cm}}
\toprule
Paper & Texts (unit) & Classification method & Reanalysis target & Use in original analysis \\
\midrule

\addlinespace
Jung (2020) &
Party manifestos (quasi-sentence) &
Moral Foundations dictionary &
Moral vs.\ non-moral rhetoric &
Aggregated to party level; independent variable for turnout \\

\addlinespace
Lacombe (2019) &
NRA and gun-rights advocacy communications &
Structural topic modeling &
Gun regulation &
Measures how advocacy organizations construct gun-owner identity; descriptive \\

\addlinespace
Mattingly et al.\ (2025) &
State-media video descriptions (description) &
LDA topic model &
Western dysfunction topic &
Aggregated to corpus proportions; descriptive, not included in a regression \\

\addlinespace
Osnabrugge et al.\ (2021) &
Parliamentary debates &
Supervised learning &
Emotive vs.\ non-emotive &
Reanalyzed as dependent variable \\

\addlinespace
Parthasarathy et al.\ (2019) &
Village assembly transcripts &
Structural topic modeling &
Fund allocation &
Measures who speaks, sets the agenda, and is heard in deliberative forums \\

\addlinespace
Schub (2022) &
Foreign-relations speech (memo) &
Supervised learning &
Political content &
Tests whether political departments convey more political-content advice \\

\addlinespace
Thrall (2025) &
Diplomat interview scripts &
Supervised learning &
Commercial issues &
Reanalyzed as dependent variable \\

\bottomrule
\end{tabular}
\end{table}

\subsection{Summary of Each Study (Validation Studies in Appendix~\ref{sec:validation_codebook})}\label{app:eachstudy_validation}

{

This section summarizes the three studies used in the validation analyses in Appendix~\ref{sec:validation_codebook}. For each study, we describe the original research and text data, the annotation procedure and coding instructions, and the classification task reproduced in our reanalysis. All three validation studies use 100 randomly sampled texts with original study's human annotations.

\medskip

\noindent \textbf{Benoit et al. (2016)}: 
\citet{benoit_crowd-sourced_2016} examine whether non-expert crowd workers can reproduce expert assessments of political parties' policy positions. Using 18,263 sentences from the Conservative, Labour, and Liberal Democrat manifestos across six British general elections, the authors compare policy positions estimated from crowd-sourced annotations with those derived from expert coding. They find that aggregated crowd annotations closely approximate expert estimates of parties' economic and social policy positions.

The original annotation procedure consists of two steps. First, coders classify each sentence as primarily concerning economic policy, social policy, or neither. Second, sentences classified as economic or social are placed on a five-point left--right scale. The authors provide detailed coding instructions in Appendix Section 6, which were administered to both experts and crowd workers. The expert annotations were produced by six coders, comprising three of the paper's authors and three PhD students in politics at New York University.

In our reanalysis, we use the original sentences as the unit of classification, providing the two preceding and two following sentences as context, consistent with the original crowd-coding procedure. We randomly sample 100 sentences from the 5,444 sentences annotated by all six expert coders. We focus on the first classification task, assigning each sentence to one of three categories: economic policy, social policy, or neither. We therefore omit the subsequent left--right scaling task and make only minimal adjustments to the original codebook, such as specifying the input and output formats for LLM annotation. We compare our annotations with the six individual expert annotations.
\medskip

\noindent \textbf{Müller and Proksch (2024)}: \citet{muller2024nostalgia} develop and validate text-based measures of nostalgic rhetoric in political communication. Using 379 party manifestos from 24 European countries, the authors compare several approaches to measuring nostalgia, including dictionaries, word embeddings, sentiment analysis, and supervised machine learning. They develop the PolNos dataset and use it to examine patterns of nostalgic rhetoric across European political parties, including differences between conservative and progressive parties.

The original annotation procedure uses translated manifesto sentences, with surrounding sentences provided as context. Four human coders apply the detailed coding instructions in Appendix Section C to identify nostalgic rhetoric. Following the definition adopted by the authors, nostalgia refers to a predominantly positive emotion associated with recalling important or meaningful past experiences. The codebook first asks coders to determine whether a sentence is nostalgic and then includes additional tasks concerning translation quality and the cultural or economic dimensions of nostalgia.

In our reanalysis, we use the processed, translated sentence-level texts provided in the authors' replication data, including the preceding and following sentences as context. We randomly sample 100 sentences from those annotated by the original four human coders. We focus on the primary binary classification task, determining whether each sentence contains nostalgic rhetoric, without requiring additional classifications of translation quality or nostalgia subtypes. We preserve the original definitions, classification instructions, and examples, making only minimal adjustments to the required output format for LLM annotation. We compare our annotations with the four individual human annotations.

\medskip

\paragraph{BFRS Dataset on Political Violence in Pakistan (Bueno de Mesquita et al. 2015; Halterman and Keith 2026).}
The BFRS Political Violence in Pakistan Dataset, originally developed by \citet{bueno2015measuring}, documents 28,731 incidents of political violence in Pakistan between 1988 and 2011. The dataset covers a broad range of violent events, including assassinations, terrorism, riots, violent political demonstrations, and attacks involving state and non-state actors. The original data were collected by researchers at Lahore University of Management Sciences, who systematically reviewed newspaper reports from the Lahore edition of \textit{The Dawn}. Human coders identified incidents of political violence and recorded their characteristics, including the type of violence, location, perpetrators, and consequences. Each incident was also accompanied by a short narrative summary. The original researchers additionally conducted a reliability assessment using independent coding by a separate research team.

More recently, \citet{Halterman2026CodebookLLMs} adapted the BFRS dataset to evaluate whether LLMs can apply detailed political science codebooks. Because the original newspaper articles were unavailable, they use the English-language event summaries written by the original BFRS annotators as classification inputs. They restructure the original coding instructions into a 12-category, single-label classification task. Each category is accompanied by a definition, clarification and exclusion rules, and illustrative examples. The task requires assigning each event to one category, such as assassination, terrorism, riot, or violent political demonstration.

In our reanalysis, we randomly sample 100 event summaries from the adapted dataset for which original human-coded labels are available. We retain the 12-category classification task and use the detailed codebook developed by \citet{Halterman2026CodebookLLMs}, making only minimal adjustments to the required output format for LLM annotation. We compare our annotations with the original BFRS human-coded event labels. Unlike the other two validation studies, however, the available BFRS data contain only one reference label per event. Consequently, we cannot directly measure agreement among the original human coders.

}

\subsection{Details of Codebook Generation}\label{app:prompt_all}

Because most of the replicated studies do not publish the codebooks used in
their original annotation, we construct a codebook for each study from the
descriptions in the published article, using GPT-5.5 (accessed through
ChatGPT) as a drafting aid. We first read each article in full to understand
the concept of interest and how it is used in the analysis. We then upload
the article as a PDF to GPT-5.5, together with the prompt below, and ask the
model to draft the paper-specific components of the codebook. Finally, we
check every generated component against the article and revise any content
that is inaccurate, incomplete, or inconsistent with how the original authors
define and use the concept. Thus, every component of the final codebooks has
been verified against the source article. The prompt is as follows:

\begin{quote}
We are developing a prompt for an LLM to classify texts used in a political science paper. I have attached the paper as a PDF. Our goal is to classify whether a text should be categorized as {category name} as defined and used in the paper's analysis. Please create a high-quality prompt for this task. Specifically, I want you to provide the following components:

\textbf{Definition:} Provide a definition of the target category in no more than 10 words. Whenever possible, base the definition on how the concept is used in the attached paper.

\textbf{Classification rules:} Provide a classification rule of no more than 10 words for each of the following:

when the label should be Yes,
when the label should be No, and
what kinds of cases are tricky or borderline.

\textbf{Label wording:} Provide five keywords and related expressions for the target category. Include alternative label names, synonyms, and close paraphrases when they are semantically relevant. Keep this concise while covering the major cases.

\textbf{Summary of the paper:} Provide a summary of the paper in no more than three sentences, covering the research question, context, and main findings.

\textbf{Summary of the data:} Provide a summary of the text data in no more than two sentences. Describe the source and context of the texts (e.g., speeches in the United Nations General Assembly or parliamentary speeches in India). Focus on the data itself rather than how it is analyzed."
\end{quote}
See Table~\ref{prompt_component} for the justification for each component. We fix the contents of prompts across papers to ensure that only the paper-specific components vary and thus the results are comparable across papers. See Table~\ref{tab:text_classification_prompts} for the components used for each paper.
%: wherever possible we replaced or corrected the definition and examples using text taken directly from the paper or its appendix, and, for papers using a structural topic model, we anchored the definition and label wording to the model's high-probability keywords. 
%When the article did not supply a usable definition, we substituted one from the Merriam-Webster or Oxford Dictionary, and when no satisfactory definition could be obtained even then, we replaced the topic with an alternative, as described in our topic selection. The verified components were stored in a paper-specific configuration file that served as the single source of truth for all subsequent prompt construction.

%For example, \cite{Schub2022}, whose topic is \emph{adversary domestic politics}, this procedure returned the definition ``Discussion of adversary domestic politics and political resolve''; the classification rules ``Yes: focuses on domestic politics, resolve, governance, or political outcomes / No: focuses mainly on military capabilities, operations, or force movements / Borderline: political consequences of military actions''; and the label wording \{political content, political attributes, domestic political conditions, resolve and political will, governance and regime dynamics\}, together with three-sentence and two-sentence summaries of the paper and its corpus of U.S.\ Cold War advisory documents.

\begin{table}[!hp]
\centering
\caption{General Prompt Components and Their Justification (Codebook-Style) }
\label{prompt_component}
\begin{tabular}{p{3.5cm} p{5cm} p{6.5cm}}
\toprule
\textbf{Component} & \textbf{Role in Prompt} & \textbf{Justification} \\
\midrule

Role & 
Assigns the model a specific analytical perspective &
Directs the model to adopt the relevant domain expertise and evaluative stance, improving task alignment and reducing inconsistent interpretations. \\

Definition & 
Provides a formal conceptualization of the construct &
Aligns the model with the intended theoretical meaning, ensuring consistency with the research concept rather than colloquial usage. \\

Classification Rules (Yes / No / Tricky Cases) & 
Specifies inclusion (Yes), exclusion (No), and ambiguous (Tricky) conditions for classification &
Operationalizes the abstract concept into an explicit decision boundary: inclusion rules define positive cases, exclusion rules prevent category leakage, and tricky cases guide decisions under ambiguity. This structured separation improves consistency, reduces false positives, and enhances robustness in edge cases. \\

Label Wording & 
Defines the vocabulary of allowable outputs &
Standardizes the semantic space of labels, ensuring consistency across observations and enabling aggregation and downstream analysis. \\

Summary of Paper/Data & 
Provides contextual information about the substantive setting and the dataset &
Anchors interpretation in the research context, helping the model infer relevance when signals are indirect or mechanism-based rather than explicit. 
Clarifies the distribution of topics and potential co-occurrence of issue areas, helping calibrate expectations and avoid overgeneralization.\\

Reasoning & 
Specifies the inferential process for classification &
Encourages the model to evaluate the text against the definition systematically and articulate the basis for its decision, improving transparency and conceptual consistency. \\

\bottomrule
\end{tabular}
\end{table}

\newcolumntype{L}[1]{>{\raggedright\arraybackslash}p{#1}}

% Table
\footnotesize
\setlength{\LTleft}{0pt}
\setlength{\LTright}{0pt}

\begin{longtable}{L{2.4cm} L{12.8cm}}
\caption{Text classification across the studies}
\label{tab:text_classification_prompts}\\

\toprule
Study & Prompt \\
\midrule
\endfirsthead

\multicolumn{2}{l}{\footnotesize\textit{Table \thetable{} continued}}\\
\toprule
Study & Prompt \\
\midrule
\endhead

\midrule
\multicolumn{2}{r}{\footnotesize\textit{Continued on next page}}\\
\endfoot

\bottomrule
\endlastfoot

Arias  \ (2022) &
\textbf{Topic}: Climate security

\textbf{Definition}: A text should be labeled Yes for climate security when Climate change is framed as an international security threat.

\textbf{Classification Rules}

Yes - Climate described as security, conflict, instability, or UNSC issue.

No - Climate discussed without security, conflict, threat, or UNSC framing.

Tricky/borderline - Existential harm or sea-level threats without security jurisdiction.

\textbf{Label Wording}

Climate security, Climate securitization, Climate as security threat, Threat to international peace and security, Climate-linked conflict, instability, terrorism, or state failure

\textbf{Summary of Paper}

Arias asks why some states frame climate change as a security issue in the United Nations, and which states are most likely to make such securitizing moves. The paper argues that securitization is shaped by agenda-control incentives: P5 states benefit from moving issues toward the Security Council, while highly vulnerable states such as SIDS may resist losing agenda control. Empirically, the paper finds that UN climate discourse overall is not securitized, P5 states are more likely to securitize climate change, and SIDS are less likely to do so.

\textbf{Summary of Data}

The text data are speeches by state representatives in the UN General Assembly General Debate, filtered to speech segments discussing climate change. The analysis uses 4,525 climate-related speech segments from 1,987 speeches, with the earliest climate-related speech segment appearing in 1984. \\

\midrule

Blaydes et al.\ (2018) & \textbf{Topic}: Art of Rulership

\textbf{Definition}  

Practical guidance on governing, exercising power, and ruling effectively.

\textbf{Classification Rules } 

YES - Describes ruler’s duties, governance practices, or statecraft strategies.

NO - Focuses on religion, personal morality, or non-political life.

Tricky / borderline - Moral advice tied directly to governing decisions or authority.

\textbf{Label Wording }

Statecraft,
Governance advice,
Ruling practices,
Political management,
Exercise of power / kingship guidance

\textbf{Summary of Paper  }

The paper compares medieval Christian and Islamic political advice texts (“mirrors for princes”) to examine how ideas of governance evolved across regions. Using text-as-data methods, it identifies shared thematic structures and tracks their evolution over time. It finds broad similarities across traditions but key divergences, including declining religious emphasis in Europe and shifting ruler-centered discourse in the Islamic world.

\textbf{Summary of Data }

The data consist of translated medieval political advice texts addressed to rulers in Christian Europe and the Islamic world. These texts include guidance on governance, morality, religion, and social order, often written by elites for kings, courts, or political audiences.
\\

\midrule

Bush \& Clayton (2023) & \textbf{Topic}: taxes

\textbf{Definition}
Tax-related material costs of climate action.

\textbf{Classification rules}

Yes -Mentions taxes, fees, prices, or cost increases.

No - Mentions non-tax harms, benefits, or general climate concern.

Tricky/borderline - Code broad “cost” claims only if tax-like or monetary.

\textbf{Label wording}

Taxes,
Tax burden,
Carbon tax / fossil-fuel tax,
Fees, penalties, mandatory payments,
Higher costs / cost-of-living increases

\textbf{Summary of Paper}

Bush and Clayton study why the gender gap in climate concern is larger in wealthier countries. They argue that as countries become wealthier, climate mitigation is perceived as more costly and less beneficial, especially among men, who are more likely to associate decarbonization with material and psychological costs. Using cross-national surveys, an original 10-country survey, and focus groups in Peru and the United States, they find that men’s climate concern declines more sharply with economic development and that men in wealthier countries more often emphasize costs of climate action.

\textbf{Summary of Data}

The relevant text data are open-ended survey responses from citizens in 10 countries in the Americas and Western Europe, collected through Netquest in 2019–2020, asking how acting to stop climate change might personally harm respondents. The “Taxes” category appears in a structural topic model of these open-ended responses; the paper labels Topic 2 as “Taxes,” associated with words such as “increase,” “tax,” and “cost,” and finds it more common in wealthier countries.
 \\

\midrule

Clark \& Dolan (2021) & \textbf{Topic}: fiscal policy

\textbf{Definition}

Government taxation, spending, debt, and budgetary policy actions

\textbf{Classification rules}

Yes - Mentions taxes, spending, deficits, debt, or fiscal measures

No - Discusses non-fiscal policies (e.g., monetary, foreign, social)

Tricky / borderline - Economic discussion without explicit fiscal instruments or government action

\textbf{Label wording}

fiscal policy / budget policy,
taxation and government spending,
deficit, debt, public finance,
revenues, expenditures, austerity,
government budget management

\textbf{Summary of Paper}

The paper examines how World Bank policy conditionality reflects the preferences of powerful states, especially the United States. It shows that countries politically aligned with the U.S. receive fewer and less stringent policy conditions. The mechanism is primarily indirect, with World Bank staff designing programs consistent with U.S. preferences rather than explicit bargaining.

\textbf{Summary of Data}

The data consist of World Bank Development Policy Financing (DPF) loan conditions from 2005–2018, including detailed text and categorical coding of policy requirements. Each observation captures policy conditions (prior actions) imposed on borrowing countries as part of loan agreements.
\\

\midrule

Feltovich \& Giovannoni (2024) & \textbf{Topic}: Contains negative campaigning

\textbf{Definition:} Criticizes opponent’s behavior, quality, or performance.

Yes - Attacks opponent’s behavior, quality, competence, or record.

No - Promotes sender without criticizing the opponent.

Tricky/borderline - Neutral opponent references without criticism are borderline.

\textbf{Label wording}

attack; criticism; opponent failure; poor performance; low quality

\textbf{Summary of Paper}

Feltovich and Giovannoni study how politicians’ track records and campaign messages interact to shape elections and voter welfare. They develop and test a laboratory election model where incumbents and challengers vary in quality, voters observe performance imperfectly, and candidates may send campaign messages before elections. The paper finds that both higher quality variability and campaigning improve voter welfare, while challengers are more likely to use negative campaigning when incumbents perform poorly.

\textbf{Summary of Data}

The text data are short, free-form campaign announcements written by candidates in a laboratory election experiment after a straw poll and before the final election. Messages were 0–140 characters, written in English, and sent by incumbents and challengers; research assistants coded whether each message was positive, negative, about quality, about past performance, or about future decisions.

\\

\midrule

Gilardi et al. \ (2021) &
\textbf{Topic}: Smoking ban enforcement

\textbf{Definition}: A text should be labeled Yes for smoking ban enforcement when Implementation and legal enforcement of smoking restrictions.

\textbf{Classification Rules}

Yes - Discusses compliance, penalties, lawsuits, or enforcement of smoking bans.

No - Discusses smoking bans without enforcement or compliance issues.

Tricky/borderline - Mentions rules, unless implementation or penalties are central.

\textbf{Label Wording}

enforcement,
compliance,
penalties or fines,
lawsuits or legal challenges,
implementation

\textbf{Summary of Paper} 

The paper studies policy diffusion at the issue-definition stage, asking whether  prior smoking-ban adoptions in other U.S. states predict how later states frame the issue.  Using structural topic models, it finds that prior adoptions predict practical, observable  frames such as rules, enforcement, bars and restaurants, casinos, and local legislation,  but not normative frames such as freedom. The paper concludes that diffusion shapes  policymaking before adoption by affecting how issues are publicly defined.

\textbf{Summary of Data}

The data are newspaper paragraphs about anti-smoking laws from 49 newspapers covering  49 U.S. states between 1996 and 2013. The final corpus contains 52,675 relevant paragraphs  after broad keyword retrieval and filtering with crowd annotation and machine-learning  classification. \\

\midrule

Jung (2020) &
\textbf{Topic}: Moral rhetoric

\textbf{Definition}: Frames politics as moral right, wrong, virtue, or transgression.

\textbf{Classification Rules}

Yes -- Uses moral values to justify political claims.

No -- Only factual, instrumental, or policy-pragmatic language.

Tricky / Borderline -- Moral words used as policy labels or proper nouns.

\textbf{Label Wording}

Moral appeal --- ethical appeal, moral framing

Right and wrong --- morally right, morally wrong

Virtue/transgression --- dignity, justice, fairness, harm

Moral foundations --- care, fairness, loyalty, authority, sanctity

Values-based rhetoric --- principles, duties, rights, common good

\textbf{Summary of Paper}

The paper asks whether parties' use of moral rhetoric affects voter behavior, focusing on party campaign communication in six English-speaking democracies. Jung argues that moral rhetoric activates positive emotions among copartisans who are exposed to party rhetoric, especially politically aware voters. The paper finds that higher moral rhetoric in party manifestos increases turnout among more educated copartisans, and that survey experiments and British Election Study data support the positive-emotion mechanism.

\textbf{Summary of Data}

The core text data are party manifestos from the Manifesto Corpus, covering 64 party manifestos from six English-speaking democracies across 18 elections. The unit of text classification is the quasi-sentence, a policy statement that may be a full sentence or part of a sentence. The appendix also validates the manifesto-based measure against a smaller set of campaign speech texts from Australia and New Zealand.
\\

\midrule

Lacombe (2019) & \textbf{Topic}: Gun regulation

\textbf{Definition}: A text should be labeled Yes for gun control when Debates over firearm laws, restrictions, ownership, and gun rights.

\textbf{Classification Rules}

Yes - Text discusses firearm policy, laws, regulation, or gun rights.

No - Text discusses guns without policy, regulation, or rights debate.

Tricky/borderline - Identity, crime, or safety claims tied to regulation.

\textbf{Label Wording}

gun control,
firearm regulation,
gun laws,
gun rights,
Second Amendment / firearms legislation

\textbf{Summary of Paper}

Lacombe studies why the NRA has been unusually effective at mobilizing gun owners and influencing U.S. gun politics. The paper argues that the NRA cultivated a politicized gun-owner social identity through long-term membership communications and programs, then used perceived threats to that identity to mobilize political action. The conclusion is that this identity-based mobilization is an important, often overlooked channel of interest-group power.

\textbf{Summary of Data }

The paper analyzes two original text corpora: NRA American Rifleman editorials from 1930-2008 and gun-control-related letters to the editor from four major U.S. newspapers: the New York Times, Arizona Republic, Atlanta Journal-Constitution, and Chicago Tribune. The newspaper letters are ordinary citizens' public arguments about gun control, excluding elites and NRA officials.
\\

\midrule

Mattingly et al.\ (2025) &  \textbf{Topic}: Western dysfunction 

\textbf{Definition}
Claims Western democracies suffer instability, racism, violence, or failure.

\textbf{Classification rules}

Yes - Criticizes Western political/social disorder or democratic failure.

No - Discusses China’s strengths without criticizing the West.

Tricky/borderline - Comparisons count only with explicit Western dysfunction.

\textbf{Label wording}

Western dysfunction,
Western political failure,
U.S./Western instability,
Racism, protests, or political violence in the West,
Democratic disorder / liberal-democratic crisis

\textbf{Summary of Paper}

Mattingly et al. study whether Chinese and American state media can persuade global audiences to prefer China’s authoritarian model or the U.S. democratic model. The paper combines content analysis of Chinese and American external media with a randomized survey experiment in 19 countries. It finds that Chinese state media, especially messages emphasizing performance, growth, stability, and competent governance, substantially increases support for the Chinese model, while U.S. messaging is less persuasive.

\textbf{Summary of Data}

The relevant text data are descriptions of external state-media video segments, especially CGTN videos posted on YouTube from 2020–2021 and ShareAmerica videos produced by the U.S. Department of State. For “Western dysfunction,” the category comes from the paper’s topic-model analysis of CGTN videos promoting China’s political model, where this theme includes stories about protests, racism, and political violence in the United States that contrast Western disorder with Chinese stability and responsiveness.
 \\

\midrule

Osnabrugge et al.\ (2021) & \textbf{Topic}: Emotive Rhetoric

\textbf{Definition}

Language eliciting emotional response beyond literal policy content

\textbf{Classification rules}

Yes- Uses emotionally charged words to evoke feelings or reactions

No - Neutral, technical, or purely informational without emotional tone
 
Tricky / borderline - Strong language without clear emotional appeal or intent

\textbf{Label wording}

emotional appeal / emotive language,
affective rhetoric / loaded language,
emotionally charged / expressive tone,
fear, outrage, admiration framing,
persuasive emotional signaling

\textbf{Summary of Paper}

The paper studies when and why politicians use emotive rhetoric in parliamentary debates. It argues that emotive rhetoric is strategically deployed to appeal to voters, especially when speeches reach a broader public audience. Using large-scale data from UK and Irish parliaments, it finds that high-profile debates (e.g., Prime Minister’s Questions) feature significantly more emotive language.

\textbf{Summary of Data}

The data consist of nearly one million parliamentary speeches from the UK House of Commons (2001–2019) and a comparable dataset from the Irish Parliament. These are formal legislative speeches delivered in different types of debates that vary in public visibility and audience size.
\\

\midrule

Parthasarathyetal et al. (2019) & \textbf{Topic}: Fund allocation

\textbf{Definition}: A text should be labeled Yes for fund allocation when Discussion of government funds, budgets, allocations, or rupee amounts.

\textbf{Classification Rules}

Yes - Mentions allocating, approving, spending, or reporting public funds.

No - Mentions services without money, budgets, or fund distribution.

Tricky/borderline - Schemes count only when funding or amounts are discussed.

\textbf{Label Wording}

Fund allocation,
Allocation of funds,
Budget / public expenditure,
Government funds / panchayat funds,
Rupee amounts / allotted funds / scheme funding

\textbf{Summary of Paper} 

The paper studies deliberation in constitutionally mandated village assemblies, or gram sabhas, in rural Tamil Nadu, India, using text-as-data methods on meeting transcripts. It asks whether these assemblies provide meaningful opportunities for citizens to speak, set the agenda, and receive responses from officials, especially under conditions of gender inequality. The authors find that the assemblies are not merely talking shops, but women are less likely than men to speak, shape discussion, or receive relevant official responses, while female village presidents improve women's likelihood of being heard.

\textbf{Summary of Data }

The data are verbatim transcripts and translations of 50 village assembly meetings held on Republic Day 2014 in rural Tamil Nadu, India. Each document is an uninterrupted speech, with speaker gender and position identified; the full corpus contains 1,736 speeches from gram sabha proceedings.
\\

\midrule

Schub (2022) & \textbf{Topic}: political content (vs. military)

\textbf{Definition}
Discussion of adversary domestic politics and political resolve.

\textbf{Classification Rules}

Yes - Focuses mainly on opponent’s political situations, such as:

-domestic political landscape:  the configuration of domestic actors, institutions, and political constraints that shape a state’s policy choices and strategic behavior.

-difficulties of converting battlefield outcomes to desired political end states.

-adversary resolve: the willingness of an opponent to continue fighting in pursuit of its objectives.

No - (military, only two categories in total)

Focuses mainly on military capabilities, quality of forces, force operations, force movements, or force locations.

Tricky / Borderline

-according to the paper, we should code “yes” if the text’s focus or objective is about political content but mentioning its causes due to other factors such as military. That is, when it’s not directly about military actions.

-mixed operational-political assessments: sometimes it may be hard to distinguish whether political content is the focus when it co-appears with other factors such as military, it may be hard to code. 

\textbf{Label Wording}

Political content,
Political attributes,
Domestic political conditions,
Resolve and political will,
Governance and regime dynamics

\textbf{Summary of Paper}

The paper examines how bureaucratic position shapes the information advisers provide to leaders during international crises. Using over 5,400 advisory texts from U.S. Cold War crises, it argues that bureaucratic specialization affects informational content and uncertainty rather than policy preferences. Foreign policy bureaucracies emphasize political attributes of adversaries and express greater uncertainty, whereas military bureaucracies emphasize military characteristics.

\textbf{Summary of Data}

The data consist of internal U.S. foreign policy deliberation texts during Cold War international crises, including memoranda, NSC discussions, presidential advisory meetings, and FRUS documents. The corpus contains adviser-level speech acts from senior officials across bureaucracies such as the State Department, Defense Department, CIA, NSC, and Joint Chiefs of Staff.
\\

\midrule

Thrall (2025) & \textbf{Topic}: commercial issues

\textbf{Definition}

Commercial diplomacy and cross-border business-related diplomatic issues.

\textbf{Classification rules}

Yes - Discusses trade, investment, firms, taxation, or commercial regulation.

No - Focuses solely on security, military, humanitarian, or cultural issues.

Tricky / Borderline - Economic development discussion without explicit business or trade relevance.

\textbf{Label wording}

trade policy / trade relations,
investment promotion / foreign investment,
business regulation / commercial regulation,
taxation / tariffs / market access,
exports, firms, commerce, industry, intellectual property

\textbf{Summary of Paper}

The paper examines how informational lobbying shapes bilateral diplomacy, arguing that diplomats prioritize issues emphasized by the interest groups supplying them with information. Using the expansion of American Chambers of Commerce abroad and oral histories from U.S. diplomats, the paper shows that diplomats exposed to active AmChams devote greater attention to commercial diplomacy. The findings suggest that business interests influence foreign policy by subsidizing diplomats’ information environment rather than through coercive lobbying.

\textbf{Summary of Data}

The paper uses approximately 1,500 oral history interviews conducted with retired U.S. diplomats by the Association for Diplomatic Studies and Training (ADST). The interviews describe diplomats’ experiences across embassy postings worldwide from roughly the 1940s–2000s, including discussions of trade, investment, security, human rights, and bilateral relations.
\\

\end{longtable}

\normalsize

\subsection{Updated Codebook}\label{app:complete_codebook}

We use the updated codebooks for Section~\ref{sec3.2.3} and
Appendix~\ref{sec:ambiguity_application}. For each paper, one of our
authors crafted a more detailed
codebook based on the description beyond our template described in Table~\ref{prompt_component}. Because these updated prompts have study-specific structures,
we present the complete prompts separately rather than in tabular form.
For Arias, the full-corpus analysis in Appendix~\ref{sec:ambiguity_application}
uses the substantively refined prompt shown here rather than the standardized
prompt~64 used for the main 100-text replication; the appendix therefore
targets annotations under the refined codebook. This implements the
refinement-before-scaling workflow described in Section~\ref{step1}.

\newpage
\noindent\textbf{\large Arias (2022)}

\medskip

\begin{tcolorbox}[
    breakable,
    colback=white,
    colframe=black!40,
    boxrule=0.4pt,
    arc=0pt,
    left=6pt,
    right=6pt,
    top=6pt,
    bottom=6pt
]
\small
\raggedright

Your role is an human classifier of textual topics from a political
science journal article.

\medskip

Carefully read the text provided below and determine whether the text
belongs to the topic of ``climate security'' or not. Before you start,
carefully review the following instructions.

\medskip

\textbf{Codebook for Climate Security}

\medskip

\textbf{Definition:} A text should be labeled Yes when it securitizes
climate change or the environment.

\medskip

This definition is met if any of the following conditions are present:

\begin{enumerate}[label=(\arabic*), leftmargin=2em, itemsep=2pt, topsep=2pt]
    \item climate change or the environment is directly described as a
    security issue;
    \item climate change or the environment is directly compared to
    traditional security issues, such as war, conflict, or terrorism; or
    \item climate change or the environment is presented as an
    existential threat that requires emergency or urgent measures.
\end{enumerate}

A text should not be considered securitizing when security terms or
issues are discussed separately from climate change or the environment.

\medskip

\textbf{Classification Rules}

\medskip

\textbf{Yes} --- Label Yes if the text explicitly frames climate change
or the environment as a security threat; links it to conflict, violence,
war, terrorism, military concerns, or instability; or presents it as a
direct existential threat requiring urgent or emergency action.

\medskip

\textbf{Positive example 1:}

``Of the major issues confronting us today there is probably none that
has captured our imagination more than the deterioration of the natural
environment. The greenhouse effect, global warming, the depletion of
the ozone layer, acid rain, waste dumping, forest depletion, and
drift-net fishing, threaten our existence. The greenhouse effect and
the resulting global warming and the rise in sea level constitute a
direct threat, no longer dismissible, to our survival. We in the
Pacific have more cause than many to be deeply concerned about it since
if the rise in sea level is significant some of our islands and coastal
areas may become permanently inundated. The possibility that entire
countries may also drown is almost beyond comprehension.''

\medskip

\textbf{Positive example 2:}

``Alone we can do little, but together we can achieve much. The
challenges that we face globally, from climate change to refugees to war
and violence, require urgent action now. The world does not have the
luxury of time. We have talked. We have debated. We have postulated and
hypothesized. We have studied and analysed. We must now act.''

\medskip

\textbf{No} --- Label No when climate change or the environment is
discussed without a security, conflict, threat, or urgent existential
framing. Also label No when security-related terms are present but refer
to separate issues rather than climate change or the environment.

\medskip

\textbf{Negative example:}

``The United Nations Conference on Environment and Development to be
held in 1992, in which so many high expectations have been placed,
offers an opportunity for the international community to address the
global ecological challenges that confront mankind. Any effective,
collective action in the major areas that the Conference is expected to
tackle---whether climate change or biodiversity and genetic
heritage---entails envisaging new methods of funding, such as the
introduction of some form of international taxation.''

\medskip

\textbf{Tricky/borderline} --- Existential, survival, or sea-level
language requires context. Label Yes only when it clearly presents
climate or environmental harm as a direct security or existential
threat and conveys urgency or the need for emergency action. Label No
when it merely describes environmental harm, vulnerability, or physical
impacts without this framing.

\medskip

\textbf{Label Wording}

Climate security

Climate securitization

Climate as a security threat

Threat to international peace and security

Climate-linked conflict, instability, terrorism, violence, or state failure

Security, military, war, conflict, instability, unrest, violence,
emergency, urgent, sovereignty, survival, existential

\medskip

\textbf{Note:} These terms are diagnostic only when they are
substantively connected to climate change or the environment; their
presence alone is not sufficient for a Yes label.

\medskip

\textbf{Summary of Paper}

\medskip

Arias asks why some states frame climate change as a security issue in
the United Nations, and which states are most likely to make such
securitizing moves. The paper argues that securitization is shaped by
agenda-control incentives: P5 states benefit from moving issues toward
the Security Council, while highly vulnerable states such as SIDS may
resist losing agenda control. Empirically, the paper finds that UN
climate discourse overall is not securitized, P5 states are more likely
to securitize climate change, and SIDS are less likely to do so.

\medskip

\textbf{Summary of Data}

\medskip

The text data are speeches by state representatives in the UN General
Assembly General Debate, filtered to speech segments discussing climate
change. The analysis uses 4,525 climate-related speech segments from
1,987 speeches, with the earliest climate-related speech segment
appearing in 1984.

\medskip

Let’s think step by step through all the above instructions when you
process each text.

\medskip

In your output, I want you to respond with `Yes' if the text closely
aligns with the definition of climate security, otherwise respond with
`No'. Respond only with `Yes' or `No'. Do not provide any other outputs
or any explanation for your output.

\end{tcolorbox}

\newpage
\noindent\textbf{\large Schub (2022)}

\medskip

\begin{tcolorbox}[
    breakable,
    colback=white,
    colframe=black!40,
    boxrule=0.4pt,
    arc=0pt,
    left=6pt,
    right=6pt,
    top=6pt,
    bottom=6pt
]
\small
\raggedright

Your role is an human classifier of textual topics from a political
science journal article.

\medskip

Carefully read the text provided below and determine whether the text
belongs to the topic described in the classification codebook. Before
you start, carefully review the following instructions.

\medskip

\textbf{Classification codebook}

\medskip

\textbf{Topic: political content}

\medskip

\textbf{Definition}

Discussion of adversary domestic politics and political resolve.

\medskip

\textbf{Classification Rules}

\medskip

\textbf{Yes}

Focuses mainly on opponent's political situations, such as:

\begin{itemize}[leftmargin=1.5em, itemsep=2pt, topsep=2pt]
    \item \textbf{domestic political landscape}: the configuration of
    domestic actors, institutions, and political constraints that shape
    a state's policy choices and strategic behavior.
    
    \item difficulties of converting battlefield outcomes to desired
    political end states.
    
    \item \textbf{adversary resolve}: the willingness of an opponent to
    continue fighting in pursuit of its objectives.
\end{itemize}

\textbf{No}

Focuses mainly on military capabilities, quality of forces, force
operations, force movements, or force locations.

\medskip

\textbf{Tricky / Borderline}

\begin{itemize}[leftmargin=1.5em, itemsep=2pt, topsep=2pt]
    \item according to the paper, we should code ``yes'' if the text's
    focus or objective is about political content but mentioning its
    causes due to other factors such as military. That is, when it's not
    directly about military actions.
    
    \item mixed operational-political assessments: sometimes it may be
    hard to distinguish whether political content is the focus when it
    co-appears with other factors such as military, it may be hard to
    code.
\end{itemize}

\textbf{Paper examples:}

\medskip

\textbf{Military Texts}

\medskip

\textbf{Example 1:} Ground Attacks on Base Camps in Cambodia: Attached
at Tab A is a brief summary of the two options for ground attacks on
enemy base camps in Cambodia submitted by General Abrams on March 30.
In developing plans for potential operations against enemy base areas,
General Abrams was asked to consider two possibilities: An attack
against targets of high military priority which could involve the use of
US forces if necessary. Any other operation which would reduce the
necessity of the involvement of US forces. With respect to military
priority, MACV considered an attack on Base Area 352/353 (COSVN Hq) to
be the most lucrative. He made the following significant points about
this base area. [1]

\medskip

\textbf{Example 2:} The Chiefs believe that ground action against the
North Vietnamese effort is adequate to reverse the situation. Air
strikes on the three targets are not necessary from a military point of
view. However, a South Vietnamese attack on their target is acceptable.
[2]

\medskip

\textbf{Political Texts}

\medskip

\textbf{Example 1:} Iran. The two leading US academic humans on Iran,
James Bill and Marvin Zonis, recently were debriefed in the Department
following their separate visits to Iran at the end of November. In a
wide range of Iranian contacts, both men found intense rage against the
Shah personally. This is a marked change from the past when Iranians
were content to blame their troubles on the Government and the Shah's
advisers. Both professors see a slim chance that the Shah might retain
a minimal role as constitutional monarch, but only if he moves quickly
to negotiate a political compromise. They assess the opposition as very
strong and extremely well-organized. Everywhere they found an eagerness
for the US to play a decisive role in promoting a political solution to
Iran's crisis. [3]

\medskip

\textbf{Example 2:} In spite of economic difficulties there is no solid
evidence that Trujillo's fall is imminent. Trujillo rules by force and
will presumably remain in power as long as the armed forces continue to
support him. While there is evidence of dissatisfaction on the part of
a few officers there is as yet no cogent evidence of large-scale
defection within the officer corps. The underground opposition to
Trujillo composed of business, student and professional people is
believed to be predominantly anti-Communist. They have substantially
increased in numbers in recent years but have been unable to move
effectively against Trujillo. In addition to opposition groups in the
Dominican Republic, there are numerous exile groups located principally
in Venezuela, Cuba, United States and Puerto Rico. In some cases these
groups have been infiltrated by pro-Castro or pro-Communist elements.
[4]

\medskip

\textbf{Label Wording}

\begin{itemize}[leftmargin=1.5em, itemsep=2pt, topsep=2pt]
    \item Political content
    \item Political attributes
    \item Domestic political conditions
    \item Resolve and political will
    \item Governance and regime dynamics
\end{itemize}

\textbf{Summary of Paper}

\medskip

The paper examines how bureaucratic position shapes the information
advisers provide to leaders during international crises. Using over
5,400 advisory texts from U.S. Cold War crises, it argues that
bureaucratic specialization affects informational content and
uncertainty rather than policy preferences. Foreign policy bureaucracies
emphasize political attributes of adversaries and express greater
uncertainty, whereas military bureaucracies emphasize military
characteristics.

\medskip

\textbf{Summary of Data}

\medskip

The data consist of internal U.S. foreign policy deliberation texts
during Cold War international crises, including memoranda, NSC
discussions, presidential advisory meetings, and FRUS documents. The
corpus contains adviser-level speech acts from senior officials across
bureaucracies such as the State Department, Defense Department, CIA,
NSC, and Joint Chiefs of Staff.

\medskip

Let’s think step by step through all the above instructions when you
process each text.

\medskip

In your output, I want you to respond with `Yes' if the text closely
aligns with the definition of political content, otherwise respond with
`No'. Respond only with `Yes' or `No'. Do not provide any other outputs
or any explanation for your output.

\end{tcolorbox}

\normalsize

\subsection{Codebook used in Validation Studies (Appendix~\ref{sec:validation_codebook})}\label{app:validate_codebook}

We use the original detailed codebooks in the validation studies described in Appendix~\ref{sec:validation_codebook} to assess whether our main findings hold when we use the original codebooks rather than those constructed by our team. When adapting these codebooks for LLM annotation, we make only minimal adjustments (e.g., specifying the required output format) to ensure that the models produce valid responses. We otherwise preserve the original coding instructions so that the original annotators, the three authors of this paper, and the LLMs classify texts using substantively identical instructions. We reproduce the codebook for each validation study below. See Appendix~\ref{sec:validation_codebook} for further details on the validation studies.

\medskip

\noindent\textbf{\large\citealt{benoit_crowd-sourced_2016}}

\begin{tcolorbox}[
    breakable,
    colback=white,
    colframe=black!40,
    boxrule=0.4pt,
    arc=0pt,
    left=6pt,
    right=6pt,
    top=6pt,
    bottom=6pt
]
\small
\raggedright

This task involves reading sentences from political texts and judging whether these deal with economic or social policy.

The sentences you will be asked about come from political party manifestos. Some deal with economic policy, some with social policy; others deal with neither economic nor social policy. We tell you below what we mean by ``economic'' and ``social'' policy.

First, you will read a short section from a party manifesto. For the sentence in the column called ``text'', enter your best judgment about whether it mainly refers to economic policy, to social policy, or to neither. The columns ``pre\_text'' and ``post\_text'' are additional context.

If the sentence refers to economic policy, select ``economic'' in the drop down menu; if it refers to social policy, select ``social''. If the sentence does not refer to either policy area, select ``not economic or social''.

Now we will tell you about what we mean by ``economic'' and ``social'' policy.

\medskip

\textbf{What is ``economic'' policy?}

``Economic'' policies deal with all aspects of the economy, including:

\begin{itemize}[
    leftmargin=1.5em,
    itemsep=2pt,
    topsep=2pt,
    parsep=0pt
]
    \item Taxation and government spending
    \item Services provided by government or other public bodies
    \item Pensions, unemployment and welfare benefits; other state benefits
    \item Property, investment and share ownership, public or private
    \item Interest rates and exchange rates
    \item Regulation of economic activity, public or private
    \item Relations between employers, workers and trade unions
\end{itemize}

\medskip

\textbf{What is ``social'' policy?}

``Social'' policies deal with aspects of social and moral life, relationships between social groups, and matters of national and social identity, including:

\begin{itemize}[
    leftmargin=1.5em,
    itemsep=2pt,
    topsep=2pt,
    parsep=0pt
]
    \item Policing, crime, punishment and rehabilitation of offenders;
    \item Immigration, relations between social groups, discrimination and multiculturalism;
    \item The role of the state in regulating the social and moral behavior of individuals;
    \item Freedom of information.
\end{itemize}

\medskip

\textbf{Examples}

Below we provide two examples of text from the manifestos and instructions on how they should be coded, and why.

\medskip

\textbf{Example 1: economic policy:}

\noindent
\textbf{pre\_text:} With a Conservative Government, all that has been changing.

\noindent
\textbf{text:} We were determined to make share-ownership available to the whole nation.

\noindent
\textbf{post\_text:} Just as with cars, television sets, washing machines and foreign holidays, it would no longer be a privilege of the few

The sentence in the column ``text'' should be coded as ``economic'' policy because it refers to ownership.

\medskip

\textbf{Example 2: social policy:}

\noindent
\textbf{pre\_text:} ALTERNATIVES TO PRISON. Every effort should be made to ensure that fine defaulters, elderly shoplifters and drunks are not sent to prison.

\noindent
\textbf{text:} Police cautions and intermediate treatment should be more widely used.

\noindent
\textbf{post\_text:} Where punishment is appropriate, it should normally be community service rather than prison

The sentence in the column ``text'' should be coded as having to do with ``social'' policy because it refers to policing.

\medskip

\textbf{Policy Area (policy\_area)}

\begin{itemize}[
    leftmargin=1.5em,
    itemsep=2pt,
    topsep=2pt,
    parsep=0pt
]
    \item Not Economic or Social (1)
    \item Economic (2)
    \item Social (3)
\end{itemize}

\medskip

For the target sentence in the column ``text'', respond only with `Economic', `Social', or `Not Economic or Social', according to the codebook. Do not provide any explanation or other output.

\end{tcolorbox}

% =========================================================
% Muller and Proksch (2024)
% =========================================================

\medskip
\noindent\textbf{\large\citealt{muller2024nostalgia}}

\medskip

\begin{tcolorbox}[
    breakable,
    colback=white,
    colframe=black!40,
    boxrule=0.4pt,
    arc=0pt,
    left=6pt,
    right=6pt,
    top=6pt,
    bottom=6pt
]
\small
\raggedright

\textbf{C.1 Instructions for Coding of Nostalgic Rhetoric}

\medskip

In this research project, we aim to measure and explain nostalgia in parties' campaign communication. Lammers and Baldwin (2018: 599) define nostalgia as ``a predominately positive emotion that is associated with recalling memories of important or momentous events, usually experienced with close others.''

\medskip

You will be asked to annotate sentences from party manifestos. The coding proceeds in five steps:

\begin{enumerate}[
    leftmargin=1.8em,
    itemsep=3pt,
    topsep=3pt,
    parsep=0pt
]
    \item You read the `target' sentence (provided in the text field), which should be coded along several dimensions. For a better context, we also provide the sentences before and after the target sentence. Make sure to read the context too.

    [The target sentence is provided as text; the sentence before it is text\_pre and the sentence after it is text\_post.]

    \item You code whether a sentence is nostalgic or not.

    \item For non-English text, code if the translation is inaccurate or very inaccurate. Leave the column blank if the sentence is understandable. Some sentences may be very short or so-called quasi sentences. Only code these sentences as inaccurate translations if they are incomprehensible after reading the context too (text\_pre and text\_post).

    \item If the sentence is not nostalgic, proceed to the next sentence.

    \item If the sentence is coded as nostalgic, please additionally code whether the sentence mentions cultural aspects, economic aspects, both cultural and economic aspects, or neither of these aspects.
\end{enumerate}

Below, we provide examples for each category.

\medskip

\textbf{Nostalgic Sentences}

\medskip

\textbf{Examples}

\begin{itemize}[
    leftmargin=1.5em,
    itemsep=3pt,
    topsep=2pt,
    parsep=0pt
]
    \item ``We need to take pride in our country again and claim back our heritage from the `chattering classes' who have disintegrated our culture, highlighted our failings as a country, rather than celebrating our successes, and tried to make us ashamed to be British.''

    \item ``As a small island nation, we need to constantly strive to build international relationships while preserving and nurturing our vibrant heritage.''

    \item ``Farmers are, and have been, stewards who for thousands of years have protected and preserved the environment, keeping it alive, clean and productive.''
\end{itemize}

\medskip

\textbf{Cultural Nostalgia}

\medskip

Cultural nostalgia mentions aspects such as the cultural plurality or cultural homogeneity within domestic societies. This may include the preservation of the autonomy of religious, and linguistic heritages within the country including special educational provisions.

\medskip

\textbf{Examples}

\begin{itemize}[
    leftmargin=1.5em,
    itemsep=3pt,
    topsep=2pt,
    parsep=0pt
]
    \item ``We recognise Irish as a unique part of our heritage belonging to all and not just the property of one section of the population.''

    \item ``We want a Europe that professes its Christian-Western roots and the ideas of the Enlightenment and lives from them.''
\end{itemize}

\medskip

\textbf{Economic Nostalgia}

\medskip

Economic nostalgia mentions aspects such as past economic growth, government spending, public services, the welfare state or state benefits, unemployment, investment, pensions, or relations between employers and trade unions.

\medskip

\textbf{Example}

\begin{itemize}[
    leftmargin=1.5em,
    itemsep=3pt,
    topsep=2pt,
    parsep=0pt
]
    \item ``History has shown that the social market economy can best create lasting prosperity.''
\end{itemize}

\medskip

\textbf{Economic and Cultural Nostalgia}

\medskip

These sentences include references to both cultural and economic aspects.

\begin{itemize}[
    leftmargin=1.5em,
    itemsep=3pt,
    topsep=2pt,
    parsep=0pt
]
    \item ``For hundreds of years, the United Kingdom has determined the rules and formed the environment where new ideas and new technologies prosper -- from financial markets to the steam train to human embryology and the code of life itself.''
\end{itemize}

\medskip

\textbf{Neither Economic nor Cultural}

\medskip

\textbf{Example}

\begin{itemize}[
    leftmargin=1.5em,
    itemsep=3pt,
    topsep=2pt,
    parsep=0pt
]
    \item ``A Conservative Government will be optimistic about Britain's future because we are comfortable with Britain's past.''
\end{itemize}

\medskip

For this classification run, return only `Yes' if the target sentence is nostalgic or `No' if it is not, according to the codebook. Do not output a translation flag, a nostalgia subtype, an explanation, or any other text.

\end{tcolorbox}

% =========================================================
% BFRS Dataset on Political Violence in Pakistan
% =========================================================

\medskip
\noindent\textbf{\large BFRS Dataset on Political Violence in Pakistan}

\medskip

\noindent For this paper, we used the prompt created by \cite{Halterman2026CodebookLLMs}.

\medskip

\begin{tcolorbox}[
    breakable,
    colback=white,
    colframe=black!40,
    boxrule=0.4pt,
    arc=0pt,
    left=6pt,
    right=6pt,
    top=6pt,
    bottom=6pt
]
\small
\raggedright

\textbf{Instructions}

\medskip

\textbf{Instruction:} You're an expert political scientist categorizing news stories from Pakistan into categories. Carefully read the definitions below, read the story, and write the Label that best matches the story. Use only the provided labels.

\medskip

\textbf{Task:} single label multiclass

\medskip

\textbf{Output Reminder:} Write the name of the Label that fits best, with no other text. For example, ``Label: ASSASSINATION'', ``Label: VIOLENT\_POLITICAL\_DEMONSTRATION'', etc.

\medskip

\textbf{Categories}

% ---------------------------------------------------------
% 1. Assassination
% ---------------------------------------------------------

\medskip

\textbf{assassination}

\smallskip
\noindent\rule{\linewidth}{0.3pt}
\smallskip

\textbf{Label:} \texttt{ASSASSINATION}

\textbf{Definition:} An attempt (successful or failed) by a non-state entity aimed to kill a specific individual.

\textbf{Clarification:} Targets may include military, political, civil society or civilian state or federal leadership. In some case, there may be a suspected government agency behind the assassination. This Label applies when a body is found with bullet wounds or other marks of violence and/or torture.

\textbf{Negative Clarification:} Does not include assassination using drone attacks (coded separately) or deaths from attacks on military, paramilitary, police or intelligence targets, or state violence against a group of civilians.

\textbf{Positive Example:} ``An Imam of a mosque was brutally murdered by unknown assailants in Jehanian on Friday.'' This is an ASSASSINATION because the imam appears to have been deliberately targeted.

\textbf{Negative Example:} ``A woman and her son were injured by unidentified assailants in Mominabad on Sunday.'' This is an example of TERRORISM, not ASSASSINATION, because the attack does not appear to be specifically targeted at the victims.

% ---------------------------------------------------------
% 2. Drone Assassination
% ---------------------------------------------------------

\medskip

\textbf{assassination (drone attack)}

\smallskip
\noindent\rule{\linewidth}{0.3pt}
\smallskip

\textbf{Label:} \texttt{DRONE\_ASSASSINATION}

\textbf{Definition:} An assassination (failed or successful attempt) carried out using an unmanned aerial vehicle (drone strike).

\textbf{Negative Clarification:} Non-drone assassination attempts are coded under the `Assassination' category, not under this category.

\textbf{Positive Example:} ``Drone fired missiles at a heavily fortified mud-house in Asoray Village in the Mirali sub-division kiliing 5 people and injuring 1.;a senior Al Qaeda commander, Hamza Rabia, was killed in the attack''

\textbf{Negative Example:} ``Two soldiers were injured in a bomb blast near an army outpost in the Hamrang area.'' No mention of a drone strike.

% ---------------------------------------------------------
% 3. Attack on State
% ---------------------------------------------------------

\medskip

\textbf{attack on state}

\smallskip
\noindent\rule{\linewidth}{0.3pt}
\smallskip

\textbf{Label:} \texttt{ATTACK\_ON\_STATE}

\textbf{Definition:} An attack on Pakistani territory targeted at the state of Pakistan or its representatives that was conducted by the armed forces of another state. This includes shelling across the Line of Control, attacks on Pakistani forces mistaken for militants near the Afghan border.

\textbf{Clarification:} All these incidents were attributed to the government of India or United States and NATO forces.

\textbf{Positive Example:} ``Indian forces resorted to machine gun firing across the LoC in what the Pakistani officials termed a clear violation of the ceasefire.'' This is an example of another state's military conducting an attack on Pakistan.

\textbf{Negative Example:} ``Five rockets were fired at a Frontier Corps checkpost in Sangsilla Chashma area of Pir Koh.'' Not attributable to the armed forces of another state.

% ---------------------------------------------------------
% 4. Conventional Attack
% ---------------------------------------------------------

\medskip

\textbf{conventional attack on military/paramilitary/police/intelligence}

\smallskip
\noindent\rule{\linewidth}{0.3pt}
\smallskip

\textbf{Label:} \texttt{CONVENTIONAL\_ATTACK\_ON\_GOV\_FORCES}

\textbf{Definition:} Incidents that involve a direct conventional attack on a military, police, paramilitary, or intelligence target, such that violence has the potential to be exchanged between the attackers.

\textbf{Clarification:} Conventional tactics are those that would be employed by state forces. Conventional attacks include ambushes, direct fire, artillery, pitched battle, and troop captures. This includes but is not limited to: attacks against military checkposts, police headquarters, the Federal Investigative Agency (FIA), the ISI.

\textbf{Negative Clarification:} ``Guerilla'' tactics are described in a separate category. Attacks by state actors (e.g.\ India) are coded as ``attack on state''.

\textbf{Positive Example:} ``At least two security personnel were injured in skirmishes between security forces and armed tribesmen. The attackers escaped after security forces retaliated using gunfire. No arrest was reported.'' Skirmishes involving gunfire is a conventional attack.

\textbf{Negative Example:} ``A vehicle carrying security personnel hit a landmine in Loti gas field area resulting in injuries to three soldiers.'' This is a guerilla attack because it involves a road-side bomb.

% ---------------------------------------------------------
% 5. Guerilla Attack
% ---------------------------------------------------------

\medskip

\textbf{guerilla attack on military/paramilitary/police/intelligence}

\smallskip
\noindent\rule{\linewidth}{0.3pt}
\smallskip

\textbf{Label:} \texttt{GUERILLA\_ATTACK\_ON\_GOV\_FORCES}

\textbf{Definition:} Incidents that involve a guerilla-like attack (e.g., a car or suicide-bombing, a kidnapping) on a military, police, paramilitary, or intelligence target.

\textbf{Clarification:} Guerilla attacks employ tactics that conventional forces do not. For example, guerilla attacks will employ road-side bombs, improvised explosive devices (IEDs), suicide attacks (e.g.\ on a checkpoint), car bombs and the like.

\textbf{Negative Clarification:} Gun battles between soldiers and militants are coded as conventional attacks.

\textbf{Positive Example:} ``A suicide bomber blew himself up in an army training area in Guliana near Kharian cantonment, where the soldiers were busy training.'' Suicide attacks are examples of guerilla attacks.

\textbf{Negative Example:} ``Mewind Rifles received information about the presence of militants in the area. When security forces reached there, militants started firing.'' This describes a conventional battle between militants and state forces and is thus not a guerilla attack.

% ---------------------------------------------------------
% 6. Government Attack on Non-State Combatants
% ---------------------------------------------------------

\medskip

\textbf{military/paramilitary/police attack on non-state combatants}

\smallskip
\noindent\rule{\linewidth}{0.3pt}
\smallskip

\textbf{Label:} \texttt{GOV\_ATTACK\_ON\_NONSTATE\_COMBATANTS}

\textbf{Definition:} Violence initiated by state (federal or provincial) combatants against combatant non-state, sub-national groups, or clandestine agents.

\textbf{Clarification:} Includes attacks by the US and NATO on militants in Pakistan that involve manned aircraft or artillery systems. For example, these may be brief episodes such as the Pakistan Army and Police assault on the Lal Masjid OR they may be sustained as the army offensive in South Waziristan in 2011 or the Frontier Corps in Aurakzai.

\textbf{Negative Clarification:} drone attacks have their own category

\textbf{Positive Example:} ``In a continuation of the crackdown on militants by the army, four more militants were killed.'' State forces attacked non-state armed actors.

\textbf{Negative Example:} ``A personnel of the Balochistan Reserve Police was injured when some armed men opened indiscriminate fire on them which they later returned.'' This is a non-state attack on state forces, so this label does not apply.

% ---------------------------------------------------------
% 7. Government Attack on Civilians
% ---------------------------------------------------------

\medskip

\textbf{military/paramilitary/police selective violence}

\smallskip
\noindent\rule{\linewidth}{0.3pt}
\smallskip

\textbf{Label:} \texttt{GOV\_ATTACK\_ON\_CIVILIANS}

\textbf{Definition:} Violence initiated by state (federal or provincial) combatants against civilians.

\textbf{Clarification:} These may include incidents where a newspaper reports that a political activist is picked up by the intelligence services and detained and beaten overnight.

\textbf{Negative Clarification:} Does not include attacks on combatants or violence in the context of a protest.

\textbf{Positive Example:} ``Advocate Ghulam Nabi of the Peshawer High Court was allegedly picked up by personnel of an intelligence agency in three vehicles near the NWFP assembly, and released after a severe over night thrashing. He is associated with Jamaat-i-Islami and a critic of the army and intelligence agencies.''

\textbf{Negative Example:} ``The police baton charged lawyers and other political activists as they tried to enter the city in order to join the throng of people outside the Supreme Court who were there supporting the sacked Chief Justice. Hundereds of people tried entering Islamabad for the purpose, which the police tried to stop by baton-charging the crowd.'' This violence takes place in the context of a protest, and thus should get the label ``violent political demonstration''.

% ---------------------------------------------------------
% 8. Riot
% ---------------------------------------------------------

\medskip

\textbf{riot}

\smallskip
\noindent\rule{\linewidth}{0.3pt}
\smallskip

\textbf{Label:} \texttt{RIOT}

\textbf{Definition:} A riot is a violent clash between two or more sizeable groups or when a single informally assembled crowd becomes violent.

\textbf{Clarification:} Neither group can be a state force. However, if there are two or more clashes around a single event in which the police are involved, then it would be a classified as a riot.

\textbf{Negative Clarification:} For example, the students of the Lal Masjid versus the police would not be considered a riot, it would be a Violent Political Demonstration.

\textbf{Positive Example:} ``Five workers of the Muslim Students Federation, the student wing of the ruling Muslim conference, sustained bullet wounds in a clash with the activists of pro-independence National Students Federation.''

\textbf{Negative Example:} ``An angry mob thrashed police personnel, accusing them of manhandling people and using obscene language when they (the police) raided a gambling den.'' This incident involves police, so it is coded as a ``violent political demonstration.''

% ---------------------------------------------------------
% 9. Terrorism
% ---------------------------------------------------------

\medskip

\textbf{terrorism}

\smallskip
\noindent\rule{\linewidth}{0.3pt}
\smallskip

\textbf{Label:} \texttt{TERRORISM}

\textbf{Definition:} Premeditated, politically motivated violence perpetrated against noncombatant targets by subnational groups or clandestine agents (22 U.S.C.\ \S~2656f(d)(2)).

\textbf{Positive Example:} For example, the attack on the Data Ganj Baksh shrine is such an example as the victims were primarily intended to be non-combatants.

\textbf{Negative Clarification:} In contrast, an attack on a police check point is not considered to be terrorism even if civilians are harmed in the conduct of the attack provided that the primary target was the check point. This is coded as an attack on police, paramilitary or military target. The intended targets of the attacks is key. A targeted attack on a specific figure is an instance of assassination, not terrorism.

\textbf{Positive Example:} ``Security forces foiled an attempt to blow up a 12-inch diameter gas pipeline in Sui tehsil.'' Even though this attack was unsuccessful, it still involved an attack by non-state groups on non-combatant targets.

\textbf{Negative Example:} ``The grandson of Mohmand Agency MNA Maulana Ghulam Mohammad Sadiq was killed when gunmen entered his house and opened fire.'' The attack seems directed at a specific prominent person, so even though it's carried out by non-state actors, it should be coded as an assassination, not terrorism.

% ---------------------------------------------------------
% 10. Threat of Violence
% ---------------------------------------------------------

\medskip

\textbf{threat of violence}

\smallskip
\noindent\rule{\linewidth}{0.3pt}
\smallskip

\textbf{Label:} \texttt{THREAT\_OF\_VIOLENCE}

\textbf{Definition:} Incidents in which the threat of violence is reported as being used for political purposes.

\textbf{Clarification:} A typical example is when a spokesman for a banned militant group threatens attacks if certain policy changes were not made, or when a party spokesman threatens violent protest if certain changes are not made. This also includes incidents in which a particular individual is threatened.

\textbf{Positive Example:} ``The President of the All Pakistan Clerks Association...alleged that he had received a phone call warning him of an explosion at his house.''

\textbf{Negative Example:} ``A man killed his cousin, accusing her of having illicit relations with a youth of the same village.'' This is an instance of violence actually being carried out, not a threat of violence.

% ---------------------------------------------------------
% 11. Violent Political Demonstration
% ---------------------------------------------------------

\medskip

\textbf{violent political demonstration}

\smallskip
\noindent\rule{\linewidth}{0.3pt}
\smallskip

\textbf{Label:} \texttt{VIOLENT\_POLITICAL\_DEMONSTRATION}

\textbf{Definition:} A violent political demonstration is characterized as mobilization of crowds in response to a political event that at some point becomes violent.

\textbf{Clarification:} Violence may be initiated by the crowd or by the police upon the crowd.

\textbf{Negative Clarification:} A violent clash between two non-state groups is a riot, not a violent political demonstration.

\textbf{Positive Example:} ``Hundreds of school children and armed supporters of Lashkar-i-Islam stormed the Bara bazaar and set the main tehsil, Qambaraband, Al Haaj Market, Sheikhan and Qasaban checkposts of the Khasadar force on fire. The Lashkar-i-Islam had warned of protests if their demands were not met.''

\textbf{Negative Example:} ``A man killed and his brother injured when they were attacked by their rivals over a land dispute.'' The motivation here is not a political event, thus this event is a riot, not a violent political demonstration.

% ---------------------------------------------------------
% 12. Other
% ---------------------------------------------------------

\medskip

\textbf{other}

\smallskip
\noindent\rule{\linewidth}{0.3pt}
\smallskip

\textbf{Label:} \texttt{OTHER}

\textbf{Definition:} Incidents that are not easily categorized such as people being hurt by unexploded ordnance or when the source of the violence is unclear.

\textbf{Clarification:} An event that does not fit into the other categories (assassination, attack on state, conventional attack on military/paramilitary/police/intelligence, guerilla attack on military/paramilitary/police/intelligence, military/paramilitary/police selective violence) should be coded as OTHER.

\textbf{Positive Example:} ``A man allegedly opened fire on his wife while she was offering prayers, and later proceeded to kill his brother on the pretext that his wife and his brother had illicit relations. He then escaped.''

\textbf{Negative Example:} ``A suicide bomber blew himself up in an army training area in Guliana near Kharian cantonment, where the soldiers were busy training.''

\end{tcolorbox}

\normalsize

\subsection{Details of Crowd Sourced Annotation}\label{app:crowd_procedure}

\subsubsection{Procedure of Data Collection}

We fielded the crowd-sourced annotation experiment on CloudResearch Connect in June 2026. Prior to fielding, we preregistered the design and analysis on the Open Science Framework (OSF), obtained Institutional Review Board approval, and fielded a small pilot survey to test our instrument.
%\footnote{OSF -- [REDACTED]
%[REDACTED]; 
%Harvard University IRB [REDACTED].
%IRB Number -- [REDACTED]
%} 
Our sample of $N=165$ is a U.S. census-matched demographic, selected from Connect’s demographic targeting template. Each respondent is tasked with coding 20 texts as belonging to a topic or not, based on a standardized codebook. Given the difficulty of the task and to ensure response quality, we further implemented the following inclusion criteria, which we preregistered: respondents must be native English speakers, pass all attention and bot checks, not exceed Connect's default maximum time spent of 48 minutes, and have a platform approval rate of at least 95\%, following prior conventions \citep{PeerVosgerauAcquisti2014, kennedy_shape_2020}.

Our survey starts with a short set of demographic and baseline questions to gauge respondents' political knowledge, which we used as moderators in exploratory analyses. Respondents are then randomly assigned to one of 15 survey tasks: one for each of the 14 replicated papers and a second Schub (2022) task with an improved codebook. The respondent reads each task's codebook (the same codebook used by LLMs and humans for each of the 14 papers) and then codes that task's 20 texts as Yes/No. Figure~\ref{fig:survey_eg} illustrates an example while Table~\ref{tab:questionnaire} details our survey questions.

\begin{figure}[H]
    \centering
    \includegraphics[width=0.78\linewidth]{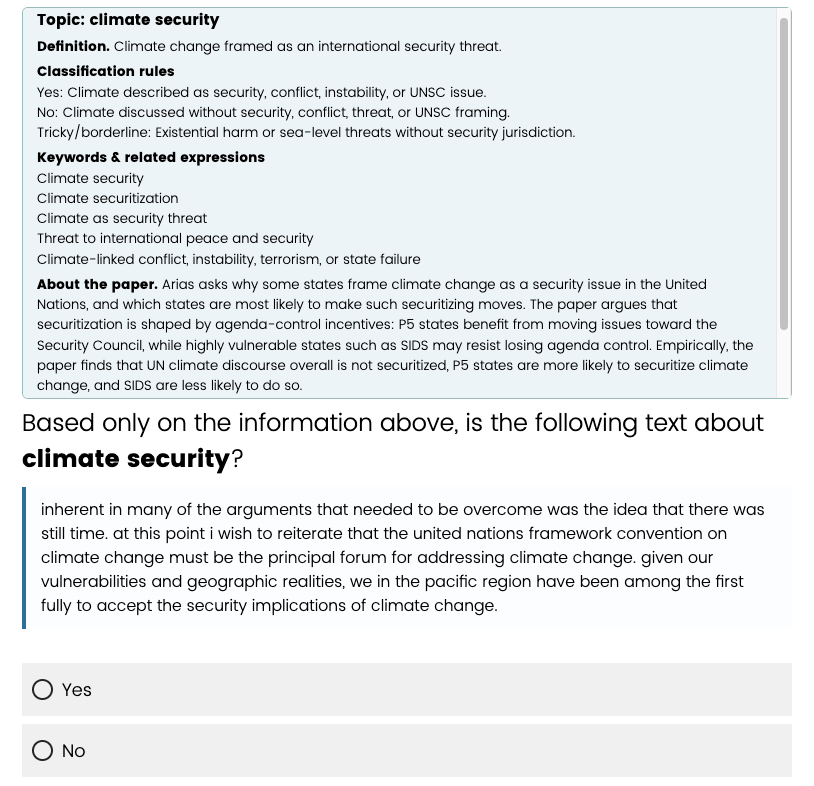}
    \caption{Example of coding task for the \cite{arias2022securitizes} paper.}
    \label{fig:survey_eg}
\end{figure}

\begin{small}
\begin{longtable}{@{}p{3.4cm}p{11.6cm}@{}}
\caption{Survey Questionnaire}\label{tab:questionnaire}\\
\toprule
Item & Question \\
\midrule
\endfirsthead
\toprule Item & Question \\ \midrule \endhead
\bottomrule \endfoot
\addlinespace[8pt]
\multicolumn{2}{@{}l}{\emph{Baseline and moderator items}} \\[8pt]
Political knowledge & Which of the following best describes a political party manifesto or platform?\newline \textit{Options:} A document describing a party's policy positions (correct); A court ruling; A private tax document; A voter registration form; Not sure \\[4pt]
News & How often do you follow news about politics or public policy?\newline \textit{Options:} Never; Once or a few times a month; A few times a week; Daily \\[4pt]
Coursework & Have you ever taken a political science, government, public policy, or international relations course?\newline \textit{Options:} No; Yes, one; Yes, more than one; Not sure \\[4pt]
\addlinespace[8pt]
\multicolumn{2}{@{}l}{\emph{Demographic items}} \\[8pt]
Age & What is your age? \\[4pt]
Gender & What is your gender? \\[4pt]
Education & What is the highest level of education you have completed? \\[4pt]
Household income & What was your total household income last year (before taxes)? \\[4pt]
\addlinespace[8pt]
\multicolumn{2}{@{}l}{\emph{Coding task}} \\[8pt]
Codebook & [Follows the same codebook as our human and LLM annotation, comprising: topic, definition, classification rules, label wording, summary of paper, summary of data] \\[4pt]
Classification ($\times$20) & Based only on the information above, is the following text about [the task's topic]? (Yes / No). \\[4pt]
\addlinespace[8pt]
\multicolumn{2}{@{}l}{\emph{Post-task feedback}} \\[8pt]
Difficulty & How difficult was the classification task? \\[4pt]
Confidence & How confident are you in how you classified the texts for the task? \\[4pt]
Basis & Which best describes how you made your classifications? Select all that apply.\newline \textit{Options:} Used the definition; Used the rules; Used keywords / label wording; Used the summary of the paper and data; Used my own intuition; Used my background knowledge about the topic; Consulted other sources; Other \\[4pt]
\end{longtable}
\end{small}

Our preregistered sample target is approximately 150 respondents, or 10 coders per task. Compared to having three human coders, having 10 coders per paper reduces the likelihood of chance agreement across all coders by 128 times. Since we randomized the task assignment, we noticed a chance imbalance of less than 10 coders for some tasks and more than 10 for some others. To ensure we have at least 10 coders for each paper, we paused the survey after the initial 150 responses, excluded papers which already hit our target from our survey block randomizer, and continued the survey until we had 10 coders or more for the remaining papers. This gives us our final sample of 165 complete responses, with 10 to 13 coders per task. We report the sample demographics in Table~\ref{tab:sample_composition} below.

\begin{table}[H]
\centering
\caption{Demographic characteristics of survey respondents.}
\label{tab:sample_composition}
\small
\sbox0{\begin{tabular}{@{}lcc@{}}
\toprule
 & $n$ & \% \\
\midrule
\multicolumn{3}{@{}l}{\emph{Age}} \\
\quad 18--24 & 12 & 7\% \\
\quad 25--34 & 32 & 19\% \\
\quad 35--44 & 37 & 22\% \\
\quad 45--54 & 31 & 19\% \\
\quad 55--64 & 27 & 16\% \\
\quad 65 or older & 25 & 15\% \\
\quad Prefer not to say & 1 & 1\% \\
\addlinespace[3pt]
\multicolumn{3}{@{}l}{\emph{Gender}} \\
\quad Woman & 88 & 53\% \\
\quad Man & 74 & 45\% \\
\quad Non-binary or another gender & 1 & 1\% \\
\quad Prefer not to say & 2 & 1\% \\
\addlinespace[3pt]
\multicolumn{3}{@{}l}{\emph{Education}} \\
\quad High school diploma or GED & 12 & 7\% \\
\quad Some college, no degree & 28 & 17\% \\
\quad Associate degree & 15 & 9\% \\
\quad Bachelor's degree & 72 & 44\% \\
\quad Graduate or professional degree & 37 & 22\% \\
\quad Prefer not to say & 1 & 1\% \\
\addlinespace[3pt]
\multicolumn{3}{@{}l}{\emph{Household income}} \\
\quad Under \$25,000 & 22 & 13\% \\
\quad \$25,000--\$49,999 & 28 & 17\% \\
\quad \$50,000--\$74,999 & 30 & 18\% \\
\quad \$75,000--\$99,999 & 34 & 21\% \\
\quad \$100,000--\$149,999 & 26 & 16\% \\
\quad \$150,000 or more & 20 & 12\% \\
\quad Prefer not to say & 5 & 3\% \\
\addlinespace[3pt]
\multicolumn{3}{@{}l}{\emph{News consumption}} \\
\quad Daily & 68 & 41\% \\
\quad A few times a week & 64 & 39\% \\
\quad Once or a few times a month & 28 & 17\% \\
\quad Never & 5 & 3\% \\
\addlinespace[3pt]
\multicolumn{3}{@{}l}{\emph{Politics/Policy coursework}} \\
\quad Yes, more than one & 46 & 28\% \\
\quad Yes, one & 53 & 32\% \\
\quad No & 57 & 35\% \\
\quad Not sure & 9 & 5\% \\
\addlinespace[3pt]
\multicolumn{3}{@{}l}{\emph{Factual knowledge item}} \\
\quad Correct & 149 & 90\% \\
\quad Incorrect & 16 & 10\% \\
\bottomrule
\end{tabular}}
\usebox0
\end{table}

\subsubsection{Text selection and Data Quality}

For each coding task we selected 20 texts from the replicated papers, taken verbatim.\footnote{Some texts are abruptly truncated at parts or contain mojibake. We chose to preserve all original texts without any modification to keep our replication exercise consistent. To reduce confusion, we informed respondents that ``some texts may appear vague or incomplete --- we ask that you answer to the best of your understanding.''} Out of the original 100 texts per paper used in our main analyses, we used stratified sampling to obtain the 10 ``most agreed'' and 10 ``most disagreed'' texts based on the LLM annotation. In other words, for each paper, we took the 10 texts with highest LLM disagreement (ranked by across-model Gini impurity) and 10 texts with lowest disagreement, calculated using the original 10 LLMs (excluding the three Claude models, which were added to our analysis after the crowd-sourced experiment). This oversamples ambiguous texts by design, so that every task spans both straightforward and difficult cases and lets us compare coders across that range. For all analyses involving crowd annotations, we only use the annotations from the subsample of 20 texts per task, from crowd workers, humans, and LLMs.

Of note, Schub (2022) contributes one set of 20 texts, coded under both its original and its improved codebook. Holding the same set of 20 texts constant allows us to compare the effects of the codebook change since assignment to each task is randomized. Across the 14 papers, our text selection procedure yields 280 unique texts. 

We capped each task at 20 texts to protect response quality and potential declines in data quality if the task was too long. We estimated the number of texts to assign for crowd coding based on the expected duration of the task, informed by our own human coding and a pilot test. In our pilot test of $N = 42$ fielded shortly before the main study (also in June 2026 and on Connect with the same demographic selection and tasks), coding 20 texts took a median of 10.7 minutes and mean of 11.3. We found this to be short enough to hold a coder's attention while still giving enough texts to compare coders within a task. Our main survey took a median of 12.8 minutes and mean of 14.8. For all tasks, text order is randomized to reduce fatigue and learning effects.

Given that our analysis compares human crowd workers with LLMs, we took extensive care to prevent AI-assisted responses \citep{Westwood2025}. First, we programmed our survey on Qualtrics to prevent any copying and pasting of text content, raising the barrier to routing a text through an LLM. Second, we added a honeypot text field that is visible to bots but not to humans. Third, we enabled Qualtrics' bot-detection features, a CAPTCHA question, and duplicate-submission prevention.\footnote{Accessed 23 July, 2026. \url{https://www.qualtrics.com/support/survey-platform/survey-module/survey-checker/fraud-detection/}} Fourth, our survey provider, CloudResearch Connect, applies its own extensive anti-bot screening.\footnote{Accessed 23 July 2026. \url{https://www.cloudresearch.com/resources/blog/ai-bot-threat-to-survey-research/}}

We used our pilot to test the survey's length and data quality checks before fielding the main study. We generally found high data quality and good engagement across both the pilot and full survey: crowd agreement with the humans is fairly high, completion times are close to what we expected, and straight-lining is rare. The final sample of 165 respondents includes only those that pass every preregistered exclusion criteria. Numerous respondents failed our attention checks and are excluded while one respondent took more than eight hours and is excluded for exceeding Connect's maximum survey time. Respondents are pre-screened on Connect for native English language proficiency and a platform approval rate of at least $95\%$, so we do not collect any data from those that do not meet these criteria.

Additionally, among our final sample of 165, three respondents straight-lined all twenty answers. Given that we did not preregister straight-lining as an exclusion criterion, we retain these three responses in the main specification and drop them only as a robustness check. Appendix~\ref{app:crowd_robustness} reports that check and the other robustness analyses.

\newpage

\section{Examples of unambiguous and ambiguous texts}
\label{app:example_ambiguous}

\scriptsize

% =========================================================
% Unambiguous examples
% =========================================================

\begin{longtable}{p{1.5cm} p{1.5cm} p{12cm}}
\caption{Examples of unambiguous texts}
\label{tab:examples_unambiguous}\\

\toprule
\textbf{Paper} & \textbf{Topic} & \textbf{Text} \\
\midrule
\endfirsthead

\multicolumn{3}{l}{\small\itshape Table \thetable\ continued from previous page}\\
\toprule
\textbf{Paper} & \textbf{Topic} & \textbf{Text} \\
\midrule
\endhead

\midrule
\multicolumn{3}{r}{\small\itshape Continued on next page}\\
\endfoot

\bottomrule
\endlastfoot

\cite{arias2022securitizes}
&
Climate security
&
Climate change is rightfully described as the most urgent threat facing mankind, and, at least for the next several months, it will remain at the top of the global diplomatic and negotiating agenda. But what is the challenge of climate change if not a risk to development, security and peace? What is the threat of climate change, if not a threat to the very notion of human survival and ecological balance? \\
\addlinespace

\cite{BlumenauLauderdale2018}
&
Finance
&
PURPOSE: to establish a new instrument for pre-accession, IPA II, in the framework of the reform of the EU external action financial instruments and following on from the unified Instrument for Pre-Accession Assistance, IPA, for the potential candidate countries to accession. PHILOSOPHY AND ACTION PLAN FOR EXTERNAL AID: what happens outside the borders of the EU can and does directly affect the prosperity and security of EU citizens. It is therefore in the interest of the EU to be actively engaged in influencing the world around us, including through the use of financial instruments. \\
\addlinespace

\cite{bush_facing_2023}
&
Taxes for climate policy
&
It would harm me through the increases in costs of energy. \\
\addlinespace

\cite{ClarkDolan2021}
&
Fiscal policy
&
The Recipient has demonstrated enhanced predictability of budget execution as reflected by the fact that, during Fiscal Year 2009, at least 85\% of Budget Heads expended between 95\% and 105\% of their total budgetary allocated funding for said year, as maintained to the date of this Agreement. \\
\addlinespace

\cite{feltovich_campaign_2024}
&
Negative campaigning
&
The other challenger either failed in maths or lied to you. \\
\addlinespace

\cite{jung_mobilizing_2020}
&
Moral rhetoric
&
Changes have been made to the Sex Discrimination Act: no one can be discriminated against based on family responsibilities, while breastfeeding is now its own separate grounds of discrimination. \\
\addlinespace

\cite{mattingly_chinese_2025}
&
Western dysfunction
&
The Wall Street showdown over GameStop share prices: is it a storm in a teacup or the tip of the iceberg? How troublesome is it for the U.S. bourse? A vaccine wall of defense against COVID slowly built. \\
\addlinespace

\cite{parthasarathy_deliberative_2019}
&
Fund allocation
&
Due to absence of ration shop facility due to rain water goes in. The food items are all spoiled. Please get it corrected. \\
\addlinespace

\cite{thrall_informational_2025}
&
Commercial issues
&
But of even greater popular and political concern was the growing trade imbalance between Japan and the United States. I think the trade deficit with Japan was somewhere in the neighborhood of \$3,000,000,000 annually. There was little point in making the argument that a negative bilateral trade imbalance had little or no economic significance. Despite all the efforts of the leading world economists beginning with Adam Smith, mercantilist theory still dominated popular thinking and the almost unanimous view was that a negative trade imbalance was a sign of economic weakness and that drastic action was necessary in order to see it eradicated or at least sharply reduced. Resentment on both sides grew as our manufacturers increasingly complained about the difficulties they were facing in exporting to Japan, which they blamed, primarily if not exclusively, on Japanese policies and practices designed to thwart imports. \\

\end{longtable}

% =========================================================
% Ambiguous examples
% =========================================================

\newpage 

\begin{longtable}{p{1.5cm} p{1.5cm} p{12cm}}
\caption{Examples of ambiguous texts}
\label{tab:examples_ambiguous}\\

\toprule
\textbf{Paper} & \textbf{Topic} & \textbf{Text} \\
\midrule
\endfirsthead

\multicolumn{3}{l}{\small\itshape Table \thetable\ continued from previous page}\\
\toprule
\textbf{Paper} & \textbf{Topic} & \textbf{Text} \\
\midrule
\endhead

\midrule
\multicolumn{3}{r}{\small\itshape Continued on next page}\\
\endfoot

\bottomrule
\endlastfoot

\cite{arias2022securitizes}
&
Climate security
&
Climate change must be further mainstreamed into the work of the whole United Nations system, with a view to supporting efforts to help the transition to low-carbon economies consistent with sustainable development, to strengthen countries' adaptation and resilience in the face of climate change, and to minimize the possible security implications. In the light of diminishing natural resources, environmental degradation, extreme poverty, hunger and diseases, and social unrest, we agree with others that sustainable development has become the defining issue of our time. Our highly globalized and interdependent world means that we share not only the same challenges but a common fate. \\
\addlinespace

\cite{arias2022securitizes}
&
Climate security
&
08-53141 40 being especially vulnerable. the recent hurricanes that left such a trail of destruction across the caribbean once again brought into sharp relief the acute vulnerabilities of small island states, such as the maldives, to global warming and climate change. for the maldives, climate change is not a distant possibility. it is happening now and is a reality that we are experiencing on a daily basis. the continuing degradation of the global environment is not only undermining our development process, but also seriously threatening the very survival of our people and the existence of our tiny country. \\
\addlinespace

\cite{blaydes_mirrors_2018}
&
Art of rulership
&
the Church of Christ. [5] We must make the same judgment about the end of the whole multitude as we do of one person.170 If, therefore, the end of a person were some good existing in that one alone, and the ultimate end of the multitude to be governed were similar in that the multitude should acquire such a good and preserve it, and if indeed such an ultimate end were a corporal one, either of one person or of the multitude, then the life and health of that body would be the duty of a physician. If the ultimate end were affluence and riches, a steward should be king of the multitude. If the good of knowing the truth were of the kind which the multitude could attain, the king would have the same duty as a professor. [ 6] It seems that the end of a multitude gathered together is to live according.\\
\addlinespace

\cite{blaydes_mirrors_2018}
&
Art of rulership
&
Any other person, have revealed to us the true sense of the oracle. Why do we then delay to offer him the crown, whom the Fates have ordained to be our king? The old men immediately quitted the sacred grove, and the chief of them, taking me by the hand, acquainted the people, who waited with impatience for their decision, that I had gained the prize. Scarcely had he done speaking, when a confused noise ran through the whole assembly. Everyone shouted for joy. The whole coast, and neighboring mountains, echoed with these words: ``May the son of Ulysses, who resembles Minos, reign over the Cretans.'' After waiting a while, I made a sign with my hand, to ask to be heard. In the meantime, Mentor whispered thus in my ear: ``Are you going to renounce your country? Will the ambition of reigning make you forget Penelope, who longs for you as her only \ldots'' \\
\addlinespace

\cite{ClarkDolan2021}
&
Fiscal policy
&
The Recipient has reduced administrative costs arising from (A) the creation of a business and (B) the transfer of urban commercial real property (mutation d'immeubles urbains b\^atis et non-b\^atis) by amending the Recipient's Registration Code (Code de l'enregistrement) so as to reduce the registration fees in connection with: (i) the creation of a business; and (2) the transfer of urban commercial real property (mutation d'immeubles urbains b\^atis et non-b\^atis), each by at least fifty percent (50\%).\\
\addlinespace

\cite{feltovich_campaign_2024}
&
Negative campaigning
&
this is the best I can do. Q7. If anyone gives you more, they will have to sacrifice themselves, which they won't. \\
\addlinespace

\cite{lacombe2019political}
&
Gun control
&
The pioneer forefathers of the present generation of Californians would find it hard to believe their eyes and their ears if they were abroad in the Golden State today. Under the ostensible leadership of a member of the California Prison Board, a state-wide campaign has been launched to secure the necessary names on a petition to force a state-wide total-disarmament bill on the election ballot this fall. The bill on which the voters will be asked to register their vote would completely prohibit the retail or wholesale distribution of pistols or revolvers or any other weapon which may be concealed upon the person, and would also completely prohibit the possession of such weapons by any except the military and police. The proponents of the measure frankly admit that criminals will continue to obtain concealable weapons, and that as a matter of fact the principal source of supply for criminals at this time is the pistol bootlegger and not the legitimate dealer. The naive claim is made, however, that this continued bootlegging of pistols will not be as serious as the present legitimate sale and possession of such arms under California's permit system, because the police, when they find a man with a gun on him, will have all the evidence they need to send him to jail for a year. At the present time the police have the same opportunity under the existing law, but the crime isn't a felony. The Sullivan Law has failed to stop crime in New York State, according to the California reformers, because it does not go far enough. \\
\addlinespace

\cite{gilardi_policy_2021}
&
Smoking-ban enforcement
&
Snellville's ordinance would not be as stringent. It would allow smoking in private offices, for example, and would not require businesses to post no-smoking signs. Council Member Bruce Garraway, an outspoken tobacco opponent, urged council members to consider adopting a tougher ban, one that prohibits smoking in private offices. The city also should require outdoor smokers to stand away from doorways so smoke will not drift inside restaurants or offices, he said. ``We'll have to look at that,'' said Oberholtzer. ``I think we'll probably have some amendments brought up.'' The ban would authorize Snellville's code officers and police to fine anyone who violated the ordinance. The fines range from \$50 for a first-time offender to \$250 for people who consistently fire up where they should not. Snellville's proposals indicate that smoking is under fire as never before. The city would not be the first Georgia municipality to clamp down on indoor smoking; Grayson, Loganville and Bainbridge have smoking bans in effect, according to Georgians Against Smoking Pollution. \\
\addlinespace

\cite{jung_mobilizing_2020}
&
Moral rhetoric
&
But you don't need a safety net unless you've turned the health system into a highwire act and families are in danger of falling off.\\
\addlinespace

\cite{mattingly_chinese_2025}
&
Western dysfunction
&
united nations human rights chief michelle bachelet urged the global community to fight discrimination against the people of asian origin if the world is to effectively combat the covid outbreak she made the remarks during a session of the un. \\
\addlinespace

\cite{osnabrugge_playing_2021}
&
Emotive rhetoric
&
May I thank the hon. Lady who has run such an effective campaign on this and the colleagues across the House who have written about this matter to my right hon. Friend the Home Secretary. As she knows, the previous Home Secretary, in her capacity as both Home Secretary and Minister for Women and Equalities, took this subject extremely seriously, as does the new Home Secretary. We are drawing together the evidence and looking at it very carefully, and we will of course let the House know the results of that review as soon as we can. \\
\addlinespace

\cite{osnabrugge_playing_2021}
&
Emotive rhetoric
& My hon Friend has a distinguished record of more than four years of campaigning hard for local health care services in Redditch and her constituents should be proud of what she has done on their behalf fighting for Redditch hospital and local services I shall be delighted to meet her to talk further about the local challenges for maternity care.\\
\addlinespace

\cite{parthasarathy_deliberative_2019}
&fund allocation
&A meeting was held by the panchayat in which 13 persons were sanctioned OAP under the scheme. But, it has actually come only to just 2 persons and the remaining 11 are yet to get. It is 6 months and no action has been taken though they have been given orders. The money has not come so far. Help should be done to get them the money.\\
\addlinespace

\cite{parthasarathy_deliberative_2019}
&fund allocation
&That is what if we install a new pipe it will be okay. Next repairing of road, building public toilet. For that government is giving 11600. We have been saying nobody is coming forward.\\
\addlinespace

\cite{Schub2022}
&political content (vs. military)
&
We have also considered diplomatic steps which we might take to stabilize the Angola-Zaire border situation. Our concern is that military assistance from French, Belgian, and Moroccan sources being funneled through Zaire to UNITA ma prompt Neto to encourage a resumption of Katangan gendarme attacks aimed at Zaire. My conclusion is that we cannot dissuade the French and Belgians from their view that their long-term interests are served by an ultimate Savimbi victory. Accordingly, we will limit ourselves to again warning Mobutu against diversion of US-supplied equipment. We will also share with him our concerns that his continued interference in Angola, even as an intermediary for others, could jeopardize Congressional support for our present economic and military programs and provoke the Angolans and Katangans into stepping up their activities in Shaba.\\
\addlinespace

\cite{Schub2022}
& political content (vs. military
&Has doubts about the Communists in charge'97CIA has no doubts. Rebels are not all of the same stripe. With [American] troops in the country it is difficult to talk with the rebels. \\
\addlinespace

\end{longtable}

\normalsize

\newpage

\newpage

\section{Additional results for Replication Studies}

\subsection{Additional results for LLM annotations against human annotations}

\subsubsection{Agreement with Human Majority-Vote Labels}\label{app:human_majority}

{
Our main analysis compares each LLM with individual human coders, so that
human and LLM annotations are treated symmetrically and LLM--human agreement
can be benchmarked directly against human--human agreement. In applied
research, however, human annotations are often aggregated, for example by
majority vote or adjudication, before being used as a reference for
validating automated measures. We therefore examine how LLM--human
agreement changes when the human reference is constructed by majority vote.
Specifically, for each text, the majority-vote label is defined as the label selected by at least two of the three human coders. Holding the texts, codebooks, and LLM outputs fixed, we calculate Krippendorff's $\alpha$ between each model's annotations and the majority-vote labels and compare it with the model's average pairwise $\alpha$ with the individual human coders.

\begin{figure}[h]
    \centering
    \includegraphics[width=1.0\linewidth]{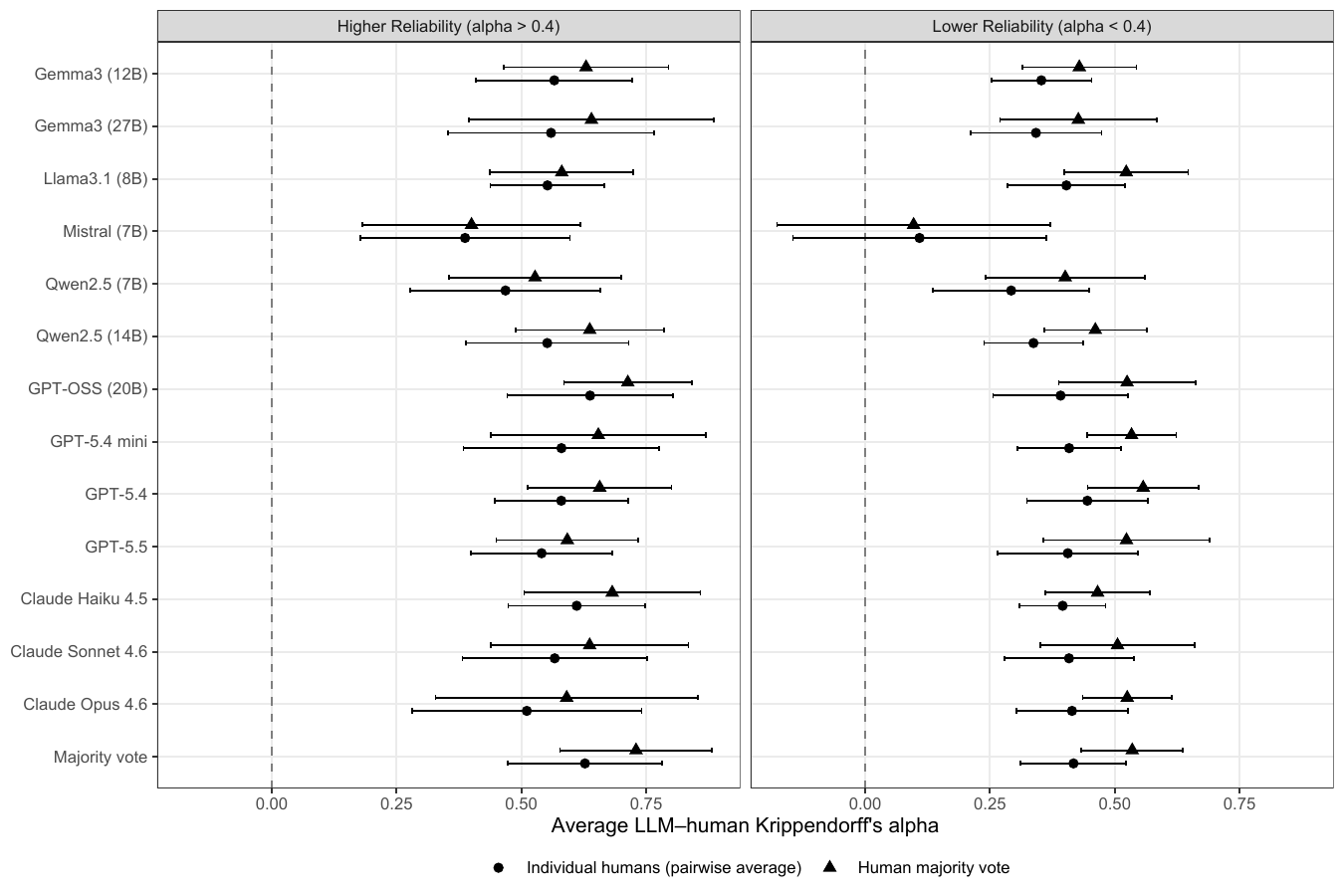}
    \caption{LLM agreement with individual human coders and human
    majority-vote labels, measured by Krippendorff's $\alpha$.
    Circles report each model's average pairwise $\alpha$ with
    the three human coders; triangles report its $\alpha$ with
    the label selected by at least two of the three coders (majority voting).
    Points are averages across studies within each reliability
    group, with horizontal bars indicating 95\% between-study
    confidence intervals. The dashed line marks $\alpha=0$.}
    \label{fig:human_majority}
\end{figure}

Figure~\ref{fig:human_majority} reports the results for studies with
relatively higher and lower human reliability. We find that majority
voting increases LLM--human agreement regardless of which group is
aggregated. Individual LLMs generally agree more closely with the
human majority-vote labels than with individual human coders.
Similarly, the majority vote across LLMs achieves higher agreement
with individual humans than the average individual LLM.
Relative to average pairwise LLM--human agreement, majority voting
across LLMs increases Krippendorff's $\alpha$ by 0.07 on average,
while majority voting across humans increases it by 0.08.
These similar gains suggest that part of the disagreement between
LLMs and humans reflects variation across individual annotators,
which can be reduced through aggregation in either group.
Importantly, greater agreement does not necessarily imply more
accurate annotations, since majority voting may override
substantively defensible minority interpretations.

}

\subsubsection{Within-Model Annotation Stability under Prompt Variation}\label{app:prompt_sensitivity}

Our main replication analysis holds the prompt fixed, using a single structure built from six components (role, topic definition, classification rules, label wording, paper and data description, and a reasoning cue; see Table~\ref{prompt_component}). This lets us compare models and studies on common ground, but raises a concern: we do not know how much our conclusions reflect this particular prompt rather than the task itself. To probe this, we toggle each component on or off to generate all $2^{6}=64$ prompts per paper and run them through seven open-source LLMs (Mistral 7B, LLaMa 3.1-8B, Qwen2.5-7B, Qwen2.5-14B, Gemma 3-12B, Gemma3-27B, GPT-OSS-20B). We test whether this sensitivity reflects \emph{substantive ambiguity} rather than arbitrary instability, by asking whether prompts disagree most on the texts that humans also find hard to code. The fully specified prompt, with all six components included, is our reference point and we refer to it as prompt~64.

The analysis uses all the 64 prompts applied to texts by each model. We measure within-model prompt disagreement using Gini impurity,
which equals zero when all 64 prompts produce the same classification and reaches its maximum of 0.5 when the prompts are evenly divided between labels 0 and 1. We then compare this prompt disagreement between texts on which all three human coders agree and texts on which at least one human assigns a different label. Because human reliability varies across studies, we report the comparison separately for relatively higher reliability studies, defined by human Krippendorff's $\alpha>0.4$, and relatively lower reliability studies, defined by $\alpha<0.4$. For each model and reliability group, the table reports the mean Gini impurity for human-agreement texts and human-disagreement texts, their difference, and the $p$-value from a two-sample comparison of the corresponding text-level Gini values.

Table~\ref{tab:prompt_gini_by_model} shows that within-model prompt disagreement is generally higher for texts on which humans disagree. This pattern is statistically significant for both Gemma3 models, Llama3.1 (8B), both Qwen2.5 models, and GPT-OSS (20B) in both reliability groups. The largest increases among Higher Reliability studies are 0.124 for GPT-OSS (20B) and 0.108 for Qwen2.5 (7B); for GPT-OSS (20B), the increase is 0.122 among Lower Reliability studies. Mistral (7B) is the principal exception: prompt disagreement is lower on human-disagreement texts in both Higher Reliability studies (difference $-0.013$, $p=0.473$) and Lower Reliability studies (difference $-0.027$, $p=0.081$), although neither difference is statistically significant. Overall, the results indicate that prompt sensitivity often tracks human-defined ambiguity, although the strength and direction of this relationship vary across models.

\begin{table}[t]
\caption{\label{tab:prompt_gini_by_model}Within-model prompt disagreement by model, human agreement, and study reliability. For each model and text, Gini impurity is calculated across the 64 prompt-specific predictions. Higher values indicate greater sensitivity to prompt specification.}
\centering
\small
\resizebox{\linewidth}{!}{%
\begin{tabular}[t]{llcccc}
\toprule
Model & Study reliability & Humans agree & Humans disagree & Difference & $p$-value\\
\midrule
Gemma3 (12B) & Higher Reliability ($\alpha>0.4$) & 0.072 & 0.137 & 0.064 & $2.19 \times 10^{-5}$\\
Gemma3 (12B) & Lower Reliability ($\alpha<0.4$) & 0.095 & 0.196 & 0.101 & $1.41 \times 10^{-12}$\\
Gemma3 (27B) & Higher Reliability ($\alpha>0.4$) & 0.058 & 0.157 & 0.098 & $2.57 \times 10^{-8}$\\
Gemma3 (27B) & Lower Reliability ($\alpha<0.4$) & 0.071 & 0.132 & 0.060 & $5.16 \times 10^{-6}$\\
Llama3.1 (8B) & Higher Reliability ($\alpha>0.4$) & 0.084 & 0.181 & 0.097 & $9.78 \times 10^{-8}$\\
\addlinespace
Llama3.1 (8B) & Lower Reliability ($\alpha<0.4$) & 0.108 & 0.235 & 0.127 & $<2.2 \times 10^{-16}$\\
Mistral (7B) & Higher Reliability ($\alpha>0.4$) & 0.186 & 0.173 & -0.013 & $4.73 \times 10^{-1}$\\
Mistral (7B) & Lower Reliability ($\alpha<0.4$) & 0.217 & 0.190 & -0.027 & $8.13 \times 10^{-2}$\\
Qwen2.5 (7B) & Higher Reliability ($\alpha>0.4$) & 0.053 & 0.161 & 0.108 & $1.73 \times 10^{-9}$\\
Qwen2.5 (7B) & Lower Reliability ($\alpha<0.4$) & 0.056 & 0.131 & 0.075 & $2.00 \times 10^{-9}$\\
\addlinespace
Qwen2.5 (14B) & Higher Reliability ($\alpha>0.4$) & 0.037 & 0.135 & 0.098 & $2.08 \times 10^{-9}$\\
Qwen2.5 (14B) & Lower Reliability ($\alpha<0.4$) & 0.040 & 0.111 & 0.071 & $1.27 \times 10^{-9}$\\
GPT-OSS (20B) & Higher Reliability ($\alpha>0.4$) & 0.041 & 0.165 & 0.124 & $2.00 \times 10^{-12}$\\
GPT-OSS (20B) & Lower Reliability ($\alpha<0.4$) & 0.055 & 0.177 & 0.122 & $<2.2 \times 10^{-16}$\\
\bottomrule
\end{tabular}}
\end{table}

To identify which prompt-design choices improve zero-shot classification, we exploit the full set of prompt variants applied across papers and models and estimate the marginal association between individual prompt components and LLM--human agreement. The dependent variable is a model prediction's mean agreement with the three human labels for a given text. The explanatory variables indicate whether the prompt includes each design component---such as an explicit role instruction, a formal concept definition, classification rules, standardized label wording, step-by-step reasoning, or paper/data context. We also control for log text length and include paper and model fixed effects so that the coefficients are identified from differences across prompt specifications within the same studies and model families.

\begin{figure}[htbp]
    \centering
    \includegraphics[width=0.75\linewidth]{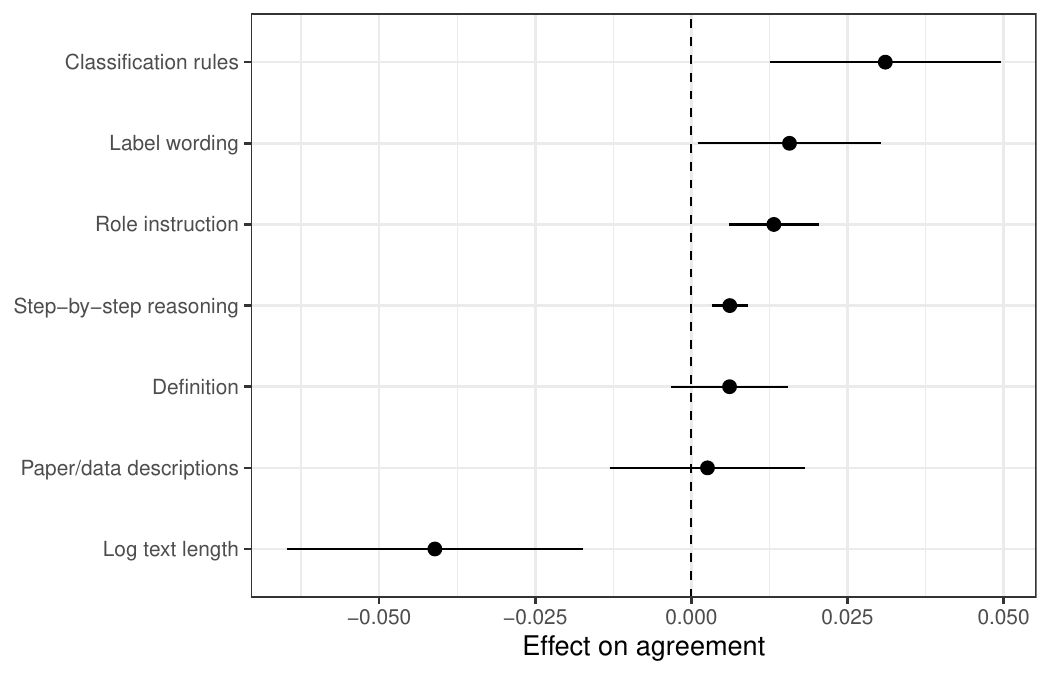}
    \caption{Marginal effects of prompt engineering components and text length on mean LLM--human agreement (OLS coefficient estimates with paper-clustered 95\% confidence intervals).}
    \label{fig:prompt_accuracy_lm}
\end{figure}

An OLS regression (Figure~\ref{fig:prompt_accuracy_lm}) analyzing the effect of prompt features on mean LLM--human agreement reveals that structural instructions provide the largest gains, led by explicit \textit{Classification rules} ($\hat{\beta} \approx +0.031$), standardized \textit{Label wording} ($\hat{\beta} \approx +0.016$), and a \textit{Role instruction} ($\hat{\beta} \approx +0.013$). Step-by-step \textit{reasoning} and topic \textit{definition} each have estimates of approximately $+0.006$, while incorporating \textit{paper/data descriptions} has an estimate near zero ($\hat{\beta} \approx +0.003$). Additionally, document length has a negative association with agreement ($\hat{\beta} \approx -0.041$). The plotted 95\% confidence intervals use CR1 standard errors clustered over the 14 papers.

\subsubsection{Within-Model Annotation Stability under Stochastic Decoding}\label{app:stochastic_decoding}

Our main analysis identifies ambiguous texts using disagreement across different LLMs. However, most applied researchers often use a single model rather than a panel of models. We therefore examine whether repeated runs of the same model provide a useful within-model measure of annotation uncertainty. For each of the seven open-source LLMs, we classify the same 100 texts from each of the 14 studies ten times under stochastic decoding with temperature $1.0$. For each model--text pair, we define instability as an indicator equal to one if the predicted label changes at least once across the ten runs. We then summarize the prevalence of instability by study and model and compare it between texts on which the humans agree and texts on which they disagree.

Figures~\ref{app:paper_stochastic_stability} and~\ref{app:within_stability} show that within-model instability is generally greater in the more difficult studies, but its magnitude varies substantially across models. LLaMA~3.1-8B and GPT-OSS-20B exhibit the greatest instability overall, followed by Mistral-7B, whereas the Gemma~3 and Qwen2.5 models produce substantially more stable annotations. Table~\ref{tab:model-human-agreement} further shows that instability is concentrated on texts that divide humans for GPT-OSS-20B ($0.406$ versus $0.112$), LLaMA~3.1-8B ($0.308$ versus $0.131$), and Qwen2.5-14B ($0.060$ versus $0.010$). All three differences are statistically significant. By contrast, the corresponding differences are small and statistically indistinguishable from zero for Mistral-7B, both Gemma~3 models, and Qwen2.5-7B. These results provide partial support for stochastic-decoding instability as a within-model proxy for ambiguity: it identifies texts that divide humans for some models, but its informativeness depends substantially on the model.

\begin{figure}[h!]
    \centering
    \includegraphics[width=1.0\linewidth]{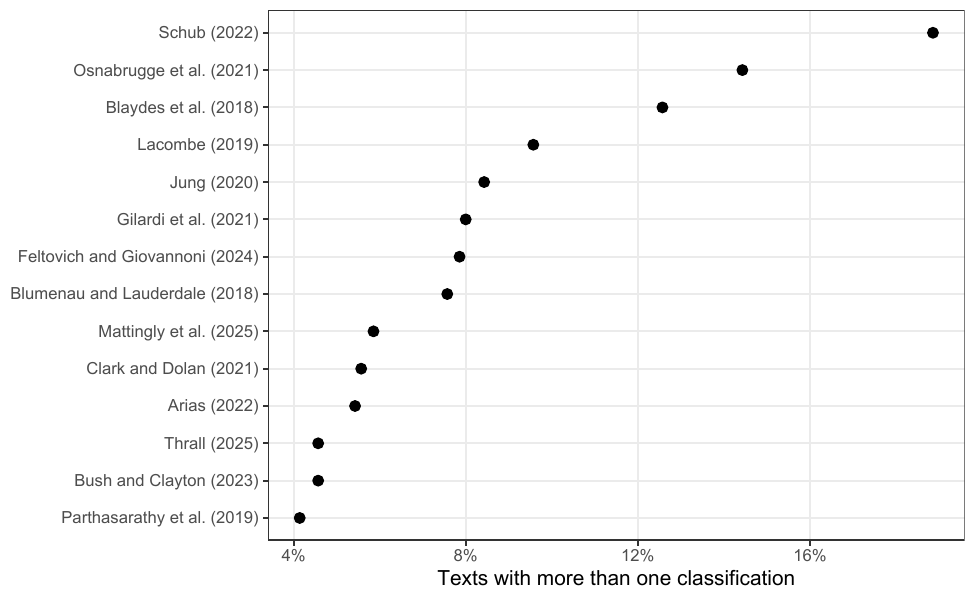}
    \caption{Instability of Text Annotations across seven open-source models for each paper. Instability is measured as the proportion of 100 randomly sampled texts that receive more than one classification label across 10 repeated runs under stochastic decoding (temperature = 1.0). The seven open-source models are GPT-OSS (20B), Gemma 3 (12B and 27B), Llama 3.1 (8B), Mistral (7B), and Qwen 2.5 (7B and 14B).}
    \label{app:paper_stochastic_stability}
\end{figure}

\begin{figure}[ht!]
    \centering
    \includegraphics[width=0.7\linewidth]{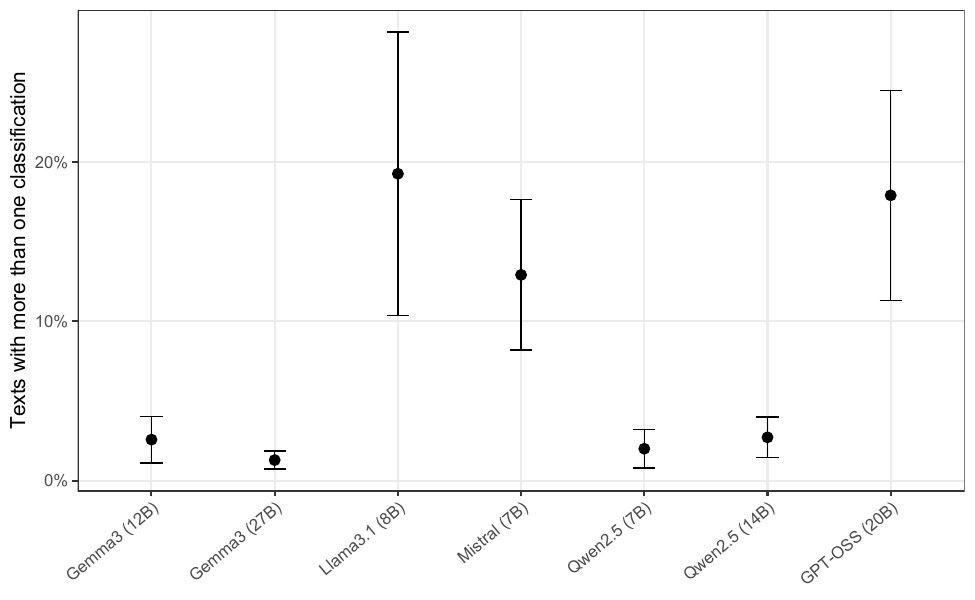}
    \caption{Instability of Text Annotations for each open-source model across 14 papers. Instability is measured as the proportion of 100 randomly sampled texts that receive more than one classification label across 10 repeated runs under stochastic decoding (temperature = 1.0).}
    \label{app:within_stability}
\end{figure}

\begin{table}[h!]
\centering
\caption{Annotation Variation for each model under Stochastic Decoding (Temperature = 1) Across 10 Runs, Stratified by Human Agreement.}
\label{tab:model-human-agreement}
\begin{tabular}{lrrrr}
\toprule
Model
& Humans agree
& Humans disagree
& Difference
& $p$-value \\
\midrule
GPT-OSS (20B)  & 0.112 & 0.406 &  0.293  & <0.001 \\
Llama3.1 (8B)  & 0.131 & 0.308 &  0.177  & 0.003 \\
Qwen2.5 (14B)  & 0.010 & 0.060 &  0.050  & 0.002 \\
Mistral (7B)   & 0.119 & 0.150 &  0.031  & 0.436 \\
Gemma3 (12B)   & 0.021 & 0.033 &  0.012  & 0.257 \\
Gemma3 (27B)   & 0.009 & 0.020 &  0.011  & 0.142 \\
Qwen2.5 (7B)   & 0.020 & 0.019 & -0.001  & 0.934 \\
\bottomrule
\end{tabular}
\end{table}

\subsubsection{Additional results under an alternative measure of within-text disagreement}\label{app:entropy_expert}

In Section~\ref{sec:disagreement_gini} and \ref{app:prompt_sensitivity}, we used Gini impurity to measure the text-level annotation heterogeneity. As a robustness check, we replicate the same results using the normalized Shannon entropy, which is defined as
\[
H=-p\log_{2}(p)-(1-p)\log_{2}(1-p),
\]
where \(p\) denotes the proportion of LLMs assigning the positive label. In this binary classification setting, \(H\) ranges from \(0\) to \(1\): \(H=0\) indicates complete agreement across models, whereas \(H=1\) indicates a perfectly even split between positive and negative predictions. We calculate entropy separately for each text across the available LLM predictions and then compare the average entropy of texts on which human coders agree with that of texts on which human coders disagree. Higher average entropy therefore indicates that the models are more divided in their interpretation of the same texts.

Table~\ref{tab:gini_disagreement_entropy} reports the same analysis as in Table~\ref{tab:gini_disagreement} using Shannon entropy. This shows that cross-model disagreement is substantially greater on texts where humans disagree. Among relatively higher reliability studies, mean entropy rises from \(0.178\) when humans agree to \(0.551\) when they disagree, a difference of \(0.372\) (\(p<0.001\)). The same pattern appears among the relatively lower reliability studies, where entropy increases from \(0.201\) to \(0.533\), a difference of \(0.332\) (\(p<0.001\)). This suggests that LLMs are especially divided on texts that are also difficult for humans to classify.

\begin{table}[h!]
\centering
\caption{Cross-model disagreement by human agreement and study reliability using entropy.
Higher values indicate greater disagreement among the 13 LLMs.}
\label{tab:gini_disagreement_entropy}
\setlength{\tabcolsep}{4pt}
\resizebox{\linewidth}{!}{%
\begin{tabular}{lcccccc}
\toprule
Study reliability
    & Humans agree
    & Humans disagree
    & Difference
    & $p$-value
    & $N$ agree
    & $N$ disagree \\
\midrule
Higher Reliability ($\alpha>0.4$)
    & 0.178
    & 0.551
    & 0.372
    & $<2.2 \times 10^{-16}$
    & 565
    & 135 \\
Lower Reliability ($\alpha<0.4$)
    & 0.201
    & 0.533
    & 0.332
    & $<2.2 \times 10^{-16}$
    & 440
    & 260 \\
\bottomrule
\end{tabular}}
\end{table}

On the other hand, Table~\ref{tab:prompt_entropy_by_model} replicates the results of within-model prompt disagreement in Table~\ref{tab:prompt_gini_by_model} using Shannon's entropy. This shows that, for most models, within-model prompt disagreement is substantially higher on texts where humans disagree than on texts where humans agree. This pattern is statistically significant for both Gemma3 models, Llama3.1 (8B), both Qwen2.5 models, and GPT-OSS (20B) in both reliability groups. The largest differences among the higher reliability studies are \(0.277\) for GPT-OSS (20B) and \(0.237\) for Qwen2.5 (7B). Mistral is the main exception: its prompt disagreement is lower on human-disagreement texts in both reliability groups, with the difference reaching statistical significance only among the lower reliability studies. Overall, the results indicate that prompt sensitivity generally concentrates on texts that are also difficult for humans to classify, although this relationship varies across model families.

\begin{table}[t]
\caption{\label{tab:prompt_entropy_by_model}Within-model prompt disagreement by model, human agreement, and study reliability. For each model and text, normalized Shannon entropy is calculated across the 64 prompt-specific predictions. Higher values indicate greater sensitivity to prompt specification.}
\centering
\small
\resizebox{\linewidth}{!}{%
\begin{tabular}[h]{llcccc}
\toprule
Model & Study reliability & Humans agree & Humans disagree & Difference & $p$-value\\
\midrule
Gemma3 (12B) & Higher Reliability ($\alpha>0.4$) & 0.170 & 0.321 & 0.151 & $3.00 \times 10^{-6}$\\
Gemma3 (12B) & Lower Reliability ($\alpha<0.4$) & 0.211 & 0.438 & 0.226 & $1.73 \times 10^{-14}$\\
Gemma3 (27B) & Higher Reliability ($\alpha>0.4$) & 0.139 & 0.348 & 0.210 & $1.17 \times 10^{-8}$\\
Gemma3 (27B) & Lower Reliability ($\alpha<0.4$) & 0.162 & 0.294 & 0.132 & $1.94 \times 10^{-6}$\\
Llama3.1 (8B) & Higher Reliability ($\alpha>0.4$) & 0.195 & 0.404 & 0.209 & $2.31 \times 10^{-8}$\\
\addlinespace
Llama3.1 (8B) & Lower Reliability ($\alpha<0.4$) & 0.239 & 0.511 & 0.272 & $<2.2 \times 10^{-16}$\\
Mistral (7B) & Higher Reliability ($\alpha>0.4$) & 0.412 & 0.383 & -0.029 & $4.34 \times 10^{-1}$\\
Mistral (7B) & Lower Reliability ($\alpha<0.4$) & 0.475 & 0.413 & -0.062 & $4.87 \times 10^{-2}$\\
Qwen2.5 (7B) & Higher Reliability ($\alpha>0.4$) & 0.121 & 0.358 & 0.237 & $2.87 \times 10^{-10}$\\
Qwen2.5 (7B) & Lower Reliability ($\alpha<0.4$) & 0.130 & 0.298 & 0.167 & $2.29 \times 10^{-10}$\\
\addlinespace
Qwen2.5 (14B) & Higher Reliability ($\alpha>0.4$) & 0.088 & 0.306 & 0.218 & $3.44 \times 10^{-10}$\\
Qwen2.5 (14B) & Lower Reliability ($\alpha<0.4$) & 0.091 & 0.256 & 0.165 & $3.13 \times 10^{-11}$\\
GPT-OSS (20B) & Higher Reliability ($\alpha>0.4$) & 0.095 & 0.372 & 0.277 & $5.81 \times 10^{-14}$\\
GPT-OSS (20B) & Lower Reliability ($\alpha<0.4$) & 0.126 & 0.393 & 0.267 & $<2.2 \times 10^{-16}$\\
\bottomrule
\end{tabular}}
\end{table}

\subsubsection{Additional analysis on predicting disagreement}\label{sec:predict_disagreement}
{

To investigate which textual characteristics predict annotation
disagreement, we consider six candidate predictors motivated by
research on readability and linguistic ambiguity: (i) text length
\citep{pitler2008revisiting}, (ii) lexical rarity or vocabulary
difficulty \citep{pitler2008revisiting}, (iii) hedging and
conditional-language frequency
\citep{vincze2008bioscope,zhang2021identifying},
(iv) referential ambiguity
\citep{jiang2022investigating,poesio2005reliability},
(v) negation-marker frequency
\citep{vincze2008bioscope,zhang2021identifying}, and
(vi) textual imperfections \citep{jiang2022investigating}.
Table~\ref{tab:disagreement_predictors} describes each feature
and its operationalization.

We estimate separate regressions predicting text-level disagreement
among human coders and across LLMs measured by Gini impurity, including all six predictors
and paper fixed effects as regressors. These models examine how each feature
is associated with disagreement among texts from the same paper,
holding the other features constant. Paper fixed effects account
for differences in baseline disagreement across papers, including
those related to the substantive topic and codebook.

\begin{table}[!ht]
\centering
\small
\caption{Text-level features used to predict annotation disagreement}
\label{tab:disagreement_predictors}
\begin{tabular}{p{4.0cm} p{5.0cm} p{6.5cm}}
\toprule
\textbf{Feature} & \textbf{Description} & \textbf{Operationalization} \\
\midrule

Text length
\citep{pitler2008revisiting}
&
Amount of textual information that an annotator must process.
&
Log number of word tokens in the text,
$\log(1+\text{number of tokens})$.
\\

\addlinespace

Lexical rarity
\citep{pitler2008revisiting}
&
Extent to which the text contains uncommon or difficult vocabulary.
&
Mean negative log unigram frequency of the words in the text,
using frequencies from an external reference corpus. Higher values
indicate less common vocabulary.
\\

\addlinespace

Hedging and conditional language
\citep{vincze2008bioscope, zhang2021identifying}
&
Language that weakens, qualifies, or makes a proposition conditional,
potentially leaving its interpretation less determinate.
&
Number of predefined hedge, modal, and conditional markers
(e.g., \textit{may}, \textit{might}, \textit{could}, \textit{perhaps},
\textit{likely}, \textit{if}, \textit{unless}) per 100 word tokens.
\\

\addlinespace

Multiple plausible referents
\citep{jiang2022investigating, poesio2005reliability}
&
Referring expressions for which more than one preceding entity could
plausibly serve as the antecedent.
&
For each third-person pronoun or other anaphoric expression, count
candidate noun-phrase antecedents in the same or preceding sentence.
Measure the proportion of referring expressions with two or more
candidate antecedents.
\\

\addlinespace

Negation
\citep{vincze2008bioscope, zhang2021identifying}
&
Explicit linguistic expressions that reverse or restrict the polarity
of a proposition.
&
Number of predefined negation markers
(e.g., \textit{no}, \textit{not}, \textit{n't}, \textit{never},
\textit{none}, \textit{neither}, \textit{nor}, \textit{without})
per 100 word tokens.
\\

\addlinespace

Imperfection
\citep{jiang2022investigating}
&
Surface defects that may make the intended meaning of a text
difficult to recover, such as typographical errors, malformed text,
or incomplete sentences.
&
Composite measure of mechanically detectable textual defects,
including misspelled or out-of-vocabulary tokens, sentence fragments
(e.g., sentences without a finite verb), and encoding or truncation
artifacts. Each component is normalized by text length before
aggregation.
\\

\bottomrule
\end{tabular}
\end{table}

Figure~\ref{fig:disagreement_predictors} provides limited evidence
of systematic associations between the six literature-motivated
features and annotation disagreement. Although the estimated
coefficients generally have the same signs in the human and
cross-LLM regressions, most are not statistically distinguishable
from zero. The only statistically significant association is
between log text length and human disagreement: longer texts
are associated with greater disagreement among human coders.
This finding is consistent with earlier evidence that longer
news articles are perceived as harder to read
\citep{pitler2008revisiting}. The corresponding coefficient for
cross-LLM disagreement is also positive but is not statistically
significant. One possible interpretation is that longer texts
place greater demands on human coders, although these regressions
do not identify the underlying mechanism or establish that the
association differs between humans and LLMs.

Overall, the evidence on general textual predictors of disagreement
remains inconclusive. One possible explanation is that ambiguity
depends on how a particular passage relates to the task's coding
rules, rather than on general linguistic features alone.
The examples in Table~\ref{tab:examples_ambiguous} illustrate
this possibility. For example, in our replication of
\citet{arias2022securitizes}, a passage describing threats to a
country's survival raises the question of whether existential
language alone constitutes climate securitization. In our
replication of \citet{osnabrugge_playing_2021}, expressions of
thanks and praise raise a different question: whether conventional
parliamentary courtesies qualify as emotive rhetoric.
These distinctions may not be captured by the general textual
features considered here. We therefore conjecture that predicting
ambiguity across diverse annotation tasks requires attention to
task-specific category boundaries and how they apply to individual
texts. We leave a systematic investigation of these possibilities to future research and generally recommend that researchers address these issues by refining the codebook and accounting for residual ambiguity.

\begin{figure}[!ht]
    \centering
    \includegraphics[width=1.0\linewidth]{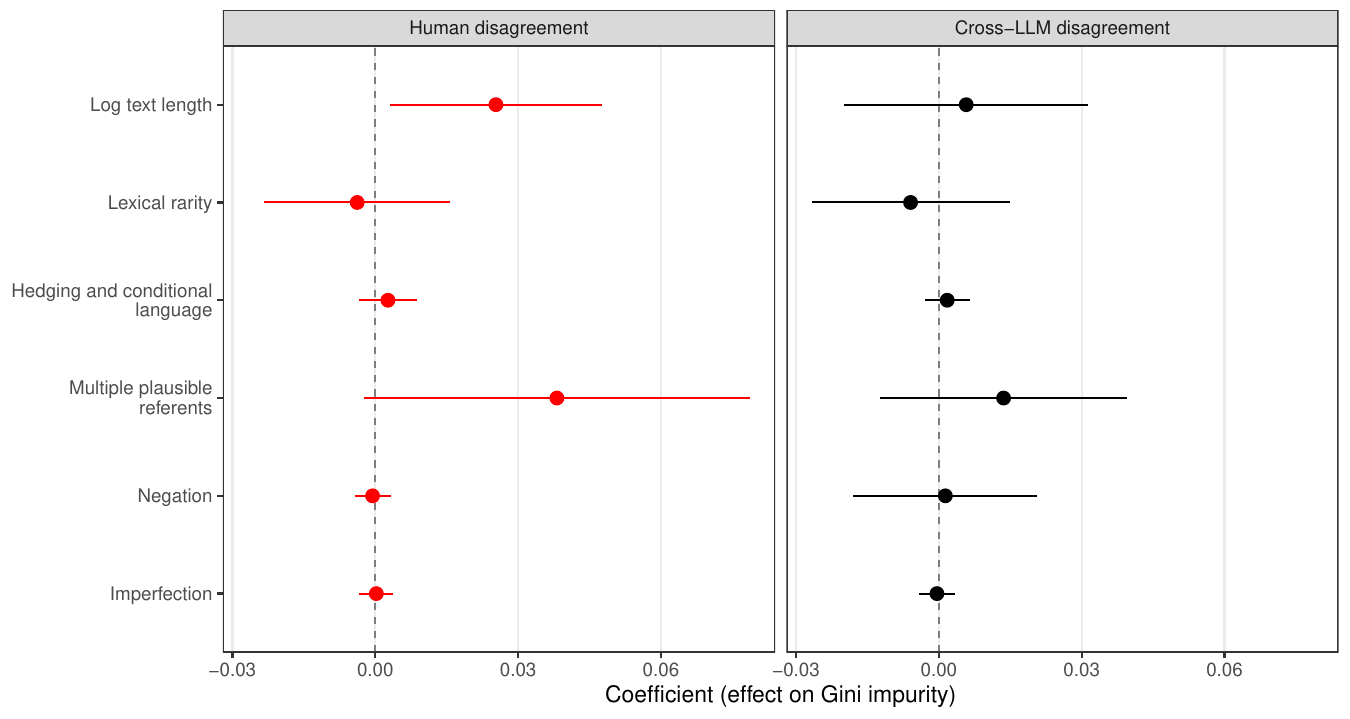}
    \caption{Coefficient of each predictors predicting annotation disagreement among
    human coders (left) and across LLMs (right). Points report
    coefficient estimates from separate regressions of text-level
    Gini impurity on all six textual features and paper fixed
    effects. Horizontal bars indicate 95\% confidence intervals,
    and dashed vertical lines mark zero. Positive coefficients
    indicate associations with greater disagreement, holding the
    other features and paper fixed effects constant.
    Table~\ref{tab:disagreement_predictors} describes the
    operationalization of each feature.}
    \label{fig:disagreement_predictors}
\end{figure}

}

\newpage 
\subsection{Additional results for Crowd Sourced Annotation}\label{app:crowd_results}

This section elaborates on the results from the crowd-sourced annotation, which affords a blinded, preregistered test of the paper's central claims. This section proceeds with reporting the results from the preregistered hypotheses for the experiment, followed by robustness checks, and alternative measures of intercoder agreement. 

\subsubsection{Preregistered Hypotheses}\label{app:crowd_prereg}

We preregistered six hypotheses that extend the paper's theory from human and LLM coders to a third set of coders: crowd workers. Among the six, three confirmatory hypotheses restate the paper's core predictions for the crowd. 
\begin{itemize}
    \item \textbf{H1 (text difficulty)}: When LLMs agree with one another, crowd-sourced workers are more likely to agree with each other. In other words, the Gini impurity of LLMs is correlated with that of crowd-sourced workers. This maps on to the main paper's Hypothesis~4a that disagreement reflects textual ambiguity rather than coder identity. 
    \item \textbf{H2 (crowd vs.\ LLM agreement with humans)}: The average agreement of crowd-sourced workers with three humans (authors’ coding) is lower than that of LLMs with three humans. In other words, the average LLM-human agreement is significantly higher than the average crowd-human agreement. This maps on to the main paper's Hypothesis~4b: LLMs align with humans better than crowd coders.
    \item \textbf{H3 (codebook improvement)}: Using an improved codebook decreases the Gini impurity among crowd-sourced workers (\textbf{H3a}) and increases the average agreement of crowd-sourced workers with humans (\textbf{H3b}). We test this on the original and improved Schub (2022) codebooks over the same 20 texts. This maps on to the main paper's Hypothesis~3: clearer coding rules reduce disagreement.

\end{itemize} 

We further probe three exploratory hypotheses on coder-and-text level sources of agreement as moderators. We preregister these as exploratory hypotheses given these analyses involve slicing an already small sample size. A full copy of our preregistered hypotheses, research design, and analysis plan are available on OSF, at \url{https://osf.io/5wp2k}.

\begin{itemize}
    \item \textbf{H4 (coder knowledge):} We test whether crowd--human agreement is higher among workers with higher baseline knowledge of politics/policy.(\textbf{H4a}), higher frequency of political news consumption, (\textbf{H4b}), or higher education levels (\textbf{H4c}).
    \item \textbf{H5 (coder attention):} We test whether crowd--human agreement is higher on texts workers spend longer on (\textbf{H5a}) or on shorter texts (\textbf{H5b}).
    \item \textbf{H6 (text heterogeneity)}: Crowd Gini impurity is lower on less contested texts than on more contested ones, with strata defined by LLM disagreement (\textbf{H6a}) or human unanimity (\textbf{H6b}). This maps on to the main paper's Hypothesis~4a, similar to H1 but contrasting texts across strata.
\end{itemize} 

% Table~\ref{tab:all_preregistered_tests} reports the full estimates, confidence intervals, and $p$-values; the three confirmatory tests are corrected together for the false-discovery rate (Benjamini--Hochberg), with H1 two-sided and H2 and H3 one-sided.

For all survey results, we use the 20-text-per-paper survey sample described in Appendix~\ref{app:crowd_procedure}. Because this subsample oversamples contested texts by design, agreement and reliability levels on it are generally lower than in the full corpus. We conduct multiple hypothesis testing for the three confirmatory hypotheses (four tests: H1, H2, H3a, H3b) using the Benjamini--Hochberg correction. We report the summary of findings for each hypothesis in Table~\ref{tab:preregistered_tests}. Below, we walk through each result in turn.

\begin{table}[H]
\centering
\caption{Preregistered tests: estimates, 95\% confidence intervals, $p$-values, and support. The four confirmatory tests share a Benjamini--Hochberg correction ($p_{\mathrm{BH}}$); exploratory tests are uncorrected.}
\label{tab:preregistered_tests}
\footnotesize
\setlength{\tabcolsep}{4pt}
\sbox0{\begin{tabular}{@{}llccccc@{}}
\toprule
Hypothesis & Test ($N$) & Est. & 95\% CI & $p$ & $p_{\mathrm{BH}}$ & Support \\
\midrule
\multicolumn{7}{@{}l}{\emph{Panel A: Confirmatory (Benjamini--Hochberg family)}} \\
\addlinespace[2pt]
H1: Crowd tracks LLM disagreement & OLS slope (280) & $0.339$ & $[0.256,\, 0.422]$ & $<$ 0.001 & $<$ 0.001 & $\checkmark$ \\
H2: LLM closer to humans ($\alpha$) & Paired $t$ (14) & $0.129$ & $[0.033,\, 0.226]$ & 0.006 & 0.008 & $\checkmark$ \\
H3a: Better codebook, less disagreement & Paired $t$ (20) & $-0.131$ & $[-0.227,\, -0.034]$ & 0.005 & 0.008 & $\checkmark$ \\
H3b: Better codebook, closer to humans & Paired $t$ (20) & $0.055$ & $[-0.061,\, 0.171]$ & 0.167 & 0.167 & ($+$) n.s. \\
\midrule
\multicolumn{7}{@{}l}{\emph{Panel B: Exploratory tests (no correction)}} \\
\addlinespace[2pt]
H4a: Political knowledge & OLS (165) & $0.063$ & $[-0.010,\, 0.137]$ & 0.090 & --- & ($+$) n.s. \\
H4b: News consumption & OLS (165) & $0.063$ & $[0.009,\, 0.117]$ & 0.022 & --- & ($+$)$^{\dagger}$ \\
H4c: Education & OLS (165) & $0.051$ & $[0.005,\, 0.096]$ & 0.030 & --- & ($+$)$^{\dagger}$ \\
H5a: Total duration (per 10 min) & OLS (165) & $-0.026$ & $[-0.0538,\, 0.0011]$ & 0.060 & --- & ($-$)$^{\ddagger}$ \\
H5b: Text length (per 100 words) & OLS (280) & $-0.010$ & $[-0.021,\, 0.001]$ & 0.067 & --- & ($-$) n.s. \\
H6a: LLM-contested $-$ uncontested Gini & Welch $t$ (280) & $0.137$ & $[0.100,\, 0.173]$ & $<$ 0.001 & --- & $\checkmark$ \\
H6b: Human-split $-$ human-agree Gini & Welch $t$ (280) & $0.101$ & $[0.064,\, 0.138]$ & $<$ 0.001 & --- & $\checkmark$ \\
\bottomrule
\end{tabular}}
\usebox0
\par\smallskip
\begin{minipage}{\wd0}
\footnotesize\emph{Notes:} $\checkmark$ supported; ($+$)/($-$), signed as predicted but not significant; n.s., not significant. $^{\dagger}$ Significant only without task fixed effects. $^{\ddagger}$ Sign opposite to the preregistered prediction but largely null.
\end{minipage}
\end{table}

\subsubsection{Survey Finding 1: Textual ambiguity drives crowd disagreement (H1 and H6).}
First, to test H1, we estimate an OLS regression of each text's crowd Gini impurity on its 13-model LLM-panel Gini impurity across all 280 texts. We exclude the improved Schub codebook for comparability. Here, a positive slope with 95\% confidence interval excluding zero means that crowd disagreement correlates positively with LLM disagreement. We use three different specifications: naive OLS, including cluster-robust errors by paper, and including both paper fixed effects and cluster-robust errors.

Table~\ref{tab:h1_specifications} reports the results. We find that crowd disagreement concentrates on exactly the texts that humans and LLMs also find ambiguous. We observe a positive and statistically significant slope: each text's crowd Gini impurity has a positive slope when regressed on LLM Gini impurity, with a coefficient of $0.34$ (95\% CI $[0.26, 0.42]$) for the naive OLS. This finding is consistent across specifications with similar estimates, which shows that it is not an artifact of which papers happen to contain more difficult texts.

\begin{table}[H]
\centering
\caption{H1, Regression of crowd disagreement on LLM-panel disagreement}
\label{tab:h1_specifications}
\small
\sbox0{\begin{tabular}{lccc}
\toprule
 & (1) & (2) & (3) \\
 & OLS & CR2 by paper & Paper FE + CR2 \\
\midrule
LLM Gini & 0.339 & 0.339 & 0.334 \\
 & [0.256, 0.422] & [0.193, 0.484] & [0.187, 0.482] \\
$p$ & $<$ 0.001 & $<$ 0.001 & $<$ 0.001 \\
\midrule
Paper fixed effects & No & No & Yes \\
$N$ (texts) & 280 & 280 & 280 \\
$N$ (papers) & 14 & 14 & 14 \\
\bottomrule
\end{tabular}}
\usebox0
\end{table}

We probe this finding in a different way in H6 by splitting texts into less versus more contested strata, based on LLM and human disagreement. This follows the same procedure as the main paper's analyses (see Section~\ref{sec:disagreement_gini}). To briefly reiterate: in H6a, disagreement among LLMs is an even split, incorporated by design given that we stratify-sample texts based on the 10 highest and lowest Gini impurity texts for each paper (computed across the original 10 LLM models). In H6b, disagreement among humans is considered ``high'' if humans are split in their coding and ``low'' for unanimous coding. We then compare mean crowd Gini impurity between the less and more contested strata with a Welch $t$-test. We find that crowd disagreement rises from $0.24$ on the least LLM-contested texts to $0.38$ on the most contested, and from $0.27$ where the humans agree to $0.37$ where they split (Table~\ref{tab:h6-strata}).
%More visually, Figure~\ref{fig:h1_replication} in the main paper shows how crowd disagreement is markedly higher when LLMs or humans disagree on those same texts. 
In all, from both H1 and H6, we demonstrate that crowd disagreement tracks the same underlying textual ambiguity that divides LLMs or humans.

\begin{table}[H]
\centering
\caption{H6, Crowd disagreement by LLM sampling stratum and by human unanimity.}
\label{tab:h6-strata}
\small
\sbox0{\begin{tabular}{@{}lcc@{}}
\toprule
 & $n$ texts & Mean crowd Gini (95\% CI) \\
\midrule
\multicolumn{3}{@{}l}{\emph{Panel A: By LLM sampling stratum}} \\
\quad Least contested (LLM agree) & 140 & 0.242 $[0.213,\, 0.271]$ \\
\quad Most contested (LLM disagree) & 140 & 0.379 $[0.357,\, 0.402]$ \\
\quad Difference (most $-$ least) & & $+0.137$ $[0.100,\, 0.173]$ \\
\addlinespace[2pt]
\multicolumn{3}{@{}l}{\emph{Panel B: By human unanimity}} \\
\quad Humans unanimous & 172 & 0.272 $[0.245,\, 0.298]$ \\
\quad Humans split & 108 & 0.373 $[0.347,\, 0.398]$ \\
\quad Difference (split $-$ unanimous) & & $+0.101$ $[0.064,\, 0.138]$ \\
\bottomrule
\end{tabular}}
\usebox0
\end{table}

\subsubsection{Survey Finding 2: Human-LLM agreement beats human-crowd agreement (H2)}

Second, to test our survey's H2, we use a one-sided paired $t$-test across the 14 papers, pairing each paper's mean LLM--human Krippendorff's $\alpha$ against its mean crowd--human $\alpha$. Each coder-pair $\alpha$ is estimated over the paper's 20 texts and then averaged across pairs. Here, we exclude the improved codebook for Schub for comparability. We find a substantial average gap between crowd--human and LLM--human $\alpha$, favoring LLMs for most models and papers.

Table~\ref{tab:h2_by_paper} reports the results. We find that the LLM panel is closer to the humans than the crowd in 11 of the 14 papers, with three reversals. Mean LLM--human $\alpha$ is $0.329$, compared with $0.199$ for crowd--human $\alpha$; the mean gap is $0.129$ (95\% CI $[0.033, 0.226]$, one-sided $p=0.006$). This gap holds for most LLM models in both the Higher Reliability and Lower Reliability paper groups (Figure~\ref{fig:app_sources_by_difficulty}). This confirms the prediction in H2: although crowd workers track ambiguity in the same way that LLMs and humans do, crowd workers show lower agreement with humans than LLMs on average. Our results corroborate existing works that find LLMs outperforming crowd workers on a range of text-annotation or related tasks \citep{gilardi_chatgpt_2023, ziems2024llm}, and extend them by showing how ``outperforming'' can be measured by agreement without defining a ground truth. We further show how the LLM-crowd-worker gap applies across a wide set of papers, across task and text difficulty, and LLM model types.

\begin{figure}[H]
    \centering
    \includegraphics[width=\textwidth]{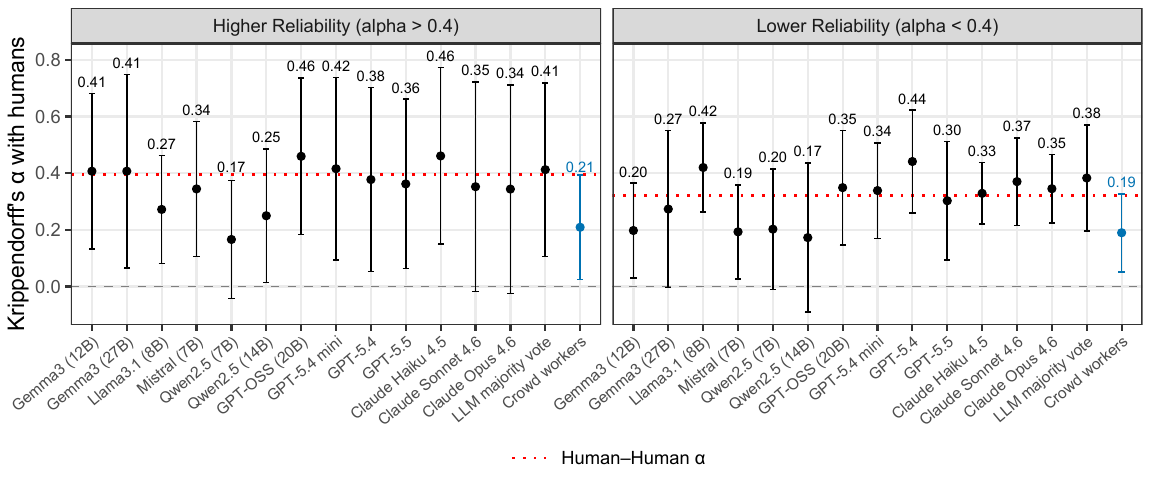}
    \caption{Krippendorff's $\alpha$ with human coders on the 20-text survey sample, for each LLM, their majority vote, and the crowd workers (in blue). Points average coder-pair $\alpha$ with the three humans within papers and then across papers; the red dotted line is the mean three-coder human $\alpha$. Levels are lower than in the full corpus because the survey sample oversamples contested texts.}
    \label{fig:app_sources_by_difficulty}
\end{figure}

\begin{table}[H]
\centering
\caption{H2, LLM--human and crowd--human Krippendorff's $\alpha$ by paper}
\label{tab:h2_by_paper}
\small
\sbox0{\begin{tabular}{lccc}
\toprule
Paper & LLM--Human $\alpha$ & Crowd--Human $\alpha$ & Gap (LLM $-$ Crowd) \\
\midrule
\quad \cite{feltovich_campaign_2024} & 0.429 & 0.261 & $+0.168$ \\
\quad \cite{bush_facing_2023} & 0.658 & 0.316 & $+0.342$ \\
\quad \cite{thrall_informational_2025} & 0.426 & 0.372 & $+0.054$ \\
\quad \cite{mattingly_chinese_2025} & 0.215 & 0.432 & $-0.217$ \\
\quad \cite{ClarkDolan2021} & 0.531 & 0.363 & $+0.168$ \\
\quad \cite{arias2022securitizes} & 0.359 & 0.015 & $+0.344$ \\
\quad \cite{BlumenauLauderdale2018} & 0.273 & 0.344 & $-0.072$ \\
\quad \cite{parthasarathy_deliberative_2019} & 0.318 & 0.083 & $+0.236$ \\
\quad \cite{jung_mobilizing_2020} & 0.397 & 0.318 & $+0.078$ \\
\quad \cite{blaydes_mirrors_2018} & 0.231 & 0.054 & $+0.178$ \\
\quad \cite{lacombe2019political} & -0.159 & -0.153 & $-0.006$ \\
\quad \cite{gilardi_policy_2021} & 0.502 & 0.121 & $+0.381$ \\
\quad \cite{Schub2022}, [original codebook] & 0.197 & 0.139 & $+0.058$ \\
\quad \cite{osnabrugge_playing_2021} & 0.229 & 0.128 & $+0.101$ \\
\midrule
Mean (14 papers) & 0.329 & 0.199 & $+0.129$ \\
\bottomrule
\end{tabular}}
\usebox0
\end{table}

\subsubsection{Survey Finding 3: A clearer codebook improves crowd coding (H3)}

Third, to test H3, we conduct a one-sided paired $t$-test across the 20 texts from Schub, assessing changes from the original codebook to the improved codebook, which includes clearer classification rules and examples. The two Schub codebooks are randomly assigned as separate tasks in our survey, while respondents code the same 20 texts regardless of which codebook they receive.

Figure~\ref{fig:app_codebook_2panel} visualizes changes in Gini impurity and Krippendorff's $\alpha$ (see Table~\ref{tab:schub-codebook} for the preregistered tests). The $\alpha$ values for crowd workers measure their average pairwise agreement with our three human coders, whereas Gini impurity measures disagreement within the crowd.  We find that sharpening the codebook improves agreement and reduces disagreement among crowd workers, consistent with our findings for LLMs and humans. For H3a, Gini impurity falls from $0.42$ to $0.29$, indicating that the improved codebook reduces disagreement. For H3b, mean crowd--human percent agreement rises from $0.60$ to $0.66$ under the improved codebook. Both findings are consistent with our preregistered expectations, although only the result for H3a is statistically significant at the 5\% level. Thus, the improved codebook leads crowd workers to agree more with one another and with humans, supporting the paper's central conclusion that clearer coding rules can reduce ambiguity and disagreement.

\begin{table}[H]
\centering
\caption{H3, Effects of codebook improvement for Schub (2022)}
\label{tab:schub-codebook}
\small
\sbox0{\begin{tabular}{lcccc}
\toprule
 & Original & Improved & & \\
 & codebook (v1) & codebook (v2) & Difference & $p$ \\
\midrule
Crowd Gini impurity (H3a) & 0.418 & 0.288 & $-0.131$ & 0.005 \\
Crowd--human agreement (H3b) & 0.603 & 0.658 & $+0.055$ & 0.167 \\
Krippendorff's $\alpha$ (crowd) & 0.047 & 0.355 & $+0.308$ & --- \\
Crowd workers ($n$) & 10 & 12 & & \\
\bottomrule
\end{tabular}}
\usebox0
\par\smallskip
\begin{minipage}{\wd0}
\footnotesize\emph{Notes:} H3a and H3b are the preregistered tests; Krippendorff's $\alpha$ is descriptive and serves as an alternative measure to agreement.
\end{minipage}
\end{table}

\begin{figure}[H]
    \centering
    \includegraphics[width=\textwidth]{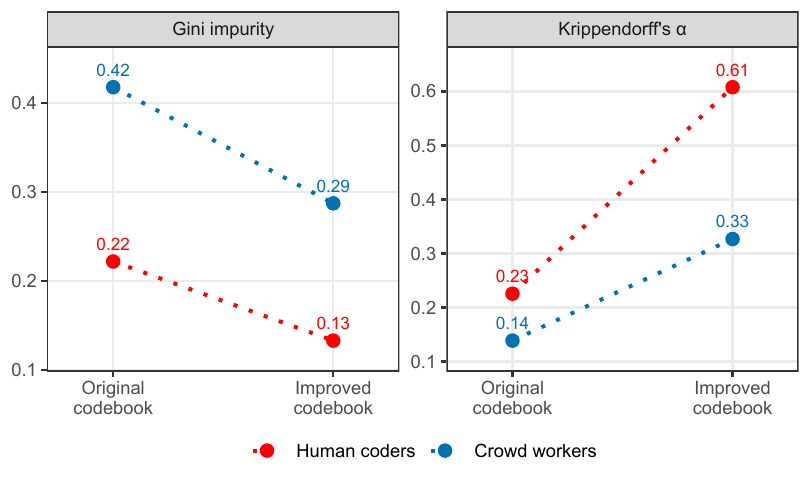}
    \caption{(Left:) Gini impurity, measure of within-source disagreement; fall in Gini corresponds to H3a (Better codebook, less disagreement). (Right:) Krippendorff's $\alpha$ for crowd workers with humans (mean pairwise crowd--human $\alpha$) and humans with one another; directionally consistent with H3b (Better codebook, closer to humans). Table~\ref{tab:schub-codebook} instead reports within-crowd $\alpha$.}
    \label{fig:app_codebook_2panel}
\end{figure}

\subsubsection{Survey Finding 4: Coder characteristics do not predict annotation quality (H4/H5).}

Finally, we test whether coder characteristics predict annotation quality (H4 and H5). We estimate H4 (coder knowledge as moderator) by respondent-level regressions of each worker's mean crowd--human agreement on their political knowledge, news frequency, and education, each binarized given the modest sample size. For hypothesis 5 (coder attention as moderator), we regress each worker's mean crowd--human agreement on their total survey duration (H5a) and per-text crowd--human agreement on text length (H5b). Here, H5a uses total survey time rather than the preregistered per-text time. This is because we could not recover the per-text time spent since we programmed our survey to randomize text order. We run all regressions with two specifications: a univariate OLS with only the predictor of interest in each respective hypothesis and a joint specification with all moderators and task fixed effects.

Figure~\ref{fig:app_moderators} plots the univariate OLS results while Table~\ref{tab:moderators} reports both specifications in full detail.  At first glance, political knowledge, news consumption, and higher education appear positively correlated with crowd--human agreement while survey duration appears negatively correlated. 
However, these results are not robust. While we find directional support for H4a,b and c, these correlations are not only marginal but are not robust when we include task fixed effects. Every moderator becomes indistinguishable from zero under the fixed effect models (see Column 2 in Table~\ref{tab:moderators}). Neither political knowledge, news consumption, education, nor survey duration reliably separates better crowd coders from worse ones. This null is consistent with the paper's central claim that annotation quality is primarily a property of the text and the coding rule, more than of the coder. Overall, we find little evidence that \emph{who} the coder is or their attention predicts how well they agree with the humans, contrary to our preregistered expectations.

%We preregistered these expectations given that we expect respondents with higher education, political knowledge, or news consumption to more closely resemble ``humans'' (the research team), particularly given the difficulty of the task and that numerous papers involve content with political jargon. Further, given that this is an online crowd sourced task where respondents' attention is limited, we expect individual text length and time spent on the survey to predict crowd--human agreement: longer texts and shorter completion times correlate with lower agreement. 

\begin{figure}[H]
    \centering
    \includegraphics[width=0.75\textwidth]{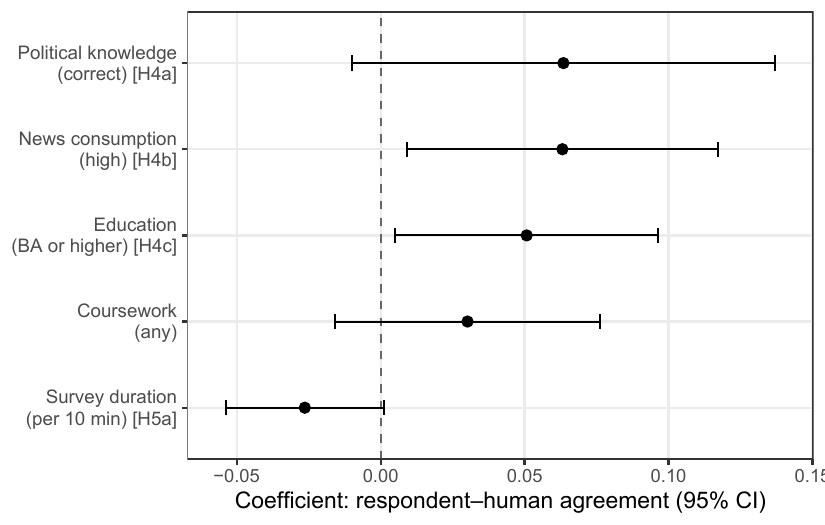}
    \caption{H4/H5, Preregistered univariate associations between coder characteristics and crowd--human agreement. Every association is indistinguishable from zero once task fixed effects are added.}
    \label{fig:app_moderators}
\end{figure}

\begin{table}[H]
\centering
\caption{Exploratory moderators of crowd coding quality (H4/H5)}
\label{tab:moderators}
\begin{tabular}{@{}lcccccc@{}}
\toprule
 & \multicolumn{3}{c}{(1) Univariate (preregistered)} & \multicolumn{3}{c}{(2) Joint, task fixed effects} \\
\cmidrule(lr){2-4} \cmidrule(lr){5-7}
Moderator & Est. & SE & $p$ & Est. & SE & $p$ \\
\midrule
\multicolumn{7}{@{}l}{\emph{Panel A: Respondent's agreement with humans ($N=165$ respondents)}} \\
\addlinespace[2pt]
H4a: Political knowledge (correct) & $0.063$ & $0.037$ & $0.090$ & $0.020$ & $0.044$ & $0.659$ \\
H4b: News consumption (high) & $0.063$ & $0.027$ & $0.022$ & $0.018$ & $0.020$ & $0.387$ \\
H4c: Education (BA or higher) & $0.051$ & $0.023$ & $0.030$ & $0.033$ & $0.037$ & $0.382$ \\
\phantom{H4d:} Coursework (any) & $0.030$ & $0.023$ & $0.196$ & $-0.018$ & $0.026$ & $0.503$ \\
H5a: Survey duration (per 10 min) & $-0.026$ & $0.014$ & $0.060$ & $0.003$ & $0.009$ & $0.755$ \\
\midrule
\multicolumn{7}{@{}l}{\emph{Panel B: Per-text crowd--human agreement ($N=280$ texts)}} \\
\addlinespace[2pt]
H5b: Text length (per 100 words) & $-0.010$ & $0.006$ & $0.067$ & $-0.003$ & $0.003$ & $0.376$ \\
\midrule
Task fixed effects & \multicolumn{3}{c}{No} & \multicolumn{3}{c}{Yes} \\
Standard errors & \multicolumn{3}{c}{Classical} & \multicolumn{3}{c}{CR2, clustered by task} \\
\bottomrule
\end{tabular}
\par\smallskip
\begin{minipage}{0.92\textwidth}
\footnotesize\emph{Notes:} Panel A's dependent variable is each respondent's mean agreement with the three human coders over their 20 texts; Panel B's is the per-text mean crowd--human agreement for the 280 texts of the 14 papers (Schub v2 arm excluded). Column (1) follows the preregistered specification using plain OLS with one moderator at a time and classical standard errors. In column (2) we introduce task fixed effects and CR2 standard errors clustered by task (15 tasks; Satterthwaite degrees of freedom). The Panel B regression likewise adds task fixed effects and CR2 clustering by task (14 tasks) to the univariate text-length specification. Moderators: political knowledge $=$ correct answer on the manifesto knowledge check; high news consumption $=$ follows news daily or a few times a week; education $=$ bachelor's degree or higher; coursework $=$ any response other than ``No'' to the related-coursework item; duration $=$ total survey duration. A joint test of the five Panel A moderators gives $F(5,14)=0.30$, $p=0.907$.
\end{minipage}
\end{table}

% Finally, we present a full summary of our preregistered crowd annotation results in Table~\ref{tab:preregistered_tests} below.

\subsubsection{Robustness to Quality Checks and Exclusions }\label{app:crowd_robustness}

We conducted a series of robustness checks to address the threat of AI use in survey research \citep{Westwood2025}. In our preregistration and Section~\ref{app:crowd_procedure}, we applied a set of inclusion/exclusion criteria to screen for our final sample of 165: (1) we sample only from respondents who indicated English as a native language on Connect, given the difficulty of the task; (2) we automatically screen out respondents that fail any of our attention or bot checks; (3) we sampled only from respondents with an approval rate of 95\% or more \citep{PeerVosgerauAcquisti2014, kennedy_shape_2020}; (4) We exclude respondents that exceed a maximum time spent of 48 minutes (Connect’s default maximum time) and drop any duplicate responses that are identical to others. Ultimately, we removed one respondent that exceeded the maximum time and found no respondents with duplicate responses across all questions; the other exclusions are automatic screens on Connect and Qualtrics.

Besides the above preregistered exclusions, we conducted robustness checks with two additional filters: we dropped respondents that straight-line all text classification responses (answered all 20 texts as ``Yes'' or all as ``No'') and those that score below 0.5 on Qualtrics' CAPTCHA scores. Both checks are soft exclusions given that they do not lead to a respondent being screened out based on our preregistration but they nonetheless raise data quality concerns and we thus remove them sequentially and rerun our main analyses without those responses. 

Table~\ref{tab:exclusion-sensitivity} reports the available headline results across the various exclusion specifications, from the preregistered sample ($N = 165$) to a strict sample that drops all flagged respondents ($N = 156$). The available results retain the same conclusions: the H1 slope (crowd disagreement rising with LLM disagreement) is $0.339$ in the main sample and $0.343$ after dropping straightliners, while the H2 $\alpha$ gap (LLMs agreeing with humans more than the crowd) is $0.129$ and $0.120$, respectively. H1 and H2 cannot be updated for the strict sample because the de-identified data omit reCAPTCHA flags. Across all three specifications, the H3a codebook effect (lower crowd Gini under the improved codebook) ranges from $-0.131$ to $-0.117$, while H3b (higher crowd--human percent agreement under that codebook) remains non-significant throughout.

\begin{table}[H]
\centering
\caption{Headline estimates under three respondent-exclusion specifications
($p$-values in parentheses).}
\label{tab:exclusion-sensitivity}
\small
\setlength{\tabcolsep}{4pt}
\sbox0{\begin{tabular}{lccccccc}
\toprule
 & & H1 & H2 & H3a & H3b & H6a & Crowd--human \\
Specification & $N$ & slope & gap & $\Delta$Gini & $\Delta$agr.\ & gap & agreement \\
\midrule
Main: preregistered exclusions & 165 & 0.339 & 0.129 & $-0.131$ & 0.055 & 0.137 & 0.650 \\
 &  & ($<$ 0.001) & (0.006) & (0.005) & (0.167) & ($<$ 0.001) &  \\
\addlinespace
Drop flagged straightliners & 162 & 0.343 & 0.120 & $-0.131$ & 0.055 & 0.141 & 0.653 \\
 &  & ($<$ 0.001) & (0.009) & (0.005) & (0.167) & ($<$ 0.001) &  \\
\addlinespace
\quad $+$ low reCAPTCHA score & 156 & --- & --- & $-0.117$ & 0.060 & 0.138 & 0.652 \\
 &  & --- & --- & (0.011) & (0.142) & ($<$ 0.001) &  \\
\bottomrule
\end{tabular}}
\usebox0
\par\smallskip
\begin{minipage}{\wd0}
\footnotesize\emph{Notes:} H2 uses Krippendorff's $\alpha$; H3b and the final crowd--human column retain preregistered percent agreement. The final row additionally drops respondents with a Qualtrics reCAPTCHA bot-score below $0.5$. %Its available cells are retained from the authors' prior run because the de-identified package does not contain the reCAPTCHA flag needed to regenerate it. H1 and H2 are unavailable for the expanded LLM panel and updated $\alpha$ measure.
\end{minipage}
\end{table}

\subsubsection{Additional results under an alternative measure of within-text disagreement}\label{app:entropy_crowd}

We reanalyze the findings that use Gini impurity under the normalized Shannon entropy introduced in Section~\ref{app:entropy_expert}. Here, we compute entropy per text over the crowd codings rather than over the LLM predictions.
%Because both measures are symmetric about $p=0.5$ and strictly increasing in $\min(p,1-p)$, entropy is a monotone transform of Gini impurity in this binary setting. The substitution therefore rescales disagreement without reordering texts, so we report it to show that our conclusions do not depend on which dispersion index we chose, not as independent evidence. Rank-based versions of these tests are identical under the two measures by construction; the group means, Welch $p$-values, and paired differences reported below are the statistics the substitution could in principle have moved. Entropy ranges to $1$ where Gini impurity ranges to $0.5$ and rises more steeply near unanimity, so entropy differences run roughly twice their Gini counterparts and only the ratios are comparable across the two measures. Unlike $\alpha$, entropy is defined per text, so it also covers H1.

Table~\ref{tab:crowd_disagreement_entropy} reports crowd disagreement on texts where the humans agree against texts where they disagree, alongside the same split by the LLM sampling stratum used to draw the survey texts. We observe that similar to Gini impurity, crowd entropy rises from $0.537$ on the least contested texts to $0.800$ on the most contested, as defined by LLM disagreement that informed our sampling of texts for the crowd survey (H6a). Likewise, crowd entropy rises from $0.593$ where the humans agree to $0.789$ where they disagree (H6b).

Splitting by study reliability, the text heterogeneity finding holds among both the Higher Reliability and Lower Reliability studies. Among the Lower Reliability studies, crowd entropy rises from $0.584$ to $0.784$ when moving from the least to most contested LLM stratum, and from $0.606$ where humans agree to $0.798$ where they disagree. Both contrasts are statistically significant ($p<0.001$).

We also reanalyze Finding 3 on codebook effects using entropy. We find that crowd entropy falls from $0.862$ under the original codebook to $0.631$ under the improved one (paired, one-sided $t$-test across the same 20 texts, $p =0.011$), reproducing the direction and the significance of the Gini impurity estimate for H3a. Therefore, we find that using Gini impurity or entropy as a measure of within-text heterogeneity does not change our substantive findings for the crowd annotation results.

\begin{table}[H]
\centering
\caption{Crowd disagreement (normalized Shannon entropy) by human agreement and by LLM sampling stratum}
\label{tab:crowd_disagreement_entropy}
\small
\sbox0{\setlength{\tabcolsep}{4pt}\begin{tabular}{lccccc}
\toprule
 & \multicolumn{2}{c}{Mean crowd entropy} & & & \\
\cmidrule(lr){2-3}
\multicolumn{6}{@{}l}{\emph{Panel A: By human agreement}} \\
 & Humans unanimous & Humans split & Diff. & $p$ & $N$ \\
\midrule
All studies & 0.593 & 0.789 & 0.196 & $<$ 0.001 & 172 / 108 \\
\quad Higher Reliability ($\alpha > 0.4$) & 0.581 & 0.780 & 0.199 & $<$ 0.001 & 89 / 51 \\
\quad Lower Reliability ($\alpha < 0.4$) & 0.606 & 0.798 & 0.192 & $<$ 0.001 & 83 / 57 \\
\midrule
\multicolumn{6}{@{}l}{\emph{Panel B: By LLM sampling stratum}} \\
 & Least contested & Most contested & Diff. & $p$ & $N$ \\
\midrule
All studies & 0.537 & 0.800 & 0.263 & $<$ 0.001 & 140 / 140 \\
\quad Higher Reliability ($\alpha > 0.4$) & 0.490 & 0.816 & 0.326 & $<$ 0.001 & 70 / 70 \\
\quad Lower Reliability ($\alpha < 0.4$) & 0.584 & 0.784 & 0.200 & $<$ 0.001 & 70 / 70 \\
\bottomrule
\end{tabular}}
\usebox0
\par\smallskip
\begin{minipage}{\wd0}
\footnotesize\emph{Notes:} Cells are the mean normalized Shannon entropy $-p\log_2 p - (1-p)\log_2(1-p)$ of each text's share $p$ of crowd Yes codes, an alternative to the Gini impurity of Table~\ref{tab:h6-strata}. $N$ counts texts (unanimous\,/\,split in Panel A, least\,/\,most contested in Panel B); $p$-values are two-sided Welch $t$-tests over texts. Schub (2022) v2 arm excluded (280 texts, 14 studies).
\end{minipage}
\end{table}

\newpage

\section{Details of Ambiguity Aware Bounds}\label{app:sensitivity}

This section extends the ambiguity-aware bounds introduced in Section~\ref{step3} beyond the mean and then applies the framework to the empirical analysis in \cite{arias2022securitizes}.

\subsection{Generalization beyond Average}
As in the main text, suppose that we have $N$ texts, indexed by $i = 1, \ldots, N$. Let $Y_i$ denote the true annotation for text $i$. Although $Y_i$ is latent, we observe an annotation produced by either a human or an LLM, denoted by $\hat Y_i$. We allow $Y_i$ to be undefined. 
When it is well-defined, however, we assume that its support, denoted by $\mathcal{Y}$, is bounded. 
Thus, the annotation may be discrete, such as a yes-or-no classification, or continuous but bounded, such as a feeling thermometer score ranging from 0 to 100.

We are interested in a parameter $\theta$ characterized by the estimating equation
\begin{align*}
\E[\bm{m}(\bD_i, Y_i; \theta)] = 0,
\end{align*}
where $\bD_i$ denotes the collection of all other variables entering the estimating equation. If there are no such variables, then $\bD_i = \emptyset$. For example, when the parameter of interest is $\theta = \E[Y_i]$, 
\begin{align*}
\bm{m}(\bD_i, Y_i; \theta) = Y_i - \theta.
\end{align*}
This formulation encompasses a wide variety of models, including regression models.

As described in the main text, we assume that researchers read a random subset of texts, indicated by $S_i = 1$, and annotate an ambiguity indicator $A_i \in \{0,1\}$, which equals one if text $i$ is ambiguous and the value of $Y_i$ is therefore undefined. We further assume that, whenever researchers determine that an LLM has failed to follow the coding instructions (i.e., if there exists some $i$ with $A_i = 0$ in which $Y_i \neq \hat Y_i$), they annotate the corresponding ground-truth label $Y_i$. We formalize this assumption as follows.

\begin{assumption}[Random Sampling]\label{random_S}
Researchers randomly select the texts to read such that
\begin{align}
S_i \ \indep \ \{A_i,Y_i\} \mid \ \hat{Y}_i, \hat{A}_i, \bD_i.
\end{align}
Moreover, the sampling probability $\pi := \Pr(S_i = 1 \mid \hat{Y}_i, \hat{A}_i, \bD_i)$ is known to the researchers.
\end{assumption}

Under this assumption, we can derive the identified set by the same procedure as in the main paper. Specifically, for each candidate value $\theta$, define the oracle bounding moments
\begin{align*}
\underline{\bm{M}}_N(\theta)
&= \frac{1}{N}\sum_{i=1}^N \Bigl[(1 - A_i)\,\bm{m}(\bD_i, Y_i;\theta)
   + A_i \min_{y \in \cY} \bm{m}(\bD_i, y;\theta)\Bigr], \\
\overline{\bm{M}}_N(\theta)
&= \frac{1}{N}\sum_{i=1}^N \Bigl[(1 - A_i)\,\bm{m}(\bD_i, Y_i;\theta)
   + A_i \max_{y \in \cY} \bm{m}(\bD_i, y;\theta)\Bigr],
\end{align*}
where the minimum and maximum are taken element-wise. The identified set is
\begin{align*}
\Theta_I = \{\theta : \underline{\bm{M}}_N(\theta) \le \bm{0} \le
\overline{\bm{M}}_N(\theta)\},
\end{align*}
which includes any value $\theta$ that is consistent with the data
if and only if some assignment of defensible annotations to the ambiguous
texts solves the estimating equation. 

Because $A_i$ and, for unambiguous texts, $Y_i$ are observed only when
$S_i=1$, we apply the design adjustment directly to each complete
bounding-moment contribution. Specifically, define the feasible bounding
moments $\underline{\bm{M}}^{\dagger}_N(\theta)$ and
$\overline{\bm{M}}^{\dagger}_N(\theta)$ by
\begin{align*}
\underline{\bm{M}}^{\dagger}_N(\theta)
&=
\frac{1}{N}\sum_{i=1}^N
\Biggl[
(1-\hat A_i)\bm{m}(\bD_i,\hat Y_i;\theta)
+\hat A_i\min_{y\in\cY}\bm{m}(\bD_i,y;\theta)
\\
&\qquad\qquad
+\frac{S_i}{\pi}
\Biggl\{
(1-A_i)\bm{m}(\bD_i,Y_i;\theta)
+A_i\min_{y\in\cY}\bm{m}(\bD_i,y;\theta)
\\
&\qquad\qquad\qquad
-(1-\hat A_i)\bm{m}(\bD_i,\hat Y_i;\theta)
-\hat A_i\min_{y\in\cY}\bm{m}(\bD_i,y;\theta)
\Biggr\}
\Biggr],
\\[4pt]
\overline{\bm{M}}^{\dagger}_N(\theta)
&=
\frac{1}{N}\sum_{i=1}^N
\Biggl[
(1-\hat A_i)\bm{m}(\bD_i,\hat Y_i;\theta)
+\hat A_i\max_{y\in\cY}\bm{m}(\bD_i,y;\theta)
\\
&\qquad\qquad
+\frac{S_i}{\pi}
\Biggl\{
(1-A_i)\bm{m}(\bD_i,Y_i;\theta)
+A_i\max_{y\in\cY}\bm{m}(\bD_i,y;\theta)
\\
&\qquad\qquad\qquad
-(1-\hat A_i)\bm{m}(\bD_i,\hat Y_i;\theta)
-\hat A_i\max_{y\in\cY}\bm{m}(\bD_i,y;\theta)
\Biggr\}
\Biggr].
\end{align*}
Here, when $A_i=1$, the term involving $Y_i$ is not evaluated because
the corresponding bounding value is determined by the minimum or maximum
over $\cY$. 
%Under Assumption~\ref{random_S}, these feasible bounding moments are design-unbiased for their oracle counterparts:
%\begin{align*}
%\E\!\left[\underline{\bm{M}}^{\dagger}_N(\theta)\right]
%&=
%\underline{\bm{M}}_N(\theta),
%&
%\E\!\left[\overline{\bm{M}}^{\dagger}_N(\theta)\right]
%&=
%\overline{\bm{M}}_N(\theta),
%\end{align*}
%where the expectation is taken with respect to the random selection of the reviewed texts. 
We then obtain the feasible analogue of the identified set as
\begin{align*}
\Theta_I^{\dagger}
=
\left\{
\theta:
\underline{\bm{M}}^{\dagger}_N(\theta)
\le \bm{0}
\le
\overline{\bm{M}}^{\dagger}_N(\theta)
\right\}.
\end{align*}

{
A key advantage of our ambiguity-aware bounds is that the estimand retains a clear substantive interpretation: it is the range of downstream quantities consistent with all substantively defensible classifications of the ambiguous texts. This differs from approaches that propagate disagreement among coders as statistical uncertainty. For example, the quasi-Bayesian procedure proposed by \citet{egami2024using} repeatedly samples one of the human annotations for each text, re-estimates the downstream quantity, and pools the resulting draws. Such an approach is appropriate when disagreement represents uncertainty about which annotation should be used for an otherwise well-defined target. It does not, however, resolve the form of ambiguity considered here, in which multiple classifications are substantively defensible and no unique annotation is defined. Averaging over those classifications instead defines a new quantity that depends on how competing interpretations are weighted. Our bounds avoid imposing such weights and instead characterize the full range of downstream results consistent with the defensible classifications.
}

\subsection{Empirical Application}\label{sec:ambiguity_application}

We finally illustrate the ambiguity-aware bounds by revisiting
\citet{arias2022securitizes}, in which a text-based measure serves as the
outcome of a linear regression. This paper asks which states frame climate change
as a security issue in the United Nations, a process known as
\emph{climate securitization}. The study argues that securitization follows
agenda-control incentives: P5 states (Permanent members of the United Nations: China, France, Russia, United Kingdom, United States) benefit from moving climate change
toward the Security Council, whereas highly vulnerable states such as Small Island Developing States (SIDS) may resist such a shift because it would reduce their agenda control. The outcome underlying this claim is constructed from text, because each climate-related speech segment must be classified as securitizing or not.
We revisit this application because, as our replication shows, the concept of securitization is demanding even for attentive human coders. Some segments, such as those invoking existential harm without referring to security institutions, remain genuinely ambiguous (Table~\ref{tab:examples_ambiguous}).

Following the original analysis, we estimate the linear probability model
\begin{align}
Y_i
    = \beta_1 \,\text{P5}_{c(i)}
    + \beta_2 \,\text{SIDS}_{c(i)}
    + \bX_{c(i),t(i)}^{\tp}\bgamma
    + \delta_{t(i)}
    + \varepsilon_i,
\label{eq:arias_reg}
\end{align}
where $Y_i$ is the true securitization label of segment $i$, delivered by
country $c(i)$ in year $t(i)$. For genuinely ambiguous segments, this label
may be undefined. The vector $\bX_{c,t}$ contains country-level controls,
including public concern about climate change, democracy, experienced
climate disasters, changes in average temperature, voting affinity with
the United States, military expenditure, logged GDP per capita, and logged
population, and $\delta_t$ denotes year fixed effects.

To implement the bounds, we first develop a codebook and conduct an independent
human review of a simple random sample of 200 segments.\footnote{We drew 200
segments at random for human review. One is a European Union observer statement
that lies outside the country-level analysis corpus because it carries no
country-year covariates, so 199 reviewed segments enter the later analysis
frame.} To develop the codebook, one of the three authors reviewed the relevant
literature on climate securitization and formulated the coding rules.\footnote{This includes \citet{BuzanWaeverDeWilde1998}, a canonical book on securitization, as well as other works that review or discuss the concept of securitization, including \citet{Levy1995,Ronnfeldt1997,Vogler2023}.} Using
this codebook, all three authors independently recorded both a binary
classification and whether the codebook left a substantively defensible
alternative classification for each segment. Finally, the author who conducted
the literature review independently reclassified all segments for which at
least one of the three authors identified ambiguity or for which the initial
binary classifications disagreed. In the end, 24 texts are classified as ambiguous.

Next, we annotate all 4,525 segments using seven open-weight LLMs (Mistral~7B, LLaMA~3.1-8B, Qwen2.5-7B, Qwen2.5-14B, Gemma~3-12B, Gemma~3-27B, and GPT-OSS-20B) using the updated prompt introduced in Appendix~\ref{app:complete_codebook}. We take their majority vote as $\hat Y_i$ and construct the ambiguity proxy $\hat A_i$ from cross-model disagreement, measured by Gini impurity.

Using these annotations, we calculate the ambiguity-aware bounds using the following procedure.
For a target coefficient $\beta_k$, let $w_{ki}$ denote the OLS weight
placed on segment $i$'s outcome when estimating that coefficient. Because
OLS is linear in the outcome, if all labels were well-defined, the
coefficient could be written as
\begin{align}
\hat\beta_k = \sum_{i=1}^N w_{ki}Y_i.
\end{align}
The weights depend only on the observed regressors, including the controls
and year fixed effects, and are therefore known once the regression design
is fixed.\footnote{Let $\bZ$ denote the full design matrix and let
$\mathbf e_k$ be the unit vector selecting coefficient $k$. Then $w_{ki}$
is the $i$th element of
$\mathbf e_k^{\tp}(\bZ^{\tp}\bZ)^{-1}\bZ^{\tp}$.}
In particular, assigning segment $i$ the label $Y_i=1$ rather than $Y_i=0$
changes $\hat\beta_k$ by $w_{ki}$. Thus, when $w_{ki}<0$, assigning the
label 1 lowers the coefficient, whereas when $w_{ki}>0$, assigning the
label 1 raises it. The oracle lower bound therefore assigns label 1 to
ambiguous segments with negative weights and label 0 to those with positive
weights; the oracle upper bound does the reverse.

Applying the design adjustment directly to the complete lower- and
upper-bound contributions gives the feasible bounds
\begin{align}
\hat\beta_k \in
\Biggl[\,
&\sum_{i=1}^{N} w_{ki}
\Biggl[
(1-\hat A_i)\hat Y_i
+\hat A_i\,\mathbbm{1}\{w_{ki}<0\}
\nonumber\\[-2pt]
&\qquad\qquad
+\frac{S_i}{\pi}
\Biggl\{
(1-A_i)Y_i
+A_i\,\mathbbm{1}\{w_{ki}<0\}
-(1-\hat A_i)\hat Y_i
-\hat A_i\,\mathbbm{1}\{w_{ki}<0\}
\Biggr\}
\Biggr],
\nonumber\\[4pt]
&\sum_{i=1}^{N} w_{ki}
\Biggl[
(1-\hat A_i)\hat Y_i
+\hat A_i\,\mathbbm{1}\{w_{ki}>0\}
\nonumber\\[-2pt]
&\qquad\qquad
+\frac{S_i}{\pi}
\Biggl\{
(1-A_i)Y_i
+A_i\,\mathbbm{1}\{w_{ki}>0\}
-(1-\hat A_i)\hat Y_i
-\hat A_i\,\mathbbm{1}\{w_{ki}>0\}
\Biggr\}
\Biggr]
\,\Biggr].
\label{eq:arias_bounds}
\end{align}
The terms inside the design adjustment are evaluated only for reviewed
segments with $S_i=1$, and $Y_i$ is required only when the reviewed segment
is unambiguous, so that $A_i=0$. Under
Assumption~\ref{random_S}, the two endpoints in
Equation~\eqref{eq:arias_bounds} are design-unbiased for the corresponding
oracle lower and upper endpoints.

For comparison, we consider two benchmark approaches. First, we estimate the original regression using the majority vote of the LLM annotations as the outcome, without accounting for ambiguity. Second, we apply design-based supervised learning (DSL; \citealt{egami2024using}), treating the majority vote of the human annotations as the ground-truth label. DSL corrects for measurement error in LLM annotations, but it does not accommodate fundamental ambiguity in the classification task because it assumes that each text has a uniquely defined gold-standard label.

\begin{figure}[!t]
    \centering
    \includegraphics[width=1.0\linewidth]{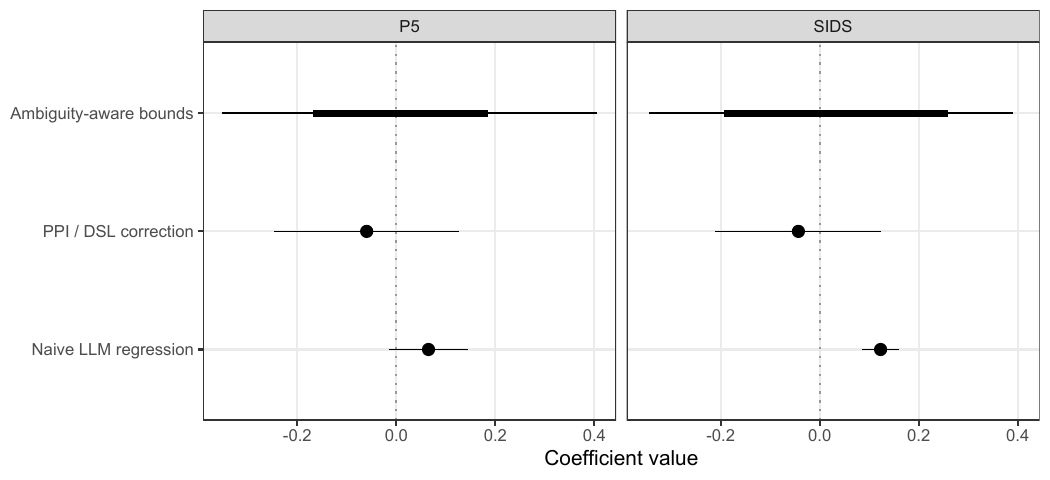}
    \caption{Coefficients of $\beta_1$ (P5) and $\beta_2$ (SIDS), along with their 95\% confidence intervals, based on (i) the proposed ambiguity-aware bounds, 
    (ii) PPI / DSL correction (which corrects the measurement error in LLM annotations by regarding human annotation as ground truth), and
    (iii) the naive LLM regression (which uses LLM predictions as the outcome without accounting for the underlying ambiguity of the concepts). }
    \label{ambiguity_bounds_results}
\end{figure}

Figure~\ref{ambiguity_bounds_results} compares the ambiguity-aware bounds with the naive LLM regression and DSL. Both benchmark approaches yield substantially more precise estimates because they impose a unique label on every text and therefore do not account for conceptual ambiguity in the annotation task. This precision can be misleading: the underlying concept of climate securitization remains ambiguous under the paper's definition,\footnote{In the paper, climate securitization is defined as ``emphasizing the security dimensions of an issue, specifically the language of existential threat'' and indicating the United Nations Security Council's (UNSC) jurisdiction over an issue \citep[p3]{arias2022securitizes}. However, some texts are borderline cases given this definition. Take the following excerpted text as an example: 
\begin{quote}\emph{
``[W]e must all assume our fair share of responsibility in effective international cooperation to find solutions to global warming and climate change, which pose, as never before, a grave threat to humankind’s survival''.}
\end{quote}
While the text does not explicitly call climate change a security issue, it refers to it as a ``grave threat to humankind's survival'' a phrase that could be interpreted as ``existential threat''; on the other hand, it contains no element of indicating UNSC jurisdiction, which is also absent from most texts that were classified as ``securitization'' in the paper's original structural topic model.

Another ambiguous example arises from the text discussing multiple issues simultaneously, including both climate change and security, leaving room for interpretation about whether the two topics are referred to jointly or separately. For example: 

\begin{quote}
\emph{ ``Alone we can do little, but together we can achieve much. The challenges that we face globally — from climate change to refugees to war and violence — require urgent action now.''}
\end{quote}
} and alternative defensible classifications of the ambiguous texts can even reverse the sign of the estimated coefficient.

These results underscore the importance of developing a precise operationalization and translating it into a clear codebook. Researchers should resolve avoidable ambiguity through substantive refinement of the coding rules before turning to statistical adjustment. The ambiguity-aware bounds are designed for the ambiguity that remains after this process, making explicit the extent to which the substantive conclusion depends on unresolved borderline cases.

\end{document}